\documentclass[11pt]{article}

\usepackage[margin=1in]{geometry}

\usepackage{microtype}
\usepackage{graphicx}
\usepackage{booktabs} %
\usepackage{float} %
\usepackage{xcolor} %

\usepackage{hyperref}

\usepackage{amsmath}
\usepackage{amssymb}
\usepackage{mathtools}
\usepackage{amsthm}
\usepackage{bbm}

\usepackage{graphicx}
\usepackage{subcaption}

\usepackage[capitalize,noabbrev]{cleveref}

\usepackage[round,authoryear]{natbib}

\usepackage{algorithm}
\usepackage{algorithmic}

\theoremstyle{plain}

\theoremstyle{definition}

\theoremstyle{remark}

\usepackage{tikz}
\usetikzlibrary{shapes.geometric, arrows}

\tikzstyle{startstop} = [rectangle, rounded corners, minimum width=3cm, minimum height=1cm,text centered, draw=black, fill=red!30]
\tikzstyle{process} = [rectangle, minimum width=3cm, minimum height=1cm, text centered, draw=black, fill=blue!30]
\tikzstyle{arrow} = [thick,->,>=stealth]
\tikzstyle{data} = [ellipse, minimum width=3cm, minimum height=1cm, text centered, draw=black, fill=green!30]
\tikzstyle{decision} = [diamond, minimum width=3cm, minimum height=1cm, text centered, draw=black, fill=yellow!30]

\newcommand{\ex}[2]{{\ifx&#1& \mathbb{E} \else \underset{#1}{\mathbb{E}} \fi \left[#2\right]}}
\newcommand{\pr}[2]{{\ifx&#1& \mathbb{P} \else \underset{#1}{\mathbb{P}} \fi \left[#2\right]}}
\newcommand{\var}[2]{{\ifx&#1& \mathrm{Var} \else \underset{#1}{\mathrm{Var}} \fi \left(#2\right)}}

\renewcommand{\epsilon}{\varepsilon}  %

\newcommand{\QAfull}{Question-Asking compression}
\newcommand{\QA}{QA-compression}
\newcommand{\BLC}{BL-CoT}  %
\newcommand{\sd}[1]{\,{\scriptsize$\pm$#1}}  %

\newif\ifshowcomments
\showcommentstrue  %

\begin{document}

\title{Haiku to Opus in Just 10 bits: \\LLMs Unlock Large Compression Gains}

\author{
Roy Rinberg$^{1}$\thanks{Corresponding author: \texttt{royrinberg@g.harvard.edu}. Work conducted while RR was participating in ML Alignment \& Theory Scholars (MATS).}
\quad Annabelle Michael Carrell$^{2}$\footnote{Work done while AC was a visiting student at Harvard.} \quad Simon Henniger$^{1}$ \\ \quad Nicholas Carlini$^{3}$ \quad Keri Warr$^{3}$ \\[0.5em]
$^{1}$\small{Harvard University} \quad $^{2}$University of Cambridge
\quad $^{3}$Anthropic}

\date{}

\maketitle

\begin{abstract}

  We study the compression of LLM-generated text across lossless and lossy regimes, characterizing a compression-compute frontier where more compression is possible at the cost of more compute. For lossless compression, domain-adapted LoRA adapters can improve LLM-based arithmetic coding by $2\times$ over compression with the base LLM alone, achieving SOTA on lossless compression of LLM-generated text. %
  For lossy compression, prompting a model for a succinct rewrite then applying arithmetic coding can achieve compression ratios of approximately 0.03, a $2\times$ improvement over compressing the original response.

   We further introduce Question-Asking compression (QA), an interactive lossy protocol inspired by the game ``Twenty Questions''. A small model iteratively refines its response by asking yes/no questions to a stronger model, transferring exactly one bit per answer. On 8 benchmarks spanning math, science, and code, 10 binary questions recover 23\% to 72\% of the capability gap between a small and large model on standard benchmarks and 7\% to 38\% on harder benchmarks, achieving compression ratios of 0.0006 to 0.004. This is over $100\times$ smaller than prior LLM-based compression \citep{delétang2024languagemodelingcompression}, suggesting that interactive protocols can transfer knowledge far more efficiently than
  transmitting full responses.

\end{abstract}

\section{Introduction}
\label{sec:intro}

Machine learning (ML) and data compression are often referred to as two sides of the same coin: ML aims to learn patterns that enable accurate prediction and generation, while compression seeks efficient representations that minimize the number of bits required to encode information.
In entropy coding, compression is a function that takes two inputs, a probability distribution and a sequence, and then it encodes the sequence according to the probability distribution, such that more likely characters are encoded using fewer bits.
Impressively, compression techniques such as arithmetic coding have been proven to achieve an optimal compression ratio \citep{cover1991elements, shannon1948mathematical}. Thus, achieving the optimal lossless compression ratio is predicated on determining the right probability distribution for a sequence.

When framed in this way, a second observation comes into light: the primary goal of Large Language Models (LLMs) is to produce probability distributions from which to sample tokens. Recent work has operationalized the connection between compression and ML by leveraging large language models (LLMs) to produce probability distributions that, when combined with arithmetic coding, enable effective text compression \citep{bellard_tszip, valmeekam2023llmziplosslesstextcompression,delétang2024languagemodelingcompression}.

These techniques already boast impressive compression ratios; whereas a common compression scheme like GZIP \citep{gzip} achieves a compression ratio of around $0.33$ on most text, LLM-based arithmetic coding schemes achieve a compression ratio closer to $0.1$ on text that is in distribution for the model, like Wikipedia \citep{delétang2024languagemodelingcompression}.

We observe that the requirement of producing a \textit{lossless} compression scheme is overly restrictive in the LLM chatbot setting. The goal of most chat interfaces is not to produce a specific sequence of text, but rather to produce a high-quality response \citep{zhang2025watermarkssandimpossibilitystrong}, and a user with no direct access to the underlying model cannot tell whether the response they received has been altered. This is a key difference between lossy compression for text and for images: image data has a well-defined ground truth from which lossy compression deviates, whereas text typically does not. Modern language models can summarize, rewrite, and edit text in an open-ended manner, enabling a new form of lossy compression in which the surface form may change but the semantic content and usefulness are largely preserved.

Additionally, it has long been recognized that certain forms of knowledge can be compactly represented through interaction. In theoretical computer science, this insight appears in the study of interactive proof systems, where interaction between two parties can dramatically reduce the amount of information that must be communicated to convey or verify a concept \citep{shafi1989,goldreich1997quantifying}. We adapt this connection to LLM communication in Section~\ref{sec:20q}.

In this work we introduce three compression methods for LLM-generated text, each independently a novel contribution and together composing into a natural cascade of fallbacks. Our headline method, Question-Asking compression (\QA), is a lossy interactive protocol that can in some cases encode an LLM's response in fewer than 10 bits, achieving compression ratios up to $100\times$ smaller than prior LLM-based lossless methods. \QA{} works well on many tasks, but does not preserve quality on every problem. To account for this, we also introduce two complementary methods that serve both as standalone baselines and as fallbacks: response-rewriting, a lossy method where the model is prompted for a succinct restatement, and LoRA-adapted arithmetic coding, which meaningfully improves prior LLM-based lossless compression. Together, these three methods form a cascade where \QA{} is attempted first; if it fails to meet a quality threshold, the response can be rewritten lossily; and as a final fallback, the response can be losslessly compressed.

\subsection{Contributions}

In this work, we investigate methods for compressing LLM-generated text using a smaller language model. Our focus is on minimizing the number of bits required to encode text produced by a large model when the decoder has access to a weaker shared model.
We explore approaches that trade additional computation for improved compression ratios, in both the lossless and lossy regimes.

Concretely, this paper makes the following contributions:\footnote{Code available at \url{https://github.com/RoyRin/model-compression-bit-limiting}.}

\begin{enumerate}
    \item \textbf{Iterative Binary Questions.} We propose a novel protocol akin to the children's game ``20 Questions'' where a small language model iteratively improves its response through yes/no questions answered by a more capable model, which we call \QAfull~(\QA) (\Cref{sec:20q}). We evaluate this protocol on 8 benchmarks on science, math, and coding tasks. In 10 bits or fewer, models recover 23--72\% of the performance gap on easier benchmarks and 7--38\% on harder benchmarks, achieving compression ratios of 0.004 to 0.0006, over $100\times$ beyond prior SOTA \citep{delétang2024languagemodelingcompression}.

    \item \textbf{Response Rewriting.} As an intuitive baseline for lossy compression, we explore two lossy compression approaches: Shortest-of-N selection (generate multiple candidates, select the most compressible) and Summarization (explicitly request a succinct rewrite). Shortest-of-N achieves compression ratios of 0.073 ($N{=}1$) to 0.063 ($N{=}10$); Summarization achieves 0.039 to 0.034, a $1.9\times$ improvement over Shortest-of-N. Selecting for compressibility does not degrade answer quality (\Cref{sec:shortest-of-n}), when evaluated on a coding and math benchmark.

    \item \textbf{Lossless Compression via LoRA Adaptation.} As an innovation on existing work on lossless compression, we demonstrate that domain-specific LoRA adapters improve arithmetic coding compression by $2\times$ over baseline models, achieving SOTA on lossless compression of LLM-generated text. A RAG-based router selects the appropriate adapter from the prompt alone, retaining a $1.9\times$ improvement (\Cref{sec:lossless}), when evaluated on the LMSYS and WildChat dataset.

    \item \textbf{Bit-Limited Chain-of-Thought (\BLC{}).} In order to isolate the effect of the question-asking process, we introduce \BLC{} as a baseline that separates the contribution of refinement from the contribution of the larger model, where the small model (e.g.\ Haiku) asks and answers its own questions (\Cref{sec:20q}).

\end{enumerate}

\subsection{Paper Organization}

\Cref{sec:background} reviews background on arithmetic coding, LoRA adaptation, and interactive compression. \Cref{sec:20q} introduces a novel interactive question-asking protocol for extreme compression.
 \Cref{sec:shortest-of-n} introduces response rewriting as a lossy compression baseline. And \Cref{sec:lossless} introduces our proposed lossless compression method based on domain-adapted LoRAs, which we present as a strong standalone baseline for comparison with the interactive protocol.  We conclude with a discussion of applications to model security and future directions in \Cref{sec:conclusions}. 

\paragraph{Models.} Throughout this work, we use three tiers of language models from the Anthropic Claude~4.5 family as reference implementations: \textbf{Haiku} (small), \textbf{Sonnet} (medium), and \textbf{Opus} (large). These span a representative capability range, with Haiku serving as the small language model (SLM) in our compression protocols and Opus as the large language model (LLM) that provides guidance. Our methods are not specific to these models; we use the Anthropic API as a convenient testbed for studying compression across capability tiers.

\section{Background \& Related Work}
\label{sec:background}

\subsection{Arithmetic Coding and LLM Compression} \label{sec:arithmetic-coding}

At the most abstract level, a compression scheme is a function that takes two inputs: a probability distribution, and a sequence; it then encodes the sequence according to the probability distribution, such that more likely elements are encoded with fewer bits than less likely elements. Arithmetic coding is a specific compression scheme that achieves near-optimal compression \citep{langdoncompression,arithmeticcoding}. 

The source coding theorem from information theory actually establishes the minimum achievable compression ratio (on average), and states that for a discrete memoryless source with distribution $P$, the minimal length for compression is equivalent to the entropy $H(P) = \mathbb{E}_{x \sim P}\!\left[-\log_2 P(x)\right]$. 
The compressibility is directly determined by how closely the probability  distribution approximates the true data-generating distribution.
When data sampled from distribution $P$ is encoded using a probabilistic model $Q$, the expected number of bits required is given by the cross-entropy $H(P, Q) = \mathbb{E}_{x \sim P}\!\left[-\log_2 Q(x)\right]$.

At a similarly abstract level, LLMs can be broken down into an algorithm that repeatedly does two things: generate a probability distribution, then sample from it. When viewed from this perspective, a natural idea is to take the probability distribution from an open-source LLM and use the as the probability-distribution  used for encoding in arithmetic coding.

A standard metric in compression is the compression ratio, defined as the size of the compressed representation relative to the original data. Recent work demonstrates that large language models paired with arithmetic coding yield state-of-the-art text compression results, achieving compression ratios of roughly 0.1 on Wikipedia corpora \citep{bellard_tszip, valmeekam2023llmziplosslesstextcompression, delétang2024languagemodelingcompression}.

Lossless compression can be evaluated on a single axis: how compressed the outputs are. Lossy compression requires at least two axes, the compression ratio and the degree to which the compressed text retains the semantics of the original. Summarization is a closely related task, as a summary is shorter than the input but intended to preserve its main meaning. The summarization literature has grappled with the same evaluation difficulty: how to measure whether a shorter text retains the information of the original. One approach is to generate questions about the source text using one LLM, and then verify that a separate model answers the questions in the same way when given the original text and when given the summary \citep{wang-etal-2020-asking,dua2025clinicallygroundedagentbasedreportevaluation}. When the prompt has a clear correct answer, this evaluation simplifies: it suffices to check whether the compressed text still produces the correct answer.

\subsection{Interactive Compression via Binary Questions}
\label{sec:20-questions}

It's easy to see how interaction can drastically reduce communication complexity through the children's game \emph{20 Questions}. One player thinks of an object, and the other identifies it by asking a sequence of yes--no questions, each conveying a single bit of information.By choosing questions adaptively, the player can identify one object among millions using roughly 20 bits—far fewer than would be required to specify the object directly (and on the order of $<2$ tokens of information).

This game reflects a general principle: interaction enables information to be revealed conditionally, focusing communication on what is most informative given previous answers. In complexity theory, this principle manifests as a separation between non-interactive and interactive verification. Non-interactive proofs, consisting of a single message from a prover to a verifier, characterize the class $\mathrm{NP}$. Allowing the verifier to ask adaptive questions yields interactive proofs, which characterize $\mathrm{IP} = \mathrm{PSPACE}$, a class believed to be strictly larger than NP \citep{goldwasser1989,ipshamir1992}.

In Section~\ref{sec:20q}, we apply this idea to LLM communication. A small model adaptively asks binary questions that are answered by a large model, allowing substantial knowledge to be transferred using only a small number of bits.

\subsection{LoRA and Domain Adaptation}
\label{sec:lora-background}

Low-Rank Adaptation (LoRA) \citep{hu2021loralowrankadaptationlarge} is a parameter-efficient fine-tuning method that adapts a pretrained model by learning low-rank update matrices. For a pretrained weight matrix $W_0 \in \mathbb{R}^{d \times k}$, LoRA learns $\Delta W = BA$ where $B \in \mathbb{R}^{d \times r}$ and $A \in \mathbb{R}^{r \times k}$ with rank $r \ll \min(d,k)$. This dramatically reduces the number of trainable parameters.

The key insight for compression is that a LoRA adapter fine-tuned on domain-specific text will assign higher probability to in-domain text, thereby reducing the bits required for arithmetic coding. If sender and receiver share a set of domain-specific LoRA adapters, they can achieve better compression by selecting the appropriate adapter for each message.

\section{Interactive Question Asking (\QAfull)} \label{sec:lossy-compression}
\label{sec:20q}

We introduce a novel interactive compression mechanism that adapts the children’s game 20 Questions to the task of LLM knowledge transfer.
 Given a prompt, an SLM first generates an initial response and then formulates $N$ binary yes/no questions about its solution strategy, for example, “Is my approach to step 3 correct?” The LLM answers each question with yes or no, transferring exactly one bit of information per response. The SLM then revises its response by incorporating all $N$ answers. Crucially, the SLM's questions are deterministic given its initial response, the prompt, and fixed hyperparameters. As a result, only the $N$ binary answers must be transmitted. \footnote{In principle, the expected cost per question can be reduced below one bit. If the SLM can predict the base-rate distribution of LLM responses, for example that “Yes” occurs with probability 0.8 for a given question, arithmetic coding over the binary answers yields an expected cost of $H(\text{answer} \mid \text{question})$ bits per question. This improvement is essentially free: if the SLM's calibration is incorrect, the LLM can transmit a single extra bit to signal this, ensuring that the protocol never performs worse than the naive one-bit-per-answer scheme.}

The receiver can reconstruct the full question-answer transcript by running an identical SLM locally, reducing the communication cost to exactly $N$ bits, independent of the response length. An important observation is that in practice the questions do no need to be sent to the LLM, as the LLM can perfectly simulate the SLM, removing the need for more bandwidth. %

\begin{figure*}[t]
    \centering
    \includegraphics[width=1.\linewidth]{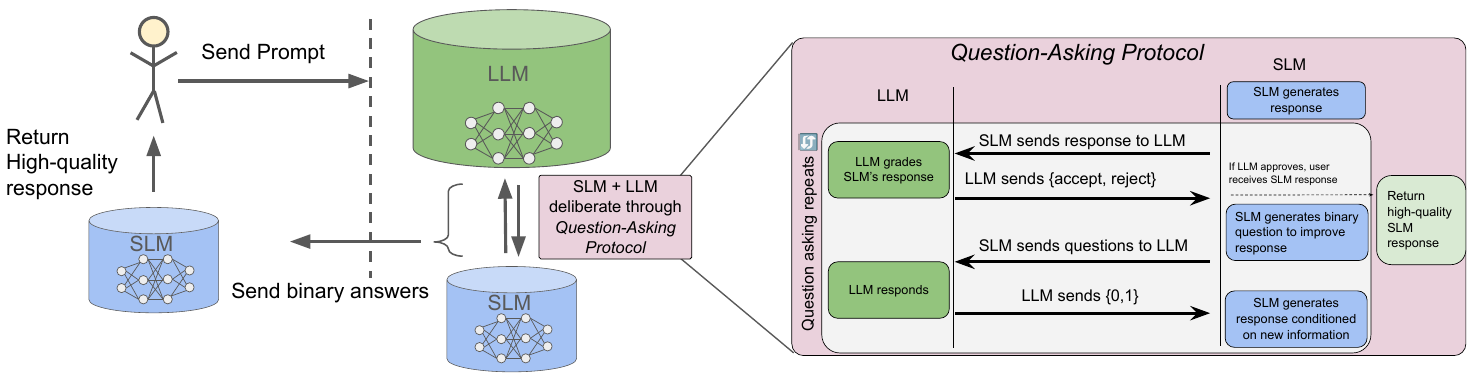}
    \caption{Overview of the compression mechanism and its use in an interactive protocol between an SLM and an LLM.}
    \label{fig:compression-overview}
\end{figure*}

\begin{algorithm}[t]
\caption{Question Asking (QA Protocol).  $a_i$: SLM's answer after round $i$; $q_i$: yes/no question; $r_i \in \{\textsc{Yes}, \textsc{No}\}$: LLM's binary response (1 bit). %
}
\label{alg:qa-protocol}
\begin{algorithmic}[1]
\REQUIRE Problem $P$, small model SLM, large model LLM, budget $N$
\STATE $a_0 \gets \text{SLM}(P)$ \COMMENT{SLM attempts to solve $P$}
\FOR{$i = 1, \ldots, N$}
    \STATE $q_i \gets \text{SLM}(P,\; a_{i-1},\; q_{<i},\; r_{<i})$ \COMMENT{SLM asks a yes/no question}
    \STATE $r_i \gets \text{LLM}(P,\; q_i)$ \COMMENT{LLM answers yes or no (1 bit)}
    \STATE $a_i \gets \text{SLM}(P,\; a_{i-1},\; q_{\leq i},\; r_{\leq i})$ \COMMENT{SLM revises its answer}
    \STATE \textbf{if} $\text{Judge}(a_i) \geq \tau$ \textbf{then break} \COMMENT{Early stop if quality threshold met [\textit{Optional}]}
\ENDFOR
\STATE \textbf{return} $a_i$
\end{algorithmic}
\end{algorithm}

We evaluate on 8 datasets, which we group into \emph{easier benchmarks}\footnote{We define ``easier benchmarks'' as those where Haiku achieves above 66\% baseline accuracy. The exception is GPQA, which we include in this group because its multiple-choice format makes answer verification straightforward, despite Haiku's lower baseline accuracy.} (GSM8K~\citep{cobbe2021gsm8k}, MATH~\citep{hendrycksmath2021}, GPQA (MC)~\citep{rein2023gpqagraduatelevelgoogleproofqa}), where current models achieve high baseline accuracy, and \emph{harder benchmarks} (MBPP~\citep{mbppaustin2021program}, AIME~\citep{AIMOValidationAIME}, Humanity's Last Exam (HLE)~\citep{phan2025humanitysexam}), where even strong models have significant room for improvement. We classify problems by difficulty based on which models solve them at baseline: \emph{Easy} (all correct), \emph{Medium} (Haiku fails, Sonnet/Opus correct), \emph{Hard} (only Opus correct), and \emph{Very Hard} (all fail). Full benchmark descriptions and difficulty distributions are in Appendix~\ref{app:dataset-stats}. Our evaluation spans diverse answer formats: GSM8K (natural language), MATH (equations), AIME (integers 0--999), MBPP (Python code), HLE (open-ended responses), and GPQA (multiple-choice; classified as easier in our analysis).

We also validate our approach with the open-source GPT-OSS-120B model (Appendix~\ref{app:gptoss-results}), and we find qualitatively similar results. This demonstrates that the QA protocol transfers across model families, and importantly shows that this technique can already be used for lossy compression, as all users have access to opensource models.

\paragraph{Evaluation.} We measure the \emph{recovery rate}: the fraction of non-easy problems answered correctly after 10 rounds of binary Q\&A. By construction, Haiku is initially wrong on Medium and Hard problems, so this equals the fraction of incorrect answers that become correct. We denote each configuration as \emph{questioner}$\to$\emph{answerer}$\to$\emph{solver}. Our primary configuration is \textbf{QA} (haiku$\to$opus$\to$haiku): Haiku generates questions, Opus answers them, and Haiku produces a final answer. This measures whether Opus's knowledge can be transferred via 10 binary answers.

However, by letting Haiku ask questions and iteratively revise its answer, we are giving it substantially more computation, and the iterative question-asking process is itself a form of structured chain-of-thought reasoning. Some of the improvement may therefore come from self-refinement rather than from Opus's knowledge. To disentangle these effects, we introduce \textbf{\BLC{}} (\textit{Bit-Limited Chain-of-Thought}, haiku$\to$haiku$\to$haiku): the identical protocol, but with Haiku answering its own questions. Any improvement from \BLC{} is self-refinement, whereas the difference between QA and \BLC{} isolates the value of Opus's knowledge transfer. Table~\ref{tab:qa-accuracy} in Appendix~\ref{app:qa-results-detailed} provides the full per-difficulty breakdown.

\begin{table}[t]
\centering
\caption{Recovery rate (\%, mean\sd{std} over 3 trials) and compression ratio by difficulty subset. \emph{Med.\ + Hard} = problems where Haiku fails but Opus succeeds; \emph{All} = all non-easy problems (includes Very Hard). \BLC{} = haiku$\to$haiku$\to$haiku; QA = haiku$\to$opus$\to$haiku; QA$^+$ = opus$\to$opus$\to$haiku. Bold = highest mean recovery in each row within each subset. Ave.\ Resp.\ Len.\ = mean Opus response length in tokens (\texttt{cl100k\_base} tokenizer). Compression = ratio of 10 QA bits to full Opus response ($\text{tokens} \times \log_2(100{,}277)$ bits).}
\label{tab:qa-asker}
\footnotesize
\setlength{\tabcolsep}{4pt}
\resizebox{\linewidth}{!}{%
\begin{tabular}{l rrrr rrrr r l}
\toprule
 & \multicolumn{4}{c}{Med.\ + Hard} & \multicolumn{4}{c}{All Non-Easy} & & \\
\cmidrule(lr){2-5} \cmidrule(lr){6-9}
Dataset & $n$ & \BLC{} & QA & QA$^+$ & $n$ & \BLC{} & QA & QA$^+$ & \shortstack{Ave.\\Resp.\\Len.} & Compression \\
\midrule
\multicolumn{11}{l}{\emph{Easier benchmarks}} \\
\addlinespace[2pt]
GSM8K            & 110 & 19.1\sd{2.0} & 23.3\sd{1.1} & \textbf{27.6}\sd{1.5} & 145 & 22.5\sd{1.2} & 26.7\sd{0.9} & \textbf{29.7}\sd{1.7} & 195 & 0.0031 \\
MATH (Algebra)   &  85 & 28.6\sd{1.5} & 27.5\sd{2.2} & \textbf{31.0}\sd{1.5} & 259 & 38.1\sd{1.6} & 39.1\sd{1.0} & \textbf{40.0}\sd{1.5} & 373 & 0.0016 \\
MATH (Geometry)  &  51 & 28.1\sd{1.8} & \textbf{37.3}\sd{3.2} & \textbf{37.3}\sd{1.6} & 190 & 33.7\sd{0.4} & \textbf{39.6}\sd{0.5} & 38.2\sd{1.5} & 687 & 0.0009 \\
MATH (Num.\ Thy) &  12 & 66.7\sd{11.8} & \textbf{72.2}\sd{15.7} & 63.9\sd{15.7} &  61 & 38.3\sd{2.0} & \textbf{43.7}\sd{4.3} & 40.4\sd{3.1} & 580 & 0.0010 \\
GPQA (MC)        &  15 & \textbf{40.0}\sd{14.4} & \textbf{40.0}\sd{14.4} & 31.1\sd{6.3} & 152 & 11.8\sd{2.7} & \textbf{13.6}\sd{0.3} & 12.1\sd{2.0} & 606 & 0.0010 \\
\addlinespace[2pt]
\midrule
\multicolumn{11}{l}{\emph{Harder benchmarks}} \\
\addlinespace[2pt]
AIME             &  34 & \textbf{10.8}\sd{1.4} & 6.9\sd{1.4} & 9.8\sd{1.4} &  51 & 18.3\sd{1.8} & 19.0\sd{2.4} & \textbf{22.2}\sd{1.8} & 1{,}083 & 0.0006 \\
MBPP             &  44 & 36.4\sd{1.9} & 37.9\sd{2.8} & \textbf{51.5}\sd{2.8} & 106 & 18.2\sd{1.8} & 18.2\sd{0.4} & \textbf{26.1}\sd{0.4} & 164 & 0.0037 \\
HLE$^\dagger$    & 206 &  6.5\sd{0.6} &  7.9\sd{0.6} & \textbf{13.1}\sd{1.4} & 306 &  5.4\sd{0.6} &  6.5\sd{0.0} & \textbf{9.3}\sd{0.8} & 711 & 0.0008 \\
\bottomrule
\end{tabular}%
}
\vspace{2pt}

{\footnotesize $^\dagger$\,For HLE, Very Hard problems were randomly subsampled to 100 (out of ${\sim}$1{,}900) due to evaluation cost.}
\end{table}

\paragraph{Compression Ratios.} The rightmost column of Table~\ref{tab:qa-asker} shows the compression achieved by the QA protocol: 10 bits (10 binary answers) compared to a full Opus response, measured as $\text{tokens} \times \log_2(|V|) \approx 17$ bits per token using the \texttt{cl100k\_base} tokenizer (an upper bound treating each token as uniform over the vocabulary). The QA scheme achieves compression ratios between 0.0006 and 0.0037, transferring only 10 bits compared to thousands in a full response.\footnote{For multiple-choice tasks like GPQA, this framing is somewhat misleading because, in theory, communicating the correct answer of a multiple-choice question should only require $\log_2(4) = 2$ bits of information. However, empirically the questions the SLM asks do not make reference to the limited choices. We include GPQA for reference but it should not be considered load-bearing for this reason.}

\paragraph{Ablations: the value of question asking.}
In any real implementation, the SLM will serve as both question-asker and final solver. However, the SLM's ability to identify its own reasoning gaps may be a bottleneck: if it cannot formulate targeted questions, even perfect answers provide limited value. To simulate future improvements in question-asking ability, we introduce replace the question-generator with another instance of Opus, while keeping Haiku as the solver, denoted QA$^+$ (opus$\to$opus$\to$haiku). This gives us insight into how this protocol's effectiveness may change as models become more capable. Table~\ref{tab:qa-asker} compares \BLC{}, QA, and QA$^+$ on two subsets: \emph{Med.\ + Hard} problems (where Opus succeeds but Haiku does not, so correct knowledge exists to transfer) and \emph{All Non-Easy} problems (which additionally includes Very Hard problems where even Opus fails). The full $3 \times 3$ breakdown across all questioner--answerer combinations is in Appendix~\ref{app:qa-results-detailed}.

The results above ask all 10 questions in a single batch, with the LLM given the ground-truth answer as reference when responding. An iterative variant lets a judge early-stop the exchange once the SLM's answer is satisfactory, potentially saving bits. We test two judge designs (objective vs.\ comparison scoring), threshold levels, and whether the judge sees the ground-truth answer; see Appendix~\ref{app:judge-ablation}. These variants do not improve over the batch protocol: the judge either early-stops prematurely on easy problems or fails to prevent regression on hard ones.

We also test scaling from 10 to 100 questions; this yields small, inconsistent gains that do not justify the $10\times$ increase in bits (Appendix~\ref{app:qa-100q}).

\paragraph{Results.}
The central question is whether the QA protocol's value comes from the \emph{structure} of iterative refinement or from the \emph{content} of the stronger model's answers. Table~\ref{tab:qa-asker} reveals that the answer depends on task difficulty. 
On easier benchmarks (GSM8K, MATH, GPQA), the majority of the recovery is captured by process of asking questions (\BLC{}), and Haiku does approximately as well as Opus as the LLM or as the Question Asker; the structured protocol alone enables Haiku to self-correct on well-posed math and science problems. 

On harder benchmarks, there is a larger gain in recovery rates when the LLM is large model.
Here, we speculate that the SLM lacks the knowledge to self-correct, so the information in Opus's binary answers becomes  more important, and a stronger question-asker (Opus in QA$^+$)  amplifies the improvement, as it can better identifies the SLM's failure modes. AIME is an exception, with uniformly low recovery across all configurations, suggesting that competition math resists the binary Q\&A format. Example transcripts illustrating the Q\&A interaction are provided in Appendix~\ref{app:qa-transcripts}.

\paragraph{Compression Ratios.} The rightmost column of Table~\ref{tab:qa-asker} shows the compression achieved by the QA protocol: 10 bits (10 binary answers) compared to a full Opus response, measured as $\text{tokens} \times \log_2(100{,}277) \approx 16.61$ bits per token using the \texttt{cl100k\_base} tokenizer (an upper bound treating each token as uniform over the vocabulary). The QA scheme achieves compression ratios between 0.0006 and 0.0037, transferring only 10 bits compared to thousands in a full response.\footnote{For multiple-choice tasks like GPQA, this framing is somewhat misleading: the correct answer contains only $\log_2(4) = 2$ bits of information, so using 10 binary questions to communicate it is technically an \emph{expansion}. The QA scheme is most compelling for open-ended tasks (math, code) where the answer space is large, not for multiple-choice tasks where the answer is already low-entropy.} To put these ratios in context, prior SOTA LLM-based lossless compression achieves approximately 0.08 on enwik9 \citep{delétang2024languagemodelingcompression}, and our own domain-adapted LoRA approach (Section~\ref{sec:lora-compression}) reaches approximately 0.03. The QA scheme's best result of 0.0006 on AIME represents over $100\times$ improvement over prior SOTA, illustrating the compressive power of interactive protocols that exploit shared priors between models rather than encoding the full response text.

\paragraph{Regression on Easy Problems.}
The \QA~protocol should not harm problems the SLM already solves. We verify this by running the protocol on \emph{easy} problems and find that regression roughly matches Haiku's own baseline inconsistency when re-evaluated with no protocol at all. Full results are in Appendix~\ref{app:easy-regression}, Table~\ref{tab:easy-reeval}.

\paragraph{Question Transcript Analysis}

Looking at the transcripts, the small model generates yes/no verification questions that walk through its own solution step by step. Interestingly, 31\% of recovered problems had every answer confirmed as ``yes'' by the frontier model; the structured self-verification alone was enough to fix the error. For example, on a GSM8K tax problem, Haiku asks questions like ``Does the solution correctly apply the 10\% tax rate only to the nonfood subtotal?'' and ``Are all five items accounted for with their correct prices''; the frontier model confirms each one, and Haiku revises its answer anyway. 

In another 37\% of recoveries, the small model asks mostly confirming questions but surfaces one or two key misconceptions. On a GSM8K word problem, Haiku asks ``Does '10 times more than the fourth friend' mean adding 10 times the fourth friend's presses?'', and the frontier model answers ``no'', pinpointing the exact misinterpretation; the other four questions are all confirmed. This pattern, where the small model's approach is mostly sound and just needs a targeted correction, accounts for the majority of successful recoveries.

  When QA fails, the transcripts reveal two characteristic failure modes. In one, the small model verifies all the steps of a wrong solution path. On a
  MATH algebra problem, Haiku carefully checks each step of its factorization and sign analysis, the frontier model confirms them all, but the fundamental setup is wrong and the questions never probe it. In the other failure mode, the small model is too confused to ask useful questions at all. This is particularly stark on code benchmarks (MBPP), where the small model asks about edge cases, type hints, and error handling (``Should the function validate that the radius is non-negative?'', ``Would it be beneficial to add type hints?'') rather than probing its actual bug, however the corrections do not help because they are orthogonal to the real issue.

  Examples of transcripts are laid out in Appendix \ref{app:qa-transcripts}, as well as uploaded to \href{https://huggingface.co/datasets/royrin/model-compression/tree/main/transcripts/iterative-qa-sweep}{HuggingFace}.

\section{Novel Baselines}

This section introduces two simple but effective compression methods that serve as competitive standalone baselines for evaluating our \QA~scheme and other approaches.

\Cref{sec:lossy-compression} explores a lossy baseline based on best-of-$N$ sampling and explicit prompting for succinctness. By selecting the shortest high-quality response, we obtain approximately a $2\times$ improvement in compression ratio. 

\Cref{sec:lossless} presents a lossless baseline that builds on arithmetic coding. Rather than compressing directly with a single language model, we pre-train multiple domain-adapted LoRA adapters and select the most relevant adapter at inference time. This specialization yields roughly a $2\times$ improvement in compression ratio relative to a non-adapted model.

\subsection{Lossy Compression Baselines: Response Rewriting}
\label{sec:shortest-of-n}

We introduce a simple approach to lossy compression is to generate multiple candidate solutions and select the one that compresses best using a standard off-the-shelf arithmetic compression scheme. This should not meaningfully change the distribution of texts that we produce, as we are selecting only from texts generated by the LLM.

\begin{figure*}[t]
    \centering
    \includegraphics[width=.99\linewidth]{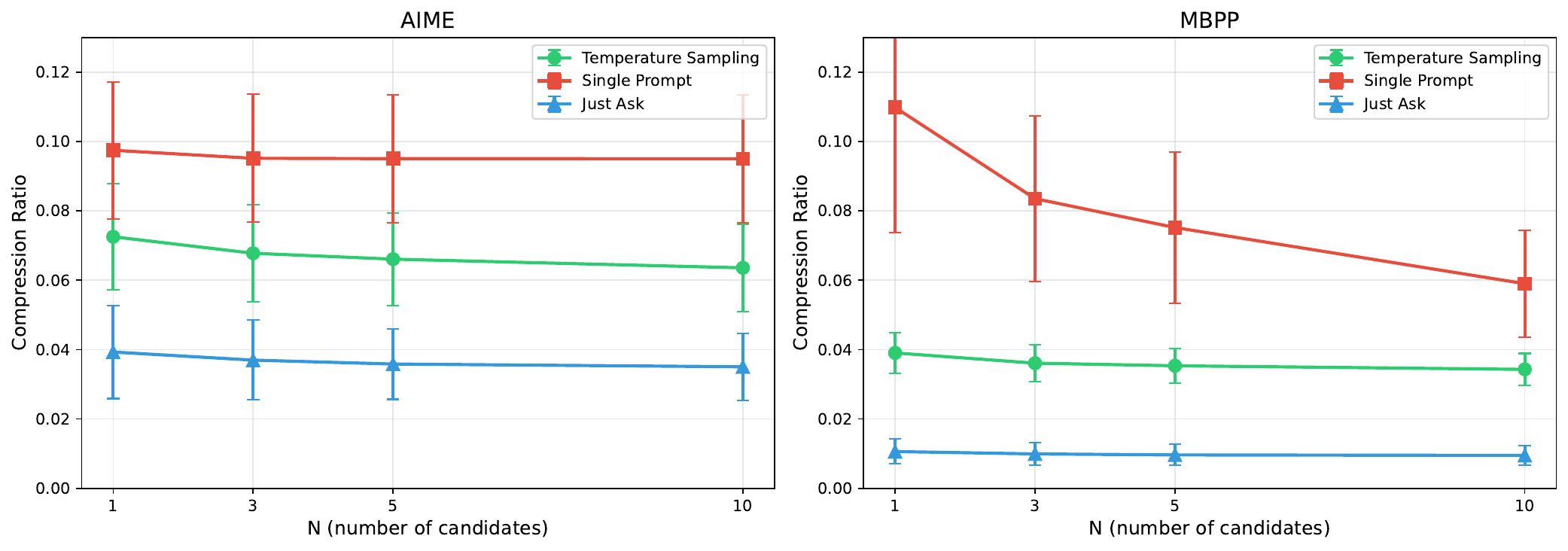}
    \caption{Compression ratio vs.\ number of candidates $N$ on AIME (left) and MBPP (right) problems, using Opus as both generator and compressor. \textbf{Just Ask} (blue) achieves the best compression by explicitly requesting succinct rewrites, roughly halving the compression ratio compared to \textbf{Shortest-of-N} methods. Among Shortest-of-N variants, \textbf{Temperature Sampling} (green) outperforms \textbf{Single Prompt} (red), likely because single-prompt solutions share structural patterns. Lower is better.}
    \label{fig:compression-comparison}
\end{figure*}

\paragraph{Shortest-of-N.}
Generate $N$ candidate solutions and select the one with the lowest compression ratio. We evaluate two variants: \textbf{Temperature Sampling}: Generate $N$ independent solutions by sampling with temperature $T > 0$ (we use $T=0.8$). \textbf{Single-Prompt}: Request $N$ distinct solutions in a single prompt, e.g., ``Provide 5 different approaches to solve this problem.'' 

\paragraph{Summarization.}
\label{sec:just-ask}
Another relatively-straightforward alternative is to \textit{directly request} a succinct rewrite. First generate a verbose solution with full reasoning, then prompt the model to rewrite it ``as succinctly as possible while preserving enough information to infer the correct answer.'' This can be combined with temperature sampling to generate $N$ succinct rewrites and select the shortest.

\paragraph{Results.}
Figure~\ref{fig:compression-comparison} compares these three approaches on AIME (90 competition math problems \citep{AIMOValidationAIME}) and Mostly Basic Python Programs (MBPP; 257 tasks \citep{mbppaustin2021program}) using Opus as the generator and Mistral-7B-Instruct as the compressor \cite{jiang2023mistral7b}, there are full benchmark descriptions are in Appendix~\ref{app:benchmarks}. 

On AIME, temperature sampling achieves compression ratios from 0.073 at $N=1$ to 0.063 at $N=10$ ($1.2\times$ improvement from sampling more candidates). Summarization significantly outperforms both, achieving 0.039 at $N=1$ ($\approx1.9\times$ better than temperature sampling) and 0.034 at $N=10$, roughly halving the compression ratio from a single sample. Results with Haiku as generator are qualitatively similar and shown in Appendix~\ref{sec:haiku-best-of-n}.

The summarized rewrites are approximately 4$\times$ shorter than verbose originals. Interestingly, the shorter text has \emph{higher} bits-per-character: by stripping away predictable boilerplate (``Step 1:'', ``Therefore,''), summarization produces denser text that is harder for the compression model to predict on a per-character basis. However, the reduction in length is larger than the increase in per-character complexity, so the net effect is still roughly 2$\times$ better total compression. See Appendix~\ref{sec:knowledge-density} for a more in-depth analysis of this trade-off.

We find that selecting for compressibility does not degrade answer quality: across all three generation strategies, choosing the most compressible candidate yields accuracy within a few percentage points of random selection. See Appendix~\ref{sec:accuracy-vs-compression} for the full analysis.

\subsection{Lossless Compression through Domain-Adapted LoRAs}
\label{sec:lora-compression} \label{sec:lossless}

While LLMs provide powerful probabilistic models for text, arithmetic coding achieves optimal compression only when the coding distribution matches the data distribution. When text drawn from $P$ is encoded using model $Q$, the expected code length increases by the KL divergence $\mathrm{KL}(P \,\|\, Q)$. A general-purpose LLM may therefore compress domain-specific data suboptimally. 

LoRA \citep{hu2021loralowrankadaptationlarge} has proven particularly effective for domain adaptation, enabling a single base model to be specialized to multiple domains via lightweight adapters. If the sender and receiver share a collection of domain-specific LoRA adapters, they can improve compression by selecting the adapter that best matches the input distribution for each message. We investigate whether domain-specialized language models can achieve better compression rates than general-purpose models.

\paragraph{Experimental Setup.}
We use LMSYS-Chat-1M~\citep{zheng2023lmsyschat1m}, a large-scale dataset of real user conversations with LLMs, as our primary testbed. We cluster the dataset into 50 topic clusters using K-means on Qwen embeddings. For ten clusters, we fine-tune a LoRA adapter~\citep{hu2021loralowrankadaptationlarge} on the training split (rank=16, $\alpha$=32) using Mistral-7B-Instruct as the base model \citep{jiang2023mistral7b}; we note that Mistral-7B is distinct from the generators (Opus / Haiku) used elsewhere.%
We evaluate compression performance on held-out test samples with at least 200 tokens, measuring compression ratio using different methods: (1) \textbf{Baseline}: the base model without any LoRA, (2) \textbf{Gzip}, (3) \textbf{Correct LoRA}: the adapter trained on the same cluster, (4) \textbf{Wrong LoRAs}: an adapter trained for each of the nine remaining incorrect clusters, and (5) \textbf{RAG LoRA}: the adapter selected by a retrieval-based router (described below). All the clusters and the the LoRAs are shared publicly on \href{https://huggingface.co/datasets/royrin/model-compression}{HuggingFace}.

For all methods, we report compression ratio as compressed size divided by uncompressed size.\footnote{For GZIP, this is compressed bytes divided by original bytes. For arithmetic coding, this is output bits divided by input bits, where input bits equals $\log_2(\text{vocabulary size}) \times \text{number of tokens}$. The arithmetic coder's bitstream can be packed into bytes with negligible overhead ($<1$ byte per message), so these metrics are directly comparable.}

\subsubsection{RAG-based LoRA Router}
\label{sec:rag-lora}

In practice, one does not have access to the exact distribution that a text was generated from, so selecting the correct LoRA requires an automatic routing method. We propose an approach akin to retrieval-augmented generation (RAG) \citep{lewis2021retrievalaugmentedgenerationknowledgeintensivenlp}.

\paragraph{Routing Protocol.} We build a FAISS index over representative texts from each cluster using Qwen embeddings. Each index entry maps to the corresponding cluster's LoRA adapter. For each input text, we embed the text, retrieve the top index, and use the corresponding LoRA adapter.

\paragraph{Routing Accuracy.} We evaluate RAG routing accuracy on held-out test samples using both the full text (prompt and response) and only the prompt. The router correctly identifies the domain cluster approximately 72\% of the time with full text and 55\% with prompt only. For cluster-specific results, see Appendix~\ref{app:lora-results}.

\paragraph{Results.}
Table~\ref{tab:lora_per_cluster} reports per-cluster compression ratios. On average, the correct LoRA achieves a compression ratio of 0.09, a $2.0\times$ improvement over the base model (0.18). Even the average wrong LoRA improves to 0.13 ($1.4\times$). The RAG-selected LoRA achieves 0.10 ($1.9\times$), approaching the oracle despite only 72\% routing accuracy. Gzip achieves compression ratios of approximately $0.75$--$0.85$, which are weaker than those typically reported for gzip on raw natural-language text. We emphasize that, in our setting, gzip is applied after tokenization rather than directly to raw text. Moreover, our reported ratio is not the conventional byte-level gzip compression ratio, i.e., compressed bytes divided by original bytes, but instead is computed under our token-normalized ratio; see Appendix~\ref{app:metric-comparison} for further discussion and Appendix~\ref{app:classical-compressors} for additional classical-compressor baselines (zstd, LZMA2, Brotli).

\begin{table}[t]
\centering
\caption{Compression performance (compression ratio) on LMSYS-Chat. Correct LoRA = adapter trained on the same cluster; Average = mean across the 9 incorrect adapters; Worst = worst-performing incorrect adapter; RAG LoRA = adapter selected by the retrieval-based router from prompt only.}
\label{tab:lora_per_cluster}
\footnotesize
\setlength{\tabcolsep}{4pt}
\resizebox{\linewidth}{!}{%
\begin{tabular}{ll|cc|ccc|c}
\toprule
 & & \multicolumn{2}{c|}{Baseline} & \multicolumn{3}{c|}{LoRA} & RAG LoRA \\
Cluster & Topic & Gzip & Base Model & Correct & Average & Worst & Prompt \\
\midrule
0 & General Chat & 0.86 & 0.21 & 0.11 & 0.13 & 0.14 & 0.11 \\
1 & Creative Writing & 0.76 & 0.18 & 0.11 & 0.12 & 0.13 & 0.11 \\
2 & Code/Technical & 0.77 & 0.28 & 0.10 & 0.20 & 0.21 & 0.10 \\
3 & Academic/Education & 0.78 & 0.16 & 0.09 & 0.11 & 0.11 & 0.10 \\
4 & Roleplay/Fiction & 0.70 & 0.11 & 0.05 & 0.08 & 0.08 & 0.05 \\
5 & Business/Professional & 0.80 & 0.20 & 0.09 & 0.13 & 0.14 & 0.09 \\
6 & Philosophy/Ethics & 0.76 & 0.19 & 0.09 & 0.14 & 0.14 & 0.09 \\
7 & Science/Math & 0.73 & 0.15 & 0.09 & 0.11 & 0.11 & 0.10 \\
8 & Translation/Language & 0.75 & 0.18 & 0.08 & 0.13 & 0.14 & 0.10 \\
9 & Casual Q\&A & 0.75 & 0.15 & 0.10 & 0.12 & 0.13 & 0.10 \\
\midrule
\textbf{Overall} & & \textbf{0.77} & \textbf{0.18} & \textbf{0.09} & \textbf{0.13} & \textbf{0.13} & \textbf{0.10} \\
\bottomrule
\end{tabular}%
}
\end{table}
\paragraph{Analysis.}
The per-cluster results reveal that improvement varies by topic. Crucially, the correct LoRA consistently outperforms the wrong LoRAs across all clusters, validating that topic-matched adapters capture domain-specific patterns that transfer to improved compression. Wrong adapters still help because the clusters all share common language patterns (e.g.\ they are from the same general distribution), so adapters learn partly transferable improvements. A significantly more diverse dataset would likely show wrong LoRAs having zero or negative effect.

We replicate these results on WildChat~\citep{zhao2024wildchat1mchatgptinteraction}, a dataset of 1M ChatGPT interaction logs (Table~\ref{tab:rag_compression_wildchat}, Appendix~\ref{app:lora-results}). The correct LoRA achieves a compression ratio of 0.10, a $1.7\times$ improvement over baseline (0.17), slightly lower than LMSYS ($2.0\times$), consistent with WildChat's greater topic diversity within clusters.

\section{Discussion}
\label{sec:conclusions}

Using the framing of the \QA~framework, we suggest a shift in how the presence of open-source models alters the structure of communication we have access to. When two parties share access to the same model, they effectively share a rich internal representation of the world. Communication no longer requires transmitting an entire idea from scratch; instead, it is sufficient to transmit only the difference between what the shared model would already generate and the intended content. In this sense, shared models enable an unusually low-bandwidth form of communication grounded in a common representation.

More broadly, question-asking serves not only as a compression primitive but also as a mechanism for probing conditional information. Quantifying the conditional information between two beings (human or LLM) is generally abstract and difficult to measure directly. By framing communication as a sequence of binary questions, QA-style interaction provides an operational way to bound how much additional information one agent must transmit to another about a specific concept. The number of questions required offers a concrete proxy for their conditional information gap.

\subsection{Applications to Model Security}
\label{sec:egress-limiting}

One promising defense for protecting model weights is \emph{egress limiting}: restricting the total    
  number of bits that exit a datacenter \citep{nevo2024securing,anthropic2025asl3,greenblatt2024preventing}. This rests on a simple            
  information-theoretic observation: if only $10$ GB leaves a datacenter, an adversary cannot steal more 
  than $10$ GB of anything, regardless of their attack vector. However, strict bit limits can harm
  legitimate users who need to receive model outputs. As we have explored in this paper, natural language is extremely compressible, which can be used to empower defenders to apply stricter egress limits, as model weights are generally more difficult to compress than text.\footnote{Though recent research seeks to find more compressible forms of model weights \citep{brown2025how}.}

\begin{figure}
        \centering
        \includegraphics[width=0.85\linewidth]{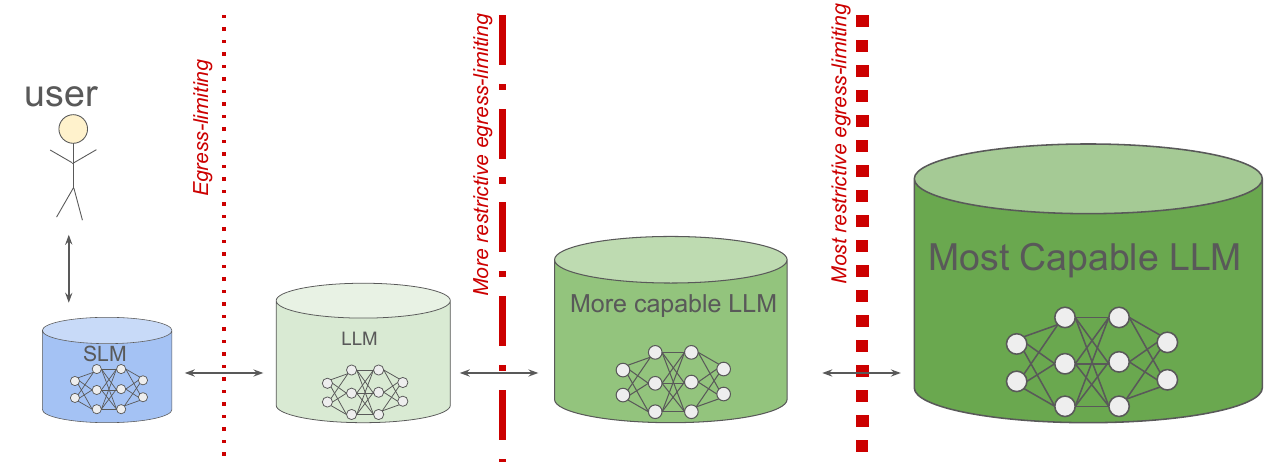}
        \caption{Layered egress-limiting architecture. More capable models are subject to stricter bit limits. A frontier model (e.g., Opus) transmits only binary yes/no answers, while a smaller gateway model (e.g., Haiku) generates full responses locally. The user receives high-quality answers via \QA~ recovery, while the frontier model's effective bandwidth remains severely constrained.}
        \label{fig:layered-egress}
    \end{figure}

This potentially motivates a layered architecture where more capable models face stricter egress limits
      (Figure~\ref{fig:layered-egress}).\footnote{However, this assumes risk scales with capability. If all models
      beyond a certain size are considered \textit{extremely dangerous}, and require severe egress limiting, then the layered approach might offer no additional benefit.} The most capable models communicate only via bit-limited question asking, providing answers to a less capable gateway model, which handles full text generation.

\subsection{Limitations}
\label{sec:limitations}

\paragraph{Selection bias in shortest-of-N.}
When we select the most compressible response from $N$ candidates, we are implicitly biasing toward certain kinds of outputs. We measure accuracy on AIME to verify that compression-selected responses are not systematically worse (Appendix~\ref{app:best-of-n}), but other quality dimensions (clarity, completeness, pedagogical value) are not evaluated. It is plausible that optimizing for compressibility selects for responses that are terse or formulaic in ways that standard accuracy metrics do not capture.

\paragraph{Benchmark-specific evaluation.}
Our \QA~ results are evaluated on math, code, and science benchmarks with verifiable answers. These domains are well-suited to the binary question format because correctness is unambiguous. The protocol's effectiveness on open-ended generation tasks (creative writing, summarization, advice) is untested and likely quite different, since these tasks lack clear ground truth against which to evaluate yes/no questions.

\paragraph{LLM-as-judge reliability.}
The iterative \QA~ protocol relies on the LLM to answer binary questions about solution correctness. When the LLM is itself wrong (as on very-hard problems), its answers can actively mislead the SLM. We observe this empirically: on frontier benchmarks where the judge fails to reliably evaluate solutions, regression on easy problems is severe (Section~\ref{sec:20q}). The protocol's utility is therefore bounded by the LLM's own accuracy on the task.

\subsection{Broader Impacts}
\label{sec:broader-impacts}
The compression primitives we develop are agnostic to what is being compressed, which creates several dual-use concerns. (i) An adversary with inference access can exfiltrate more proprietary information per bit than naive transmission allows, using the same protocols we characterize for legitimate egress-limited deployments. (ii) In the lossy regime, the user receives a reconstructed response rather than the LLM's original output, and errors introduced by the rewrite or recovery step can compound in unpredictable ways for downstream tasks. This matters most in high-stakes domains where users may not realize a substitution has occurred. (iii) Selecting outputs by compressibility may systematically deprioritize phrasings that are longer for legitimate reasons, including pedagogical explanations, multilingual content, and accessible restatements. We do not propose mitigations in this work but flag side-channel monitoring and fairness audits of downstream tasks as necessary for any deployed compression scheme.

\subsection{Future Work}
\label{sec:future-work}

\paragraph{Improving question quality for \QA.}
The \QA~ protocol is inherently limited by the SLM's ability to ask useful questions and integrate the answers. Our results show that on frontier benchmarks, upgrading the question-asker to a stronger model yields meaningful gains (Section~\ref{sec:20q}), suggesting that question quality is a bottleneck. Future work could explore improving the SLM's question-asking ability, whether through prompt engineering, fine-tuning on transcripts from successful \QA~ sessions, or reinforcement learning on recovery rate.

  \paragraph{Alternative interactive schemes.}
  
  Our \QA~protocol relies entirely on the SLM's ability to ask useful questions. However, this is limiting:
  consider "What are the first 5 decimal places of $\sqrt{17}$?" A small model lacks the capacity to
  derive the answer, so no sequence of binary questions will help. A natural extension is to allow the
  LLM to \emph{inject} information, providing information that the SLM would not discover through questioning alone. Other low-bit interaction schemes are also possible. For example, the SLM could generate $N$ candidate
  responses, and the LLM could select the best one in $\log_2 N$ bits.\footnote{We explored this direction but found that this meaningfully lengthened responses, eliminating compression gains. However, we view this as an
  engineering challenge rather than a fundamental barrier, and we believe alternative strategies may yet make such schemes viable.}

\paragraph{Robustness to Non-determinism.}
Both our lossless and interactive compression schemes assume that the encoder and decoder reproduce identical model outputs. In practice, modern LLM inference pipelines are not strictly deterministic: numerical discrepancies arise from floating-point effects, hardware variation, batching, and low-level kernel implementations \citep{yuan2025understandingmitigatingnumericalsources,he2025nondeterminism,rinberg2025verifyingllminferencedetect,karvonen2025difrinferenceverificationdespite}. 

For arithmetic coding, exact agreement between encoder and decoder probability distributions is required. Because decoding is autoregressive, even small mismatches can compound and lead to catastrophic decoding failure. The \QA~scheme is similarly sensitive: if the large model fails to perfectly simulate the small model and generates a different question, synchronization breaks and the protocol collapses. Recent work such as PMATIC \citep{pmatic} introduces auxiliary ``helper bits'' to realign encoder and decoder distributions. In our work, we proposed a complementary approach, \emph{block emission encoding}, which periodically emits the midpoint of the current coding interval and restarts from a fresh $[0,1]$ range after every $N$ bits of information, thereby limiting error propagation to a single block; however, more improvements need to be made for this to be a robust solution.\footnote{A more detailed discussion appears in Appendix~\ref{app:arithmetic-coding}, and we outline three deployment regimes that recover the protocol's egress guarantee under non-determinism in Appendix~\ref{app:nondeterminism}.} Despite these advances, designing compression schemes that are robust to cross-machine non-determinism remains an open problem.

\paragraph{Combining lossless and lossy methods.}
Our lossless and lossy methods currently operate independently, but they could be composed. For example, one could apply LoRA-adapted arithmetic coding to summarized text (combining rewriting with domain adaptation), or use LoRA-based compression scoring as the selection criterion in shortest-of-$N$. These combinations may yield compression ratios between the lossy and lossless regimes while maintaining higher fidelity than pure lossy methods.

\paragraph{Scaling to open-ended tasks.}
Our \QA~ evaluation focuses on benchmarks with verifiable answers. Extending the protocol to open-ended tasks (code review, writing assistance, multi-step reasoning) requires new evaluation methodology, since there is no single correct answer against which to measure recovery. The protocol itself transfers naturally to these settings, but measuring whether the transferred bits are useful is an open problem.

\newpage
\section{Acknowledgements}\label{sec:acks}

RR's work was supported by a grant from Coefficient Giving, and this work was conducted in part during RR's participation in MATS. SH's work was supported by the Harvard Paulson SEAS Prize Fellowship and the German Academic Fellowship Organization, funded by the German Federal Ministry for Economic Affairs and Energy. Compute credits provided by the ML Alignment \& Theory Scholars (MATS) program as well as the Kempner Institute for the Study of Natural and Artificial Intelligence at Harvard University.

\bibliographystyle{plainnat}
\bibliography{citations}

\newpage
\appendix

\section{Extended Background}
\label{app:background}

This appendix provides extended discussion of background concepts summarized in Section~\ref{sec:background}.

\subsection{Arithmetic Coding: Extended Discussion}
\label{app:arithmetic-coding}

Arithmetic coding is an entropy coding technique that achieves near-optimal lossless compression by encoding a sequence of symbols using a number of bits that approaches the Shannon entropy of the source \citep{witten1987arithmetic}. Given a probabilistic model over a discrete alphabet, arithmetic coding converts probability estimates directly into code lengths, assigning shorter codes to more likely sequences and longer codes to less likely ones.

In the context of language modeling, let $p$ denote a language model and let $x_1, x_2, \ldots, x_n$ be a sequence of tokens. When arithmetic coding is applied using the model's conditional probabilities, the resulting code length is approximately
\[
\sum_{i=1}^{n} -\log_2 p(x_i \mid x_1, \ldots, x_{i-1}),
\]
which corresponds to the model's cross-entropy evaluated on the sequence. Consequently, compression performance is governed almost entirely by the quality of the language model's predictive distribution.

Building on this observation, several recent works have demonstrated that large language models can serve as effective entropy models for text compression when paired with arithmetic coding, achieving strong empirical results in both online and offline settings \citep{bellard_tszip, valmeekam2023llmziplosslesstextcompression, delétang2024languagemodelingcompression}.

\subsection{Distribution Mismatch: Extended Discussion}
\label{app:distribution-mismatch}

Arithmetic coding achieves optimal compression only when the coding distribution matches the true data-generating distribution. When text drawn from a distribution $P$ is encoded using a model $Q$, the expected code length increases by the Kullback-Leibler divergence $\mathrm{KL}(P \,\|\, Q)$, yielding an expected cost of $H(P) + \mathrm{KL}(P \,\|\, Q)$. As a result, a general-purpose LLM may perform suboptimally when compressing domain-specific data whose statistics deviate from those seen during pretraining.

This motivates domain adaptation approaches. We address this challenge through lightweight domain adaptation using LoRA fine-tuning (Section~\ref{sec:lora-compression}), which significantly improves compression performance by reducing distributional mismatch.

\subsection{Non-Determinism in LLM Inference: Extended Discussion}
\label{app:nondeterminism}

Arithmetic coding requires exact agreement between the probability distributions used by the encoder and decoder. Modern LLM inference pipelines, however, are not strictly deterministic: small numerical discrepancies arise across runs due to floating-point effects, hardware variation, kernel implementations, or batching \citep{he2025nondeterminism,rinberg2025verifyingllminferencedetect,karvonen2025difrinferenceverificationdespite}. Even vanishingly small probability mismatches can cause decoder desynchronization and catastrophic decoding failure.

This fragility is amplified by the sequential nature of arithmetic coding. Because decoding is autoregressive, any mismatch alters the decoder's context and thus the predicted distribution for all subsequent tokens, causing errors to accumulate over long sequences. Many prior works do not observe this issue because they run compression and decompression on a single machine, where the cross-run non-determinism is very small \citep{delétang2024languagemodelingcompression,valmeekam2023llmziplosslesstextcompression}. However, real-world deployments necessarily involve separate encoder and decoder machines, where such discrepancies become unavoidable.

Robust inference under nondeterminism is not the focus of this paper, but we outline three deployment regimes that recover the protocol's egress guarantee, in roughly increasing order of operational cost.

\paragraph{Determinism holds.} LLM inference can be made deterministic in practice with matched inference stacks (same hardware, kernels, batch sizes, and precision) and fixed seeds \citep{yuan2025understandingmitigatingnumericalsources}. In this regime \QA{} works as described, and only $N$ bits of egress are required.

\paragraph{Data diodes for ingress.} When determinism cannot be guaranteed, the SLM's questions can be transmitted to the LLM over an unconstrained \emph{inbound} channel, for example a unidirectional data diode that permits high-bandwidth ingress while strictly limiting egress. Under this deployment, the LLM still returns only $N$ binary answers, so the egress bit budget is preserved regardless of how nondeterministic the SLM's question generation is.

\paragraph{PMATIC self-compression.} When neither matched determinism nor an asymmetric channel is available, the SLM can compress its own transcript using its own logits as the arithmetic-coding distribution. Nondeterminism in the arithmetic coder can be handled by schemes like PMATIC \citep{pmatic}, which uses auxiliary helper bits to absorb GPU-level numerical drift. Empirically, we observe roughly $40\times$ compression on QA transcripts, so even in the fully nondeterministic case the egress cost grows modestly rather than catastrophically; see Appendix~\ref{app:pmatic-results}.

\section{Benchmarks, Evaluation Setup, and Dataset Statistics}
\label{app:dataset-stats}

\subsection{Benchmarks}
\label{app:benchmarks}

We evaluate our lossy compression methods on eight benchmarks spanning mathematics, code generation, science, and expert-level reasoning. These were chosen to cover a range of task types and difficulty levels, all with verifiable answers that enable automatic evaluation.

\begin{itemize}
    \item \textbf{GSM8K} \citep{cobbe2021gsm8k}: 1{,}319 grade-school math word problems requiring multi-step arithmetic reasoning. Answers are integers.
    \item \textbf{MATH} \citep{hendrycksmath2021}: Competition-level mathematics. We evaluate three subjects: Algebra (1{,}187 problems), Geometry (479), and Number Theory (540). Answers are typically expressions in \LaTeX{} (e.g., $\boxed{5\sqrt{5}}$).
    \item \textbf{GPQA (MC)} \citep{rein2023gpqagraduatelevelgoogleproofqa}: 198 graduate-level science questions (physics, chemistry, biology) in multiple-choice format with four options.
    \item \textbf{MBPP} \citep{mbppaustin2021program}: 257 entry-level Python programming tasks from the sanitized split. Evaluated by executing generated code against unit tests.
    \item \textbf{AIME} \citep{AIMOValidationAIME}: 90 problems from the American Invitational Mathematics Examination. Answers are integers in $\{0, \ldots, 999\}$.
    \item \textbf{HLE} \citep{phan2025humanitysexam}: 2{,}158 text-only problems from Humanity's Last Exam, spanning diverse expert-level domains. Includes both exact-match and multiple-choice questions.
\end{itemize}

\subsection{Difficulty Classification}
\label{app:difficulty-classification}

We classify each problem into one of four difficulty tiers based on which models solve it correctly in a single baseline evaluation (temperature 0):
\begin{itemize}
    \item \textbf{Easy}: All three models (Haiku, Sonnet, Opus) answer correctly.
    \item \textbf{Medium}: Haiku fails, but Sonnet and/or Opus succeed.
    \item \textbf{Hard}: Haiku and Sonnet fail, but Opus succeeds.
    \item \textbf{Very Hard}: All three models fail.
\end{itemize}
This classification is based on a single deterministic run per model; as we discuss in Appendix~\ref{app:easy-regression}, this can be unreliable for harder benchmarks (AIME, HLE) where model outputs are not consistently reproducible.

\subsection{Claude 4.5 Model Family Results}

Table~\ref{tab:all-datasets-v45} summarizes model accuracies across all evaluated datasets using the Claude 4.5 model family.

\begin{table}[H]
\centering
\caption{Model accuracy across all datasets (Claude 4.5 family).}
\label{tab:all-datasets-v45}
\begin{tabular}{lrrrr}
\toprule
Dataset & $n$ & Haiku & Sonnet & Opus \\
\midrule
GSM8K & 1,319 & 90.3\% & 97.7\% & 97.9\% \\
MATH (Algebra) & 1,187 & 82.0\% & 87.1\% & 86.5\% \\
MATH (Geometry) & 479 & 68.3\% & 73.3\% & 72.0\% \\
MATH (Num.\ Theory) & 540 & 92.4\% & 92.0\% & 92.8\% \\
GPQA (MC) & 198 & 29.3\% & 28.3\% & 33.8\% \\
MBPP & 257 & 60.3\% & 72.4\% & 75.9\% \\
AIME & 90 & 57.8\% & 52.2\% & 94.4\% \\
HLE & 2,158 & 4.7\% & 6.6\% & 12.2\% \\
\bottomrule
\end{tabular}
\end{table}

\subsection{Claude 3.5/4 Model Family Results}

Table~\ref{tab:all-datasets-v35} shows results using the older Claude 3.5/4 model family for comparison.

\begin{table}[H]
\centering
\caption{Model accuracy across all datasets (Claude 3.5/4 family).}
\label{tab:all-datasets-v35}
\begin{tabular}{lrrrr}
\toprule
Dataset & $n$ & Haiku & Sonnet & Opus \\
\midrule
GSM8K & 1,319 & 93.6\% & 97.0\% & 97.0\% \\
MATH (Algebra) & 1,187 & 76.7\% & 87.0\% & 83.1\% \\
MATH (Geometry) & 479 & 47.6\% & 72.4\% & 68.9\% \\
MATH (Num.\ Theory) & 540 & 73.9\% & 92.0\% & 90.9\% \\
GPQA (MC) & 198 & 33.3\% & 31.8\% & 31.3\% \\
MBPP & 257 & 65.4\% & 69.3\% & 70.0\% \\
AIME & 90 & 3.3\% & 36.7\% & 46.7\% \\
HLE & 2,158 & 3.9\% & 5.4\% & 6.3\% \\
\bottomrule
\end{tabular}
\end{table}

\subsection{Difficulty Distribution (Claude 4.5)}

Table~\ref{tab:difficulty-v45} shows the difficulty distribution for each dataset based on Claude 4.5 model performance.

\begin{table}[H]
\centering
\caption{Difficulty distribution across datasets (Claude 4.5 family).}
\label{tab:difficulty-v45}
\begin{tabular}{lrrrrr}
\toprule
Dataset & $n$ & Easy & Medium & Hard & V.Hard \\
\midrule
GSM8K & 1,319 & 89.0\% & 8.0\% & 0.4\% & 2.7\% \\
MATH (Algebra) & 1,187 & 78.2\% & 5.1\% & 2.1\% & 14.7\% \\
MATH (Geometry) & 479 & 60.3\% & 6.3\% & 4.4\% & 29.0\% \\
MATH (Num.\ Theory) & 540 & 88.7\% & 1.7\% & 0.6\% & 9.1\% \\
GPQA (MC) & 198 & 23.2\% & 3.0\% & 4.5\% & 69.2\% \\
MBPP & 257 & 58.8\% & 10.5\% & 6.6\% & 24.1\% \\
AIME & 90 & 43.3\% & 6.7\% & 31.1\% & 18.9\% \\
HLE & 2,158 & 1.4\% & 2.3\% & 7.2\% & 89.1\% \\
\bottomrule
\end{tabular}
\end{table}

\subsection{Model Identifiers}
\label{app:model-ids}

Table~\ref{tab:model-ids} lists the Anthropic API model identifiers used in our experiments. We report results for both model families: Claude 3.5/4 (Table~\ref{tab:all-datasets-v35}) and Claude 4.5 (Table~\ref{tab:all-datasets-v45}).

\begin{table}[H]
\centering
\caption{Anthropic API model identifiers.}
\label{tab:model-ids}
\small
\begin{tabular}{lll}
\toprule
Model Name & Claude 3.5/4 (used in results) & Claude 4.5 \\
\midrule
Haiku & \texttt{claude-3-5-haiku-20241022} & \texttt{claude-haiku-4-5-20251001} \\
Sonnet & \texttt{claude-sonnet-4-20250514} & \texttt{claude-sonnet-4-5-20250929} \\
Opus & \texttt{claude-opus-4-20250514} & \texttt{claude-opus-4-5-20251101} \\
\bottomrule
\end{tabular}
\end{table}

\section{LoRA Compression Results}
\label{app:lora-results}

\subsection{Compression metric: token-based vs.\ bytes/bytes}
\label{app:metric-comparison}

The original ICML submission reported gzip's compression ratio as $\text{compressed bytes} / \text{original bytes}$, the standard metric in the classical-compression literature. We now use the same denominator $(\log_2 |V|) \times \text{num tokens}$ for every method, including gzip, to enable apples-to-apples comparison. Under this uniform denominator, gzip's average compression ratio is higher (worse) than under bytes/bytes, because the token-based denominator is smaller for the texts we evaluate. Importantly, our LoRA-adapted compressors stay at $\approx 0.09$ regardless of denominator, so the qualitative ordering is unchanged.

\subsection{Classical-compressor baselines}
\label{app:classical-compressors}

We evaluate four classical lossless compressors on 1{,}000 WildChat messages: gzip, zstd, LZMA2, and Brotli, at batch sizes 1 and 100. All four compressors fall in the 0.36 to 0.49 range under our uniform metric; gzip, zstd, and LZMA2 are within 4\% of each other, and Brotli is the strongest classical compressor (roughly 20\% better than gzip) at significantly higher compute cost. Our LoRA-adapted approach (compression ratio 0.09) dominates all classical baselines.

\subsection{Full Compression Results (All 10 LMSYS Clusters)}

Table~\ref{tab:lora_all_clusters} provides the full per-cluster breakdown of compression performance across all 10 LMSYS clusters.

\begin{table}[h]
\centering
\caption{Compression performance (compression ratio) for all 10 LMSYS clusters. Correct LoRA = adapter trained on the same cluster; Average = mean across the 9 incorrect adapters; RAG LoRA = adapter selected by the RAG router using full text or prompt only.}
\scriptsize
\setlength{\tabcolsep}{3pt}
\begin{tabular}{llccccc}
\toprule
Cluster & Topic & Baseline & Gzip & Correct LoRA & Worst Wrong & Avg Wrong \\
\midrule
0 & General Chat & 0.21 & 0.49 (0.44x) & 0.11 (2.02x) & 0.14 (1.48x) & 0.13 (1.65x) \\
1 & Creative Writing & 0.18 & 0.42 (0.42x) & 0.11 (1.67x) & 0.13 (1.42x) & 0.12 (1.47x) \\
2 & Code/Technical & 0.28 & 0.47 (0.61x) & 0.10 (2.96x) & 0.21 (1.32x) & 0.20 (1.43x) \\
3 & Academic/Education & 0.16 & 0.44 (0.36x) & 0.09 (1.71x) & 0.11 (1.43x) & 0.11 (1.49x) \\
4 & Roleplay/Fiction & 0.11 & 0.40 (0.26x) & 0.05 (2.12x) & 0.08 (1.32x) & 0.08 (1.38x) \\
5 & Business/Professional & 0.20 & 0.44 (0.45x) & 0.09 (2.20x) & 0.14 (1.47x) & 0.13 (1.56x) \\
6 & Philosophy/Ethics & 0.19 & 0.45 (0.43x) & 0.09 (2.10x) & 0.14 (1.33x) & 0.14 (1.38x) \\
7 & Science/Math & 0.15 & 0.39 (0.38x) & 0.09 (1.60x) & 0.11 (1.34x) & 0.11 (1.40x) \\
8 & Translation/Language & 0.18 & 0.43 (0.41x) & 0.09 (2.08x) & 0.14 (1.31x) & 0.13 (1.35x) \\
9 & Casual Q\&A & 0.15 & 0.44 (0.35x) & 0.10 (1.57x) & 0.13 (1.22x) & 0.12 (1.26x) \\
\midrule
\textbf{Average} & & \textbf{0.18} & \textbf{0.44 (0.41x)} & \textbf{0.09 (1.99x)} & \textbf{0.13 (1.36x)} & \textbf{0.13 (1.44x)} \\
\bottomrule
\end{tabular}
\label{tab:lora_all_clusters}
\end{table}

\subsection{RAG Routing Accuracy (LMSYS)}

\begin{table}[h]
\centering
\caption{RAG Routing Accuracy by Cluster (LMSYS)}
\begin{tabular}{llccc}
\toprule
Cluster & Topic & Full & Prompt & Total \\
\midrule
0 & General Chat & 60.0\% & 32.0\% & 50 \\
1 & Creative Writing & 18.0\% & 16.0\% & 50 \\
2 & Code/Technical & 78.0\% & 56.0\% & 50 \\
3 & Academic/Education & 64.0\% & 36.0\% & 50 \\
4 & Roleplay/Fiction & 98.0\% & 88.0\% & 50 \\
5 & Business/Professional & 90.0\% & 68.0\% & 50 \\
6 & Philosophy/Ethics & 88.0\% & 84.0\% & 50 \\
7 & Science/Math & 58.0\% & 30.0\% & 50 \\
8 & Translation/Language & 78.0\% & 48.0\% & 50 \\
9 & Casual Q\&A & 88.0\% & 92.0\% & 50 \\
\midrule
\textbf{Overall} & & \textbf{72.0\%} & \textbf{55.0\%} & \textbf{500} \\
\bottomrule
\end{tabular}
\label{tab:rag_accuracy}
\end{table}

\begin{table}[h]
\centering
\caption{Full vs Prompt RAG Routing Compression, Bits per Token (BPT) and Compression Ratio (LMSYS)}
\begin{tabular}{llcccc}
\toprule
Cluster & Topic & Full BPT & Prompt BPT & Full Comp. Ratio & Prompt Comp. Ratio \\
\midrule
0 & General Chat & 1.601 & 1.653 & 0.1069 & 0.1104 \\
1 & Creative Writing & 1.636 & 1.715 & 0.1093 & 0.1146 \\
2 & Code/Technical & 1.322 & 1.528 & 0.0884 & 0.1021 \\
3 & Academic/Education & 1.417 & 1.504 & 0.0947 & 0.1005 \\
4 & Roleplay/Fiction & 0.747 & 0.762 & 0.0499 & 0.0509 \\
5 & Business/Professional & 1.340 & 1.345 & 0.0895 & 0.0899 \\
6 & Philosophy/Ethics & 1.408 & 1.417 & 0.0940 & 0.0947 \\
7 & Science/Math & 1.444 & 1.490 & 0.0965 & 0.0996 \\
8 & Translation/Language & 1.326 & 1.550 & 0.0886 & 0.1036 \\
9 & Casual Q\&A & 1.464 & 1.496 & 0.0978 & 0.1000 \\
\midrule
\textbf{Overall} & & \textbf{1.370} & \textbf{1.446} & \textbf{0.0916} & \textbf{0.0966} \\
\bottomrule
\end{tabular}
\label{tab:full_vs_prompt_lmsys}
\end{table}

\subsection{WildChat Results}

We replicate the LoRA compression pipeline on WildChat~\citep{zhao2024wildchat1mchatgptinteraction}, a dataset of 1M ChatGPT interaction logs in the wild. The same clustering, training, and RAG routing procedure is applied to 10 WildChat clusters.

\paragraph{RAG Routing Accuracy.} Table~\ref{tab:rag_accuracy_wildchat} reports per-cluster RAG routing accuracy on WildChat. Overall accuracy is 77.0\% (full text) and 53.0\% (prompt only), which is comparable to LMSYS.

\begin{table}[h]
\centering
\caption{RAG Routing Accuracy by Cluster (WildChat)}
\begin{tabular}{llccc}
\toprule
Cluster & Topic & Full & Prompt & Total \\
\midrule
0 & General Chat & 56.0\% & 46.0\% & 50 \\
1 & Creative Writing & 98.0\% & 0.0\% & 50 \\
2 & Code/Technical & 94.0\% & 96.0\% & 50 \\
3 & Academic/Education & 56.0\% & 32.0\% & 50 \\
4 & Roleplay/Fiction & 92.0\% & 72.0\% & 50 \\
5 & Business/Professional & 52.0\% & 18.0\% & 50 \\
6 & Philosophy/Ethics & 38.0\% & 22.0\% & 50 \\
7 & Science/Math & 98.0\% & 80.0\% & 50 \\
8 & Translation/Language & 88.0\% & 70.0\% & 50 \\
9 & Casual Q\&A & 98.0\% & 94.0\% & 50 \\
\midrule
\textbf{Overall} & & \textbf{77.0\%} & \textbf{53.0\%} & \textbf{500} \\
\bottomrule
\end{tabular}
\label{tab:rag_accuracy_wildchat}
\end{table}

\paragraph{Compression Performance.}
Table~\ref{tab:rag_compression_wildchat} reports compression ratios on WildChat. The correct LoRA achieves a $1.7\times$ improvement over baseline on average (0.10 vs.\ 0.17), slightly lower than LMSYS ($2.0\times$), consistent with WildChat's greater topic diversity within clusters.

\begin{table}[h]
\centering
\caption{Compression performance (compression ratio) for all 10 Wildchat clusters. Correct LoRA = adapter trained on the same cluster; Average = mean across the 9 incorrect adapters; RAG LoRA = adapter selected by the RAG router using full text or prompt only.}
\scriptsize
\setlength{\tabcolsep}{3pt}
\begin{tabular}{llccccc}
\toprule
Cluster & Topic & Baseline & Gzip & Correct LoRA & Worst Wrong & Avg Wrong \\
\midrule
0 & General Chat & 0.20 & 0.49 (0.40x) & 0.13 (1.50x) & 0.17 (1.15x) & 0.17 (1.20x) \\
1 & Creative Writing & 0.19 & 0.38 (0.51x) & 0.04 (4.92x) & 0.16 (1.23x) & 0.13 (1.51x) \\
2 & Code/Technical & 0.14 & 0.36 (0.39x) & 0.10 (1.43x) & 0.13 (1.06x) & 0.12 (1.13x) \\
3 & Academic/Education & 0.14 & 0.51 (0.28x) & 0.10 (1.38x) & 0.13 (1.06x) & 0.12 (1.14x) \\
4 & Roleplay/Fiction & 0.12 & 0.43 (0.28x) & 0.09 (1.42x) & 0.11 (1.09x) & 0.10 (1.16x) \\
5 & Business/Professional & 0.16 & 0.52 (0.30x) & 0.12 (1.30x) & 0.15 (1.05x) & 0.14 (1.13x) \\
6 & Philosophy/Ethics & 0.17 & 0.45 (0.37x) & 0.13 (1.32x) & 0.14 (1.16x) & 0.14 (1.19x) \\
7 & Science/Math & 0.24 & 0.46 (0.52x) & 0.09 (2.75x) & 0.20 (1.18x) & 0.19 (1.24x) \\
8 & Translation/Language & 0.17 & 0.46 (0.36x) & 0.12 (1.44x) & 0.15 (1.11x) & 0.14 (1.17x) \\
9 & Casual Q\&A & 0.16 & 0.45 (0.36x) & 0.11 (1.55x) & 0.15 (1.08x) & 0.14 (1.14x) \\
\midrule
\textbf{Average} & & \textbf{0.17} & \textbf{0.45 (0.38x)} & \textbf{0.10 (1.67x)} & \textbf{0.15 (1.12x)} & \textbf{0.14 (1.20x)} \\
\bottomrule
\end{tabular}
\label{tab:rag_compression_wildchat}
\end{table}

\begin{table}[h]
\centering
\caption{Full vs Prompt RAG Routing Compression, Bits per Token (BPT) and Compression Ratio}
\scriptsize
\setlength{\tabcolsep}{3pt}
\begin{tabular}{llcccc}
\toprule
Cluster & Topic & Full BPT & Prompt BPT & Full Comp. Ratio & Prompt Comp. Ratio \\
\midrule
0 & General Chat & 2.007 & 2.035 & 0.1341 & 0.1360 \\
1 & Creative Writing & 0.590 & 0.590 & 0.0394 & 0.0394 \\
2 & Code/Technical & 1.454 & 1.471 & 0.0972 & 0.0983 \\
3 & Academic/Education & 1.575 & 1.589 & 0.1052 & 0.1062 \\
4 & Roleplay/Fiction & 1.273 & 1.298 & 0.0850 & 0.0867 \\
5 & Business/Professional & 1.836 & 1.884 & 0.1227 & 0.1259 \\
6 & Philosophy/Ethics & 1.930 & 1.968 & 0.1290 & 0.1315 \\
7 & Science/Math & 1.304 & 1.318 & 0.0871 & 0.0881 \\
8 & Translation/Language & 1.721 & 1.743 & 0.1150 & 0.1164 \\
9 & Casual Q\&A & 1.573 & 1.580 & 0.1051 & 0.1055 \\
\midrule
\textbf{Overall} & & \textbf{1.526} & \textbf{1.547} & \textbf{0.1020} & \textbf{0.1034} \\
\bottomrule
\end{tabular}
\label{tab:full_vs_prompt_wildchat}
\end{table}
\subsection{LoRA Compression on enwik9}
\label{app:enwik-lora}

To evaluate LoRA-based compression beyond conversational data, we replicate the lmsys pipeline (Section~\ref{sec:lora-compression}) on \texttt{enwik9}, a 1\,GB prefix of an English Wikipedia XML dump commonly used as a compression benchmark. Unlike the lmsys dataset, which consists of multi-turn chat conversations, enwik9 contains raw encyclopedic text spanning diverse topics.

\subsubsection{Data Preparation}

We extract individual Wikipedia articles from the XML dump, clean the text (removing markup, templates, and metadata), and filter to articles with at least 100 words, yielding 136{,}572 articles. We embed each article using the Qwen 0.6B embedding model and cluster with $K$-means into 10 clusters, then split each cluster 90/10 into train and test sets.

\subsubsection{Cluster Summary}

Table~\ref{tab:enwik-clusters} summarizes the 10 clusters. The largest cluster (002) consists almost entirely of US Census-style geographic entries with highly formulaic text, which we expect to compress particularly well with a specialized LoRA.

\begin{table}[H]
\centering
\caption{enwik9 cluster summary (10 clusters, 136{,}572 articles total).}
\label{tab:enwik-clusters}
\small
\begin{tabular}{rlrr}
\toprule
Cluster & Topic & Train & Test \\
\midrule
0 & General knowledge / social concepts & 8,464 & 941 \\
1 & Entertainment / arts / pop culture & 12,111 & 1,346 \\
2 & US geography / census data & 27,366 & 3,041 \\
3 & History and mythology & 8,108 & 901 \\
4 & Geography and nature (international) & 8,718 & 969 \\
5 & Politics and government & 7,030 & 782 \\
6 & Science and nature & 11,227 & 1,248 \\
7 & Infrastructure and miscellany & 9,026 & 1,003 \\
8 & Biographies and institutions & 10,830 & 1,204 \\
9 & Technology and computing & 7,734 & 860 \\
\midrule
\textbf{Total} & & \textbf{122,909} & \textbf{13,663} \\
\bottomrule
\end{tabular}
\end{table}

Cluster~2 is notably larger than the others (${\sim}2.5\times$ the mean), reflecting Wikipedia's large number of auto-generated US locality stubs that share near-identical structure (population, area, demographics).

\subsubsection{Experimental Setup}

We train one LoRA adapter per cluster on Mistral-7B-Instruct-v0.2, using the same hyperparameters as the lmsys pipeline (rank 16, $\alpha = 32$, 3 epochs). For LoRA selection at test time, we build a FAISS index over BGE-large-en-v1.5 embeddings of 1{,}000 training samples per cluster and route each test text via $k$-nearest-neighbor majority vote ($k=10$).

We evaluate three compression conditions using arithmetic coding (64-bit precision) on held-out test texts (minimum 200 tokens):
\begin{enumerate}
    \item \textbf{Baseline}: Mistral-7B with no LoRA adapter.
    \item \textbf{Correct LoRA} (oracle): Mistral-7B with the LoRA trained on the same cluster as the test text.
    \item \textbf{RAG-selected LoRA}: Mistral-7B with the LoRA chosen by the RAG router.
\end{enumerate}

\subsubsection{RAG Routing Accuracy}

Table~\ref{tab:enwik-rag-accuracy} reports RAG routing accuracy per cluster, evaluated on 50 test texts per cluster. Overall accuracy is 80\%, with substantial variation across clusters.

\begin{table}[H]
\centering
\caption{RAG routing accuracy on enwik9 (50 test texts per cluster, $k=10$ nearest neighbors).}
\label{tab:enwik-rag-accuracy}
\small
\begin{tabular}{rlrr}
\toprule
Cluster & Topic & Correct / Total & Accuracy \\
\midrule
0 & General knowledge / social concepts & 40 / 50 & 80.0\% \\
1 & Entertainment / arts / pop culture & 45 / 50 & 90.0\% \\
2 & US geography / census data & 50 / 50 & 100.0\% \\
3 & History and mythology & 45 / 50 & 90.0\% \\
4 & Geography and nature (international) & 39 / 50 & 78.0\% \\
5 & Politics and government & 37 / 50 & 74.0\% \\
6 & Science and nature & 43 / 50 & 86.0\% \\
7 & Infrastructure and miscellany & 31 / 50 & 62.0\% \\
8 & Biographies and institutions & 32 / 50 & 64.0\% \\
9 & Technology and computing & 38 / 50 & 76.0\% \\
\midrule
\textbf{Overall} & & \textbf{400 / 500} & \textbf{80.0\%} \\
\bottomrule
\end{tabular}
\end{table}

Cluster~2 (US geography/census) achieves perfect routing accuracy, consistent with its highly distinctive, formulaic text. Clusters~7 (infrastructure) and~8 (biographies) are the hardest to route, likely because their content overlaps with multiple other clusters.

\subsubsection{Compression Results (50 Clusters)}

We extend the enwik9 experiment to 50 clusters for finer-grained topic specialization. We split enwik9 into chunks of approximately 2,000 tokens each and cluster using k-means on \texttt{all-MiniLM-L6-v2} embeddings. Each cluster's LoRA is trained on Mistral-7B-Instruct-v0.2 with rank $r=16$, $\alpha=32$, for 3 epochs.

Table~\ref{tab:enwik9_50_compression} shows per-cluster compression performance, sorted by baseline compression ratio. The compression ratio is computed as:
\[
\text{compression ratio} = \frac{\text{compressed bits}}{\text{tokens} \times \log_2(\text{vocab\_size})}
\]
where $\log_2(32000) \approx 14.97$ bits per token represents uniform encoding over Mistral's vocabulary.

\begin{table}[H]
\centering
\caption{Enwik9 compression performance (50 clusters, sorted by baseline). Lower compression ratio is better.}
\label{tab:enwik9_50_compression}
\scriptsize
\setlength{\tabcolsep}{3pt}
\begin{tabular}{rrcccr}
\toprule
Cluster & Samples & Baseline & RAG LoRA & Correct LoRA & RAG Acc \\
\midrule
27 & 524 & 0.060 & 0.044 & 0.044 & 96.0\% \\
36 & 1,327 & 0.064 & 0.046 & 0.045 & 70.4\% \\
0 & 818 & 0.075 & 0.053 & 0.053 & 71.4\% \\
11 & 469 & 0.077 & 0.055 & 0.055 & 90.4\% \\
2 & 346 & 0.103 & 0.078 & 0.078 & 51.7\% \\
44 & 121 & 0.138 & 0.110 & 0.110 & 87.6\% \\
13 & 299 & 0.147 & 0.116 & 0.117 & 78.3\% \\
9 & 330 & 0.164 & 0.141 & 0.140 & 51.2\% \\
19 & 246 & 0.166 & 0.111 & 0.111 & 87.4\% \\
41 & 233 & 0.189 & 0.160 & 0.160 & 95.7\% \\
16 & 307 & 0.193 & 0.159 & 0.159 & 81.4\% \\
10 & 257 & 0.194 & 0.162 & 0.162 & 80.2\% \\
34 & 188 & 0.194 & 0.161 & 0.160 & 67.0\% \\
38 & 122 & 0.195 & 0.169 & 0.169 & 78.7\% \\
39 & 269 & 0.197 & 0.174 & 0.174 & 65.8\% \\
22 & 118 & 0.197 & 0.162 & 0.163 & 92.4\% \\
4 & 158 & 0.199 & 0.159 & 0.160 & 63.9\% \\
6 & 176 & 0.200 & 0.165 & 0.165 & 72.7\% \\
32 & 150 & 0.205 & 0.167 & 0.167 & 88.7\% \\
26 & 250 & 0.208 & 0.182 & 0.182 & 74.4\% \\
7 & 228 & 0.208 & 0.178 & 0.178 & 90.4\% \\
15 & 355 & 0.210 & 0.183 & 0.183 & 73.2\% \\
24 & 343 & 0.210 & 0.183 & 0.182 & 64.7\% \\
1 & 373 & 0.212 & 0.184 & 0.184 & 60.1\% \\
31 & 111 & 0.214 & 0.183 & 0.183 & 81.1\% \\
46 & 179 & 0.214 & 0.182 & 0.183 & 73.7\% \\
23 & 340 & 0.215 & 0.190 & 0.190 & 45.0\% \\
49 & 100 & 0.218 & 0.181 & 0.181 & 90.0\% \\
48 & 214 & 0.219 & 0.191 & 0.191 & 89.7\% \\
42 & 143 & 0.219 & 0.194 & 0.194 & 83.2\% \\
25 & 257 & 0.220 & 0.188 & 0.188 & 77.0\% \\
3 & 169 & 0.220 & 0.190 & 0.190 & 86.4\% \\
14 & 246 & 0.221 & 0.189 & 0.189 & 80.5\% \\
28 & 219 & 0.222 & 0.198 & 0.198 & 79.9\% \\
40 & 198 & 0.225 & 0.194 & 0.194 & 79.3\% \\
33 & 303 & 0.225 & 0.196 & 0.196 & 55.8\% \\
30 & 166 & 0.227 & 0.196 & 0.196 & 90.4\% \\
29 & 205 & 0.227 & 0.173 & 0.173 & 89.3\% \\
21 & 268 & 0.228 & 0.190 & 0.189 & 40.7\% \\
45 & 220 & 0.228 & 0.203 & 0.203 & 64.5\% \\
35 & 142 & 0.228 & 0.202 & 0.202 & 69.0\% \\
43 & 229 & 0.230 & 0.199 & 0.199 & 76.4\% \\
8 & 107 & 0.235 & 0.209 & 0.209 & 90.7\% \\
47 & 192 & 0.240 & 0.209 & 0.208 & 83.9\% \\
17 & 245 & 0.244 & 0.209 & 0.209 & 77.1\% \\
18 & 182 & 0.247 & 0.215 & 0.215 & 68.7\% \\
20 & 312 & 0.248 & 0.220 & 0.219 & 63.1\% \\
12 & 285 & 0.248 & 0.219 & 0.219 & 66.7\% \\
5 & 158 & 0.248 & 0.224 & 0.224 & 68.4\% \\
37 & 245 & 0.253 & 0.224 & 0.224 & 64.1\% \\
\midrule
\textbf{Total} & \textbf{13,442} & \textbf{0.176} & \textbf{0.148} & \textbf{0.148} & \textbf{73.6\%} \\
\bottomrule
\end{tabular}
\end{table}

\paragraph{Key findings.}
\begin{itemize}
    \item \textbf{16\% compression improvement}: LoRA adaptation reduces the average compression ratio from 0.176 (baseline) to 0.148 (RAG LoRA), a 16\% reduction in compressed size.
    \item \textbf{High variance across clusters}: Compression ratios range from 0.044 (cluster 27) to 0.224 (clusters 5, 37), reflecting Wikipedia's content diversity. The best-compressing clusters likely contain highly structured content (tables, lists, formulaic entries).
    \item \textbf{RAG LoRA $\approx$ Correct LoRA}: Despite 26.4\% routing errors, RAG LoRA achieves nearly identical compression to the oracle (0.148 vs 0.148). This indicates that misrouted samples are typically sent to clusters with similar compression characteristics.
    \item \textbf{Routing accuracy varies}: RAG accuracy ranges from 40.7\% (cluster 21) to 96.0\% (cluster 27), with an overall accuracy of 73.6\%.
\end{itemize}

\paragraph{Notes on baseline compression.}
Our baseline compression ratio of 0.176 is higher (worse) than ratios reported in prior work on neural text compression \cite{delétang2024languagemodelingcompression}, which report values closer to 0.1 for the same model size. This discrepancy is likely due to several factors: (1) differences in base models (we use Mistral-7B-Instruct rather than models specifically optimized for compression), and (2) differences in experimental setup (e.g., how text is chunked, minimum token thresholds, evaluation protocol). The key takeaway is not the absolute compression ratio, but rather the \emph{relative improvement} from LoRA adaptation: a consistent 16\% reduction in compressed size across all clusters.

\paragraph{Comparison to conversational data.}
Interestingly, the conversational datasets (LMSYS and WildChat) achieve substantially better compression ratios than enwik9. LMSYS achieves a baseline ratio of 0.09 and WildChat achieves 0.10 (Tables~\ref{tab:full_vs_prompt_lmsys} and~\ref{tab:full_vs_prompt_wildchat}), compared to 0.176 for enwik9. This suggests that LLM-generated conversational text is ${\sim}2\times$ more compressible than raw Wikipedia text. This is intuitive: conversational text produced by LLMs likely has more predictable patterns and less information density than encyclopedic content, which contains diverse factual information, proper nouns, dates, and technical terminology.

\section{``Just Ask'' Compression}
\label{app:best-of-n}

\subsection{Quality Preservation in Lossy Compression}
\label{app:quality-preservation}

When using Summarization (Section~\ref{sec:just-ask}), a natural concern is whether the succinct rewrite preserves the quality of the original reasoning, or merely preserves enough surface-level information to guess the answer.

\paragraph{Answer-Masking Check.}
We implement a simple quality check. Before requesting the succinct rewrite, we mask the final answer in the verbose solution, replacing $\boxed{A}$ with $\boxed{???}$. The model then rewrites the masked solution, and must re-derive the correct answer from its own succinct version. If it produces an incorrect answer, we consider the compression too lossy.

\paragraph{Limitations.}
This check is imperfect. The succinct version might ``leak'' information about the answer through numerical patterns or structural cues (e.g., ``the answer is a two-digit prime'') without actually preserving the full reasoning chain. A human reader might find the succinct version harder to follow or less informative than the verbose original, even if a language model can still extract the correct answer.

More rigorous assessment would require human evaluation of reasoning quality, or measuring performance on downstream tasks that depend on understanding the solution (not just knowing the final answer). We leave such evaluation to future work. Other work, such as \citet{wang2020askingansweringquestionsevaluate, dua2025clinicallygroundedagentbasedreport} propose alternative mechanisms for evaluating that a summary retains the core information from the body, largely premised on using an LLM to generate questions about the original text, and then prompting another LLM to answer the questions based on the original and answer the questions based on the summarized text, and comparing alignment.

\subsection{Effect of Compression-Based Selection on Accuracy}
  \label{sec:accuracy-vs-compression}
  A natural concern with best-of-$N$ selection by compressibility is whether choosing the most compressible candidate degrades answer quality. To
  investigate this, we compare the accuracy of two selection strategies across 90 AIME problems: (1) \textbf{random selection}, which picks uniformly among
   the $N$ candidates (equivalently, the expected accuracy of a single draw), and (2) \textbf{best-compression selection}, which picks the candidate with
  the lowest compression ratio.
  \begin{figure}[H]
      \centering
      \includegraphics[width=.65\linewidth]{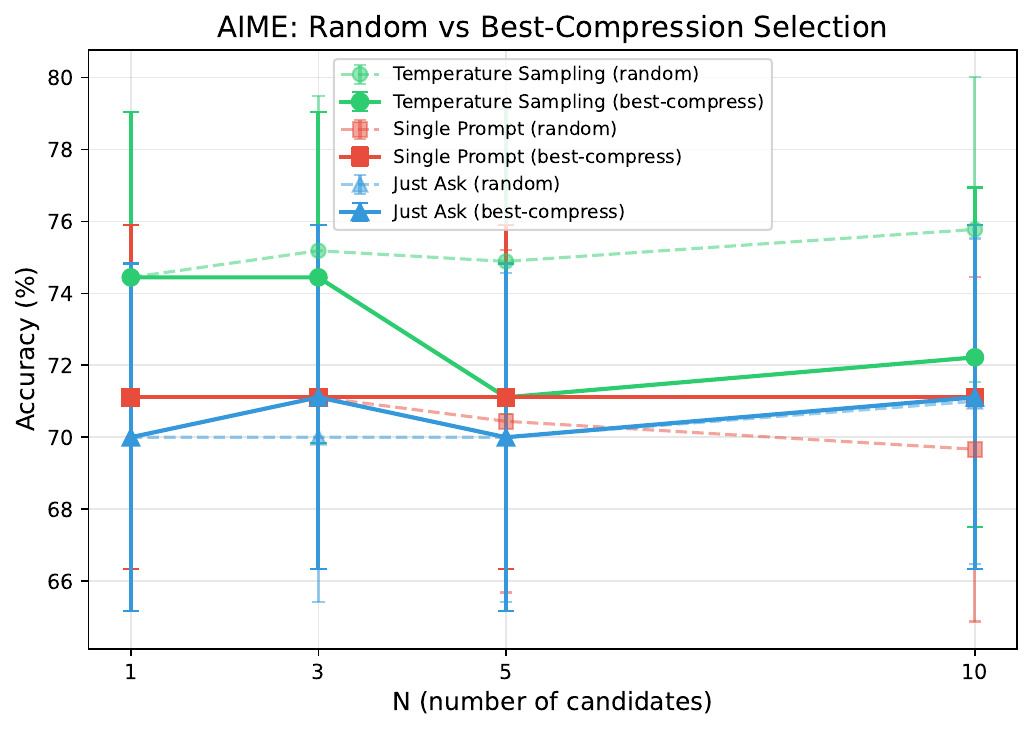}
      \caption{Accuracy of random selection (dashed) versus best-compression selection (solid) on 90 AIME problems using Opus. Random selection reports the expected
  accuracy of picking uniformly among $N$ candidates. Across all three generation strategies, selecting the most compressible candidate yields accuracy within a few percentage points of random selection.}
      \label{fig:accuracy-versus-compression}
  \end{figure}
  Figure~\ref{fig:accuracy-versus-compression} shows the results for all three generation approaches using Opus. For all three strategies, compression-based selection yields accuracy within a few percentage points of random selection across all values of $N$. For \textit{temperature sampling}, random selection benefits slightly from oracle best-of-$N$ at large $N$ (rising from ${\sim}70\%$ to ${\sim}76\%$ at $N{=}10$), while compression-based selection remains relatively flat around $72$--$74\%$. For \textit{single prompt} and \textit{just ask}, both selection strategies yield nearly identical accuracy (${\sim}70$--$71\%$) across all $N$.

  These results indicate that compression-based selection does not sacrifice accuracy. For the weaker Haiku model, compression-based selection can even \emph{improve} accuracy (see \S\ref{sec:haiku-best-of-n}). %

\subsection{Relative Compression Ratios}
\label{sec:relative-compression}

\begin{figure}[H]
    \centering
    \includegraphics[width=.99\linewidth]{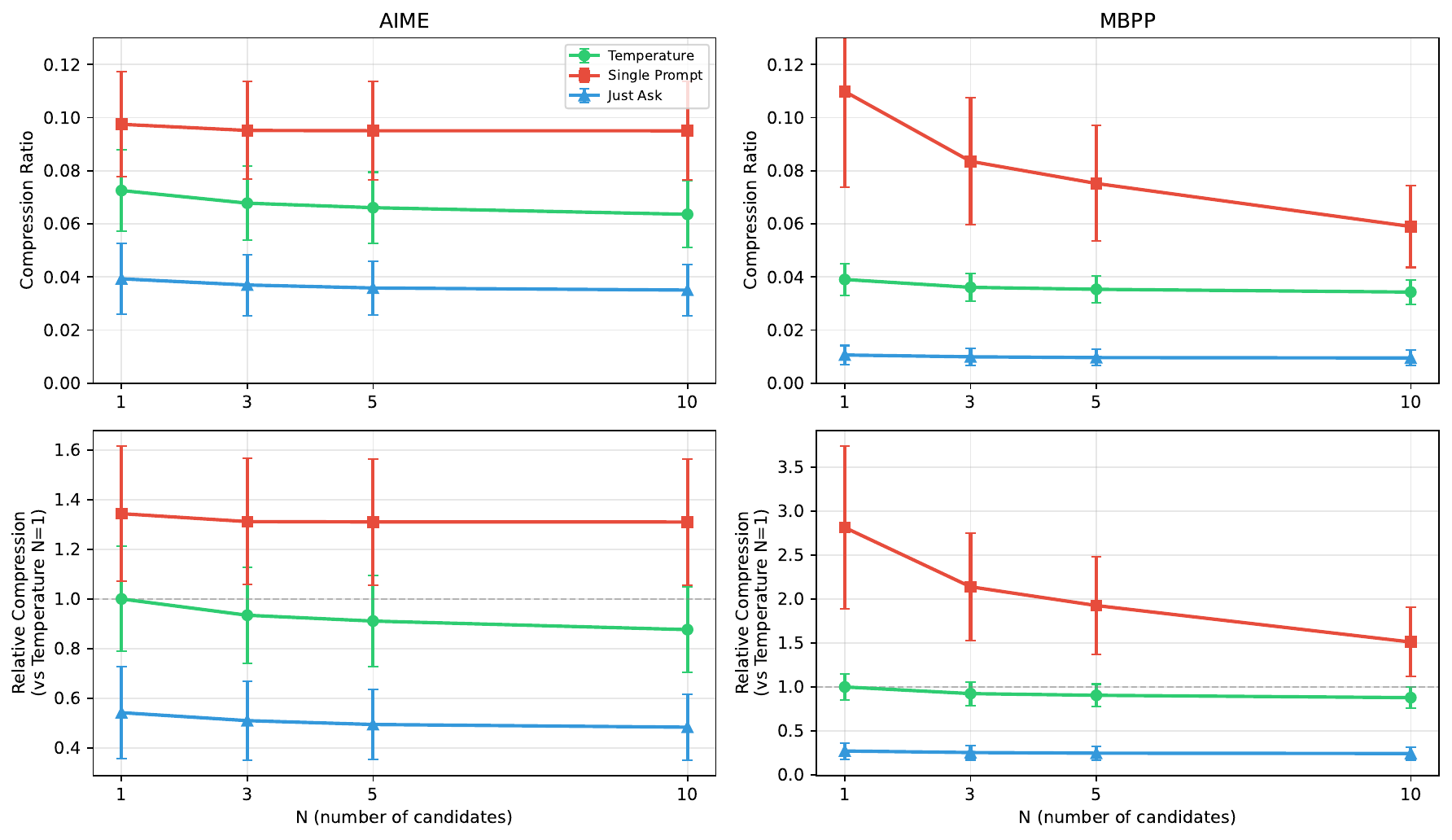}
    \caption{Absolute compression ratio (top) and relative compression ratio normalized to Temperature Sampling at $N{=}1$ (bottom) on AIME (left) and MBPP (right) using Opus. The dashed gray line at 1.0 indicates the baseline: values below 1.0 represent improved compression, while values above 1.0 indicate worse compression than the baseline. Lower is better.}
    \label{fig:opus-relative-compression}
\end{figure}

\begin{figure}[H]
    \centering
    \includegraphics[width=.99\linewidth]{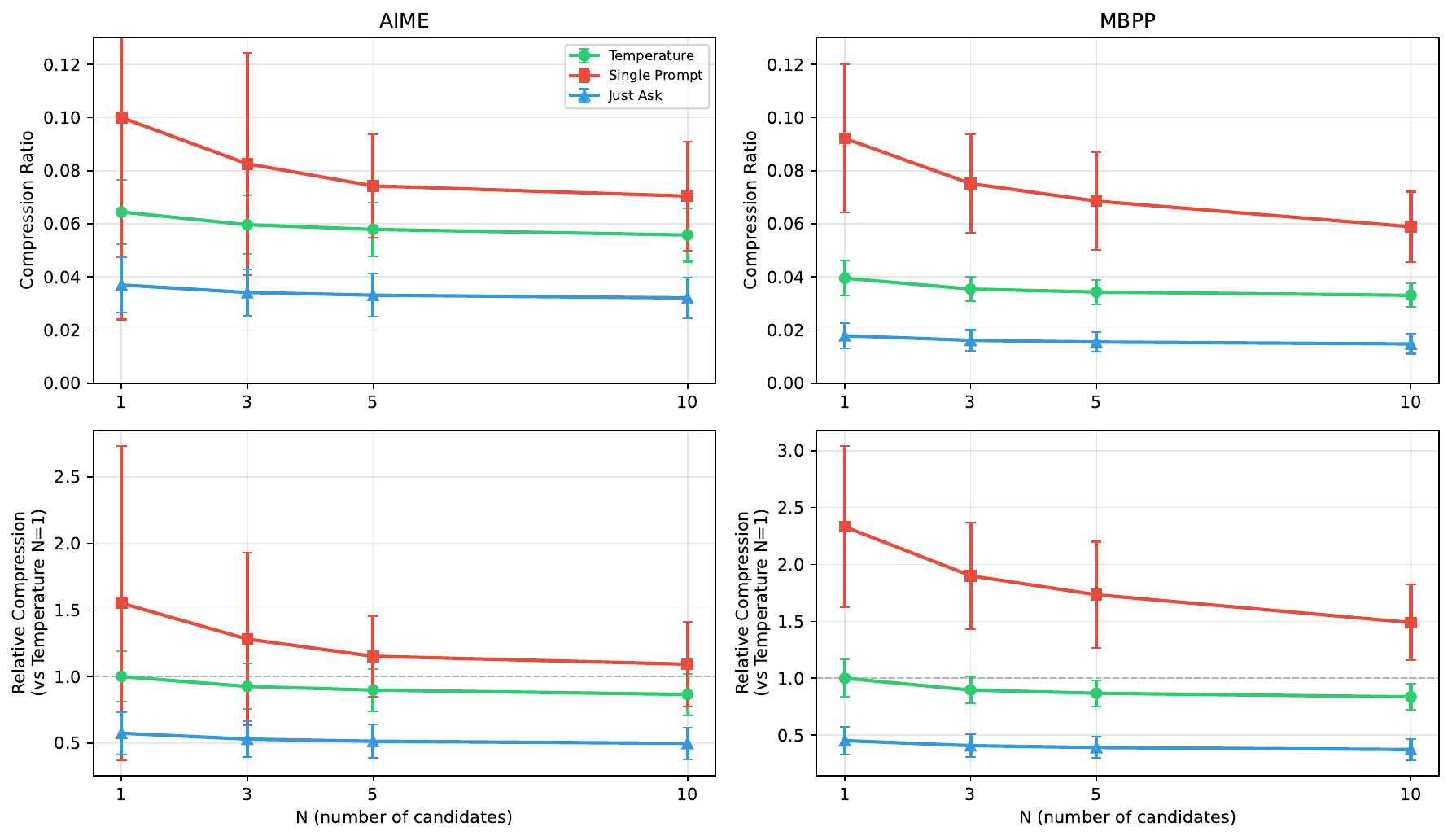}
    \caption{Absolute compression ratio (top) and relative compression ratio normalized to Temperature Sampling at $N{=}1$ (bottom) on AIME (left) and MBPP (right) using Haiku. The dashed gray line at 1.0 indicates the baseline: values below 1.0 represent improved compression, while values above 1.0 indicate worse compression than the baseline. Lower is better.}
    \label{fig:haiku-relative-compression}
\end{figure}

Figures~\ref{fig:opus-relative-compression} and~\ref{fig:haiku-relative-compression} show both the absolute compression ratio (top row) and the relative compression ratio normalized to Temperature Sampling at $N{=}1$ (bottom row) for Opus and Haiku respectively. A relative compression ratio of 1.0 (dashed gray line) corresponds to no improvement over the baseline; lower values indicate better compression. On AIME, Just Ask achieves a relative compression ratio of ${\sim}0.5$ for both models, meaning it compresses to roughly half the size of the temperature sampling baseline. On MBPP, the gains are even more pronounced: Just Ask reaches ${\sim}0.27$ for Opus and ${\sim}0.45$ for Haiku.

\subsection{Haiku Results}
\label{sec:haiku-best-of-n}

\begin{figure}[H]
    \centering
    \includegraphics[width=.99\linewidth]{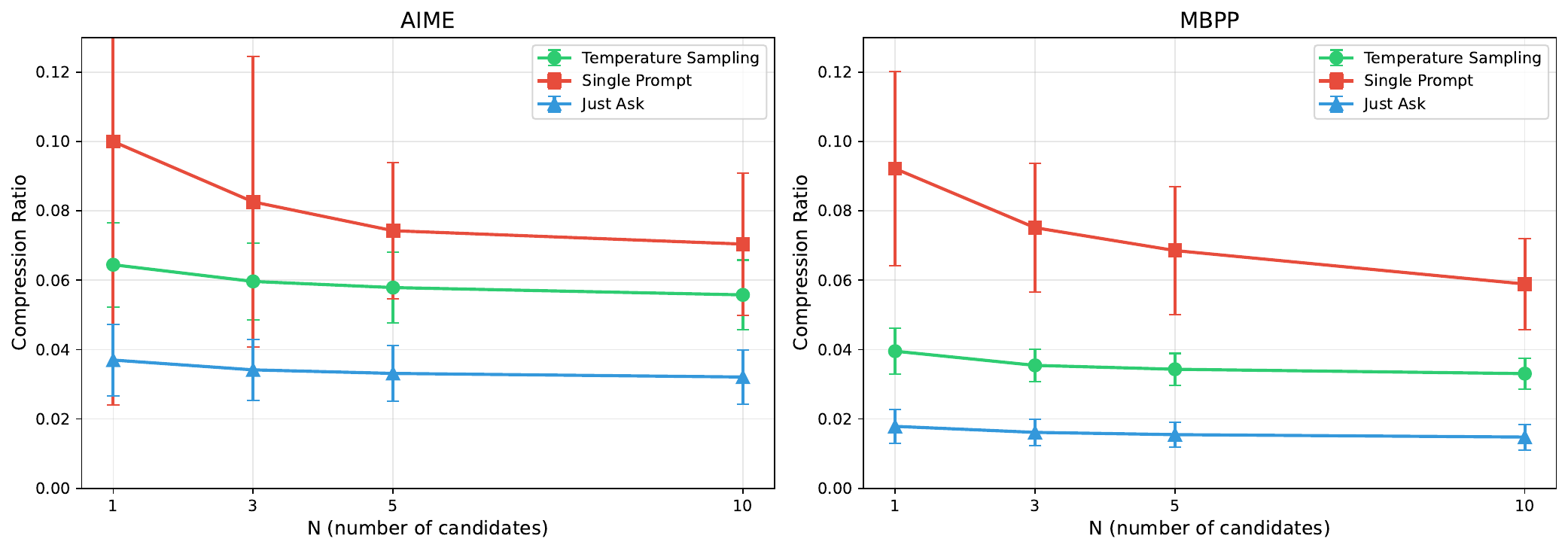}
    \caption{Compression ratio vs.\ number of candidates $N$ on AIME (left) and MBPP (right) problems, using Haiku as both generator and compressor. The qualitative pattern matches Opus (Figure~\ref{fig:compression-comparison}): Just Ask achieves the best compression, followed by Temperature Sampling, with Single Prompt performing worst.}
    \label{fig:haiku-compression-comparison}
\end{figure}

Figure~\ref{fig:haiku-compression-comparison} shows the compression comparison using Haiku as both generator and compressor. The qualitative pattern matches Opus: summarization achieves the best compression, followed by temperature sampling, with single-prompt generation performing worst. On AIME, temperature sampling achieves compression ratios from 0.065 at $N=1$ to 0.055 at $N=10$, while summarization achieves 0.036 to 0.032.

\begin{figure}[H]
    \centering
    \includegraphics[width=.65\linewidth]{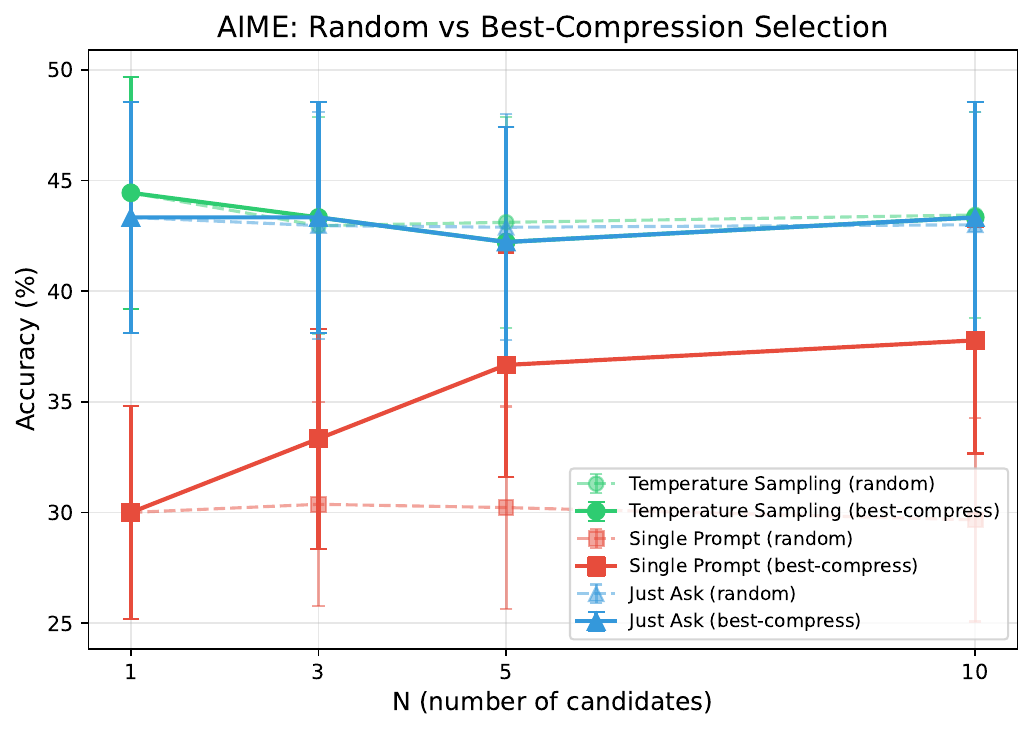}
    \caption{Accuracy of random selection (dashed) versus best-compression selection (solid) on 90 AIME problems using Haiku. For single prompt generation, compression-based selection improves accuracy by up to 8 percentage points at $N{=}10$, suggesting that correct solutions follow more predictable reasoning patterns.}
    \label{fig:haiku-accuracy-versus-compression}
\end{figure}

Figure~\ref{fig:haiku-accuracy-versus-compression} shows the accuracy trade-off for Haiku. Notably, for \textit{single prompt}, compression-based selection \emph{improves} accuracy from 30\% (random) to 38\% at $N{=}10$, an 8 percentage point gain. This can be explained by examining the compressibility of correct versus incorrect solutions: correct answers are on average more compressible than incorrect ones ($-0.65\%$ and $-1.71\%$ difference in compression ratio for temperature sampling and single prompt, respectively), suggesting that correct solutions follow more predictable reasoning patterns. For \textit{just ask}, the correlation is negligible ($+0.23\%$), consistent with the flat accuracy curves.

\subsection{Knowledge Density and Compressibility}
\label{sec:knowledge-density}

\begin{figure}[H]
    \centering
    \includegraphics[width=1.\linewidth]{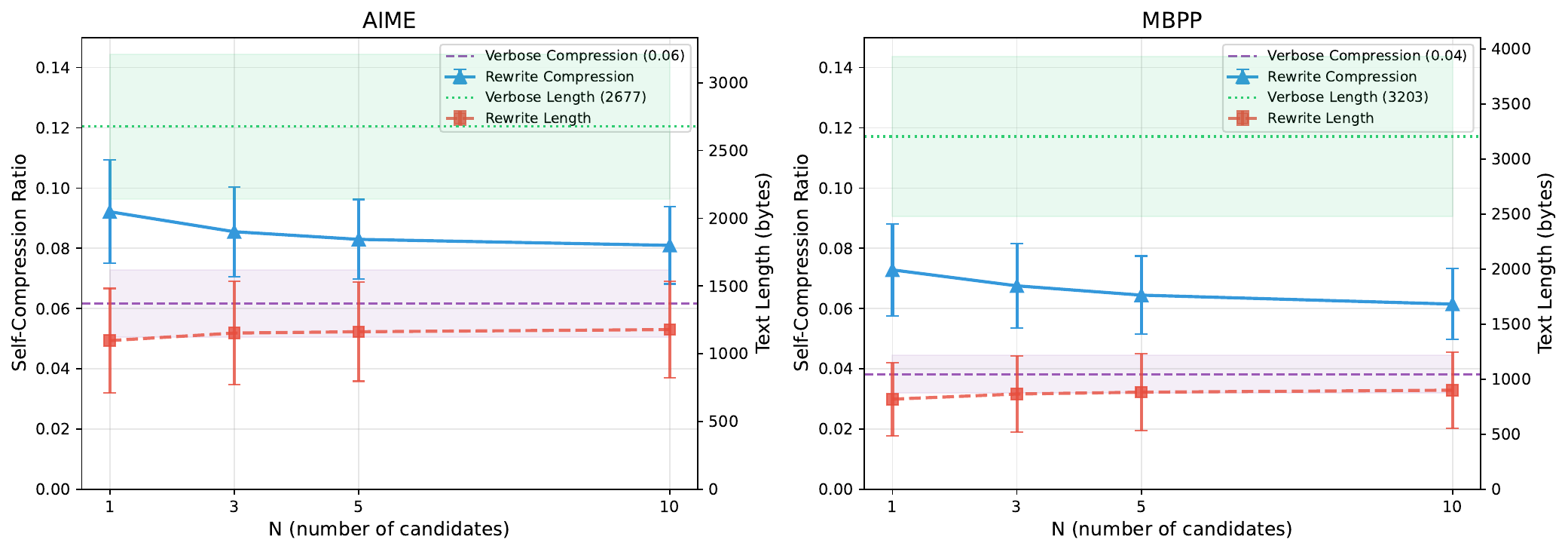}
    \caption{Compression ratio (bits-per-character) for verbose original solutions versus succinct rewrites on AIME problems using Haiku. Summarized text has higher bits-per-character (less redundancy per character) but is substantially shorter, yielding lower total compressed size.}
    \label{fig:just-ask-self-compression}
\end{figure}

An interesting phenomenon emerges when we examine what happens to text after summarization: the rewritten text is shorter but \emph{less compressible per character}. Figure~\ref{fig:just-ask-self-compression} illustrates this on AIME problems using Haiku. The verbose original solutions contain predictable boilerplate (``Step 1:'', ``Therefore,'', ``Let us consider'') that an LLM easily predicts, resulting in low bits-per-character. When we ask the model to rewrite the solution as succinctly as possible, the boilerplate is stripped away and what remains is denser, less predictable text: variable names, key equations, and compressed reasoning steps that the LLM used for compression cannot as easily anticipate.

However, the reduction in length is larger than the increase in per-character complexity. A typical verbose solution of ${\sim}2{,}100$ characters at ${\sim}5.3\%$ compression ratio becomes a ${\sim}590$-character rewrite at ${\sim}9.5\%$ compression ratio. Despite the nearly doubled bits-per-character, the total compressed size drops by roughly $40\%$ because the text is $3.6\times$ shorter. In other words, summarization increases the density of information of the text while still reducing the total information cost.

\section{Negative Results \& Ablations}
\label{app:negative-results}

In this section, we document approaches that did not yield the expected improvements, as well as ablation studies that provide insight into the mechanisms underlying our methods.

\subsection{Judge Design}
\label{app:judge-design}

The iterative QA protocol requires a judge to decide whether the SLM's updated answer is satisfactory (and should be accepted) or whether to continue asking questions. Two natural options arise:

\begin{enumerate}
    \item \textbf{Objective (standalone) judging.} The judge evaluates the SLM's answer on its own merits, scoring solution quality on a 1--10 scale. This does not require a reference solution, but relies on the judge's ability to assess correctness independently.
    \item \textbf{Comparison judging.} The judge compares the SLM's answer against the LLM's own solution, scoring how similar or aligned they are. This gives the judge a concrete reference, but the LLM's own answer may itself be wrong on hard problems.
\end{enumerate}

We evaluate both options below, along with ablations on the quality threshold and gold-answer access.

\subsubsection{Quality-Thresholding Judge in QA Compression}
\label{app:judge-ablation}

A natural idea for improving the QA protocol is to add a \emph{quality-thresholding judge}: after each round of questions, an LLM evaluates the SLM's updated answer on a 1--10 scale. If the score meets a threshold ($\geq 7$), the protocol accepts the current answer and early-stops; otherwise, it continues asking questions in hopes of further improvement.

We test this by comparing the standard QA protocol against an iterative variant that adds an objective judge. Both protocols use the same model configurations, \BLC{} (Haiku$\to$Haiku$\to$Haiku) and QA (Haiku$\to$Opus$\to$Haiku), evaluated on Claude 4.5 models, with a budget of 10 binary yes/no questions each. The protocols differ in two ways:
\begin{enumerate}
    \item \textbf{Quality thresholding:} The iterative variant adds a judge that evaluates the SLM's answer on a 1--10 scale (mathematical soundness, calculation correctness, reasoning clarity) after each batch of 5 questions. If the score $\geq 7$, the protocol accepts the current answer and early-stops.
    \item \textbf{Gold answer access:} In the standard protocol, the LLM answering questions is given the gold answer as reference. In the iterative variant, the LLM generates its own solution first, then uses it as reference for answering questions.
\end{enumerate}
These two changes are confounded: the iterative variant both removes gold answer access and adds judging. However, since QA compression in practice cannot rely on gold answers, the iterative protocol represents the more realistic setting.

Table~\ref{tab:judge-ablation} shows that the iterative protocol consistently \emph{reduces} recovery rates compared to the standard protocol. For the QA configuration (Haiku$\to$Opus$\to$Haiku), recovery drops on all 8 datasets. The ``Standard'' column reproduces the results from Table~\ref{tab:qa-accuracy}.\footnote{Sample sizes differ slightly between the two protocols: the standard protocol evaluates all non-easy problems, while the iterative variant caps at 150 per dataset.}

\begin{table}[H]
\centering
\caption{Recovery rates (\%) for the standard QA protocol (Table~\ref{tab:qa-accuracy}) vs.\ the iterative variant. ``Standard'': 10 questions in one batch, LLM given gold answer, no judge. ``Iterative'': 2 batches of 5 questions, LLM generates own solution, objective judge with threshold $\geq 7/10$. $\Delta$: change in recovery rate. Negative values indicate the iterative protocol performs worse.}
\label{tab:judge-ablation}
\footnotesize
\setlength{\tabcolsep}{4pt}
\begin{tabular}{l rr rr rr}
\toprule
 & \multicolumn{3}{c}{\BLC{} (h$\to$h$\to$h)} & \multicolumn{3}{c}{QA (h$\to$o$\to$h)} \\
\cmidrule(lr){2-4} \cmidrule(lr){5-7}
Dataset & Std & Iter & $\Delta$ & Std & Iter & $\Delta$ \\
\midrule
GSM8K            & 23.4 & 26.9 & +3.5  & 31.0 & 27.6 & $-$3.4 \\
MATH (Algebra)   & 40.5 & 36.0 & $-$4.5 & 38.2 & 36.0 & $-$2.2 \\
MATH (Geometry)  & 38.4 & 33.3 & $-$5.1 & 38.9 & 31.3 & $-$7.6 \\
MATH (Num.\ Thy) & 36.1 & 44.3 & +8.2  & 44.3 & 42.6 & $-$1.7 \\
GPQA (MC)        & 13.2 &  8.0 & $-$5.2 & 17.1 &  8.7 & $-$8.4 \\
MBPP             & 15.1 & 14.2 & $-$0.9 & 17.9 & 17.0 & $-$0.9 \\
AIME             & 19.6 &  9.8 & $-$9.8 & 17.6 & 11.8 & $-$5.8 \\
HLE              &  5.2 &  2.7 & $-$2.5 &  7.8 &  1.3 & $-$6.5 \\
\midrule
\textbf{Mean}    & 23.9 & 21.9 & $-$2.0 & 26.6 & 22.0 & $-$4.6 \\
\bottomrule
\end{tabular}
\end{table}

\paragraph{Analysis of judge behavior.}
To understand the role of the quality-thresholding judge specifically, Table~\ref{tab:judge-early-stop} examines its behavior in the iterative protocol. Note that ``early stop'' means the judge was \emph{satisfied} (score $\geq 7$) and terminated the protocol, accepting the current answer.

\begin{table}[H]
\centering
\caption{Judge behavior statistics (objective judge, QA config). Early-stop \%: fraction of problems where the judge scored $\geq 7$ and accepted the answer before all 10 questions were used. Avg.\ Score: mean judge rating (1--10). Score $<7$: fraction of individual evaluations below the acceptance threshold.}
\label{tab:judge-early-stop}
\footnotesize
\setlength{\tabcolsep}{4pt}
\begin{tabular}{l rr rr}
\toprule
Dataset & Early Stop (\%) & Avg Qs Used & Avg Score & Score $<7$ (\%) \\
\midrule
GSM8K            & 89.0 & 5.9 & 8.5 & 19.0 \\
MATH (Algebra)   & 98.7 & 5.1 & 9.7 &  3.0 \\
MATH (Geometry)  & 78.7 & 6.3 & 7.2 & 36.6 \\
MATH (Num.\ Thy) & 90.2 & 5.7 & 8.4 & 20.0 \\
GPQA (MC)        & 90.0 & 6.1 & 7.5 & 25.5 \\
MBPP             & 92.5 & 5.5 & 8.3 & 12.3 \\
AIME             & 19.6 & 9.4 & 3.0 & 91.8 \\
HLE              & 50.7 & 7.7 & 5.4 & 64.7 \\
\bottomrule
\end{tabular}
\end{table}

The data reveals two distinct failure modes:

\begin{enumerate}
    \item \textbf{Easy-to-judge datasets} (MATH Algebra: avg.\ score 9.7, 98.7\% early stopped). The judge gives high scores and accepts the answer after just one round of 5 questions, before the SLM has received enough guidance from the Q\&A exchange. The no-judge protocol would have continued for a full 10 questions, giving the SLM more information to work with.

    \item \textbf{Hard-to-judge datasets} (AIME: avg.\ score 3.0, 80.4\% \emph{not} early stopped; HLE: avg.\ score 5.4, 49.3\% not early stopped). The judge scores are persistently low, so the protocol runs all 10 questions but the final answer is still scored poorly. On AIME, problems that were \emph{not} early stopped show severe regression: 7 out of 41 were initially correct, but only 1 remained correct after the Q\&A exchange, a regression rate of 86\%.
\end{enumerate}

In neither regime does the judge provide a net benefit. On easy problems, it terminates the protocol prematurely; on hard problems, it fails to prevent regression. The judge lacks the ability to distinguish between ``the answer improved but is still imperfect'' (where it should continue) and ``the Q\&A guidance is making things worse'' (where it should revert to the original answer).

Beyond the judge, the removal of gold answer access likely also contributes to the drop in recovery. When the LLM must generate its own solution to answer questions, its yes/no responses may be less reliable, particularly on hard problems where the LLM itself struggles. To partially disentangle these factors, we run two additional ablations:
\begin{enumerate}
    \item \textbf{Higher threshold ($\geq 9$):} Raising the quality threshold from 7 to 9 should reduce premature early stopping, since the judge will accept fewer answers. This isolates whether the threshold level is the primary issue.
    \item \textbf{Gold-answer judge:} The judge is given the gold answer for evaluation, while the LLM still answers questions using its own solution. This isolates the judge's evaluation quality from the LLM's question-answering quality.
\end{enumerate}

\paragraph{Higher threshold ($\geq 9$).}
Table~\ref{tab:threshold-ablation} compares recovery rates under the default threshold ($\geq 7$) and a stricter threshold ($\geq 9$). The stricter threshold reduces premature early stopping: for example, on GPQA the early-stop rate drops from 90\% to 61\%, and on HLE from 51\% to 7\%. Despite forcing the protocol to ask more questions, recovery rates do not improve. 

\begin{table}[H]
\centering
\caption{Recovery rates (\%) for the iterative protocol with quality threshold $\geq 7$ (default) vs.\ $\geq 9$. Both use the objective judge, no gold answer. $\Delta$: change in recovery rate.}
\label{tab:threshold-ablation}
\footnotesize
\setlength{\tabcolsep}{4pt}
\begin{tabular}{l rr rr rr}
\toprule
 & \multicolumn{3}{c}{\BLC{} (h$\to$h$\to$h)} & \multicolumn{3}{c}{QA (h$\to$o$\to$h)} \\
\cmidrule(lr){2-4} \cmidrule(lr){5-7}
Dataset & $t{=}7$ & $t{=}9$ & $\Delta$ & $t{=}7$ & $t{=}9$ & $\Delta$ \\
\midrule
GSM8K            & 18.8 & 14.8 & $-$3.9 & 19.5 & 20.3 & $+$0.8 \\
MATH (Algebra)   & 27.3 & 26.6 & $-$0.8 & 27.3 & 25.8 & $-$1.6 \\
MATH (Geometry)  & 25.8 & 19.4 & $-$6.5 & 23.4 & 22.6 & $-$0.8 \\
MATH (Num.\ Thy) & 24.4 & 19.5 & $-$4.9 & 19.5 & 19.5 & $+$0.0 \\
GPQA (MC)        &  5.1 &  5.1 & $+$0.0 &  6.5 &  6.5 & $+$0.0 \\
MBPP             & 12.7 & 14.7 & $+$2.0 & 16.7 & 18.6 & $+$2.0 \\
AIME             &  5.3 &  2.6 & $-$2.6 &  5.3 &  2.6 & $-$2.6 \\
HLE              &  2.8 &  0.7 & $-$2.1 &  1.4 &  1.4 & $+$0.0 \\
\midrule
\textbf{Mean}    & 15.3 & 12.9 & $-$2.3 & 15.0 & 14.7 & $-$0.3 \\
\bottomrule
\end{tabular}
\end{table}

\paragraph{Gold-answer judge.}
Table~\ref{tab:goldjudge-ablation} compares the default iterative protocol (judge evaluates without the gold answer) to a variant where the judge is given the gold answer for evaluation. The LLM still answers questions using its own solution in both cases, so this isolates the judge's evaluation quality from the LLM's question-answering quality. 

\begin{table}[H]
\centering
\caption{Recovery rates (\%) for the iterative protocol without vs.\ with gold answer access for the judge. Both use the objective judge with threshold $\geq 7$. The LLM answers questions using its own solution in both cases. $\Delta$: change in recovery rate.}
\label{tab:goldjudge-ablation}
\footnotesize
\setlength{\tabcolsep}{4pt}
\begin{tabular}{l rr rr rr}
\toprule
 & \multicolumn{3}{c}{\BLC{} (h$\to$h$\to$h)} & \multicolumn{3}{c}{QA (h$\to$o$\to$h)} \\
\cmidrule(lr){2-4} \cmidrule(lr){5-7}
Dataset & Base & Gold & $\Delta$ & Base & Gold & $\Delta$ \\
\midrule
GSM8K            & 18.8 & 20.3 & $+$1.6 & 19.5 & 19.5 & $+$0.0 \\
MATH (Algebra)   & 27.3 & 26.6 & $-$0.8 & 27.3 & 27.3 & $+$0.0 \\
MATH (Geometry)  & 25.8 & 19.4 & $-$6.5 & 23.4 & 25.8 & $+$2.4 \\
MATH (Num.\ Thy) & 24.4 & 24.4 & $+$0.0 & 19.5 & 19.5 & $+$0.0 \\
GPQA (MC)        &  5.1 &  5.8 & $+$0.7 &  6.5 &  8.0 & $+$1.4 \\
MBPP             & 12.7 & 14.7 & $+$2.0 & 16.7 & 17.6 & $+$1.0 \\
AIME             &  5.3 &  7.9 & $+$2.6 &  5.3 &  5.3 & $+$0.0 \\
HLE              &  2.8 &  1.4 & $-$1.4 &  1.4 &  1.4 & $+$0.0 \\
\midrule
\textbf{Mean}    & 15.3 & 15.1 & $-$0.2 & 15.0 & 15.6 & $+$0.6 \\
\bottomrule
\end{tabular}
\end{table}

\paragraph{Comparison-mode judge.}
We also test a \emph{comparison-mode} judge, which rates the SLM's answer relative to the LLM's own solution rather than evaluating it in isolation. Table~\ref{tab:judge-mode-comparison} shows recovery rates across all three protocols: the standard (no-judge) protocol from Table~\ref{tab:qa-accuracy}, the iterative protocol with an objective judge, and the iterative protocol with a comparison judge. Both iterative variants use threshold $\geq 7$, and the LLM answers questions using its own solution (no gold answer).

\begin{table}[H]
\centering
\caption{Recovery rates (\%) across three protocols. ``Standard'': no judge, LLM given gold answer, 10 questions in one batch. ``Objective'': standalone quality judge, threshold $\geq 7$, no gold answer. ``Comparison'': judge compares SLM answer to LLM answer, threshold $\geq 7$, no gold answer. Bold indicates the highest value in each row.}
\label{tab:judge-mode-comparison}
\footnotesize
\setlength{\tabcolsep}{4pt}
\begin{tabular}{l rr rr rr}
\toprule
 & \multicolumn{2}{c}{Standard} & \multicolumn{2}{c}{Objective} & \multicolumn{2}{c}{Comparison} \\
\cmidrule(lr){2-3} \cmidrule(lr){4-5} \cmidrule(lr){6-7}
Dataset & \BLC{} & QA & \BLC{} & QA & \BLC{} & QA \\
\midrule
GSM8K            & 23.4 & \textbf{31.0} & 26.9 & 27.6 & 27.6 & 28.3 \\
MATH (Algebra)   & \textbf{40.5} & 38.2 & 36.0 & 36.0 & 34.0 & 36.7 \\
MATH (Geometry)  & 38.4 & \textbf{38.9} & 33.3 & 31.3 & 29.3 & 31.3 \\
MATH (Num.\ Thy) & 36.1 & \textbf{44.3} & \textbf{44.3} & 42.6 & 42.6 & 39.3 \\
GPQA (MC)        & 13.2 & \textbf{17.1} &  8.0 &  8.7 & 10.0 & 11.3 \\
MBPP             & 15.1 & 17.9 & 14.2 & 17.0 & 16.0 & \textbf{19.8} \\
AIME             & \textbf{19.6} & 17.6 &  9.8 & 11.8 &  5.9 & 11.8 \\
HLE              &  5.2 & \textbf{7.8} &  2.7 &  1.3 &  0.7 &  2.0 \\
\bottomrule
\end{tabular}
\end{table}

The standard protocol achieves the highest recovery on 6 of 8 datasets, confirming that the primary driver of the gap is gold answer access rather than judge mode. Between the two judge modes, comparison slightly outperforms objective for QA (mean 22.6\% vs 22.0\%) while slightly underperforming for \BLC{} (20.8\% vs 21.9\%). The comparison judge preserves more of the QA-over-\BLC{} advantage ($+$1.8pp vs $+$0.1pp for objective), likely because it evaluates relative to the LLM's own answer, which is stronger in the QA configuration (Opus) than in \BLC{} (Haiku). However, neither judge mode closes the gap to the standard protocol.

\section{Arithmetic Coding in Practice}
\label{app:block-coding}

This appendix provides detailed pseudocode for our block emission arithmetic coding scheme. The key challenge with standard arithmetic coding is that encoder and decoder must agree exactly on probability distributions. In practice, floating-point non-determinism, hardware differences, and batching effects can cause small mismatches that accumulate over long sequences, eventually causing decoding failures.

Block emission addresses this by periodically emitting the current interval midpoint whenever the range becomes too small, then restarting with a fresh $[0,1]$ interval. This limits how much drift can accumulate before being ``reset,'' making the scheme robust to minor probability discrepancies.

\subsection{Algorithms}
\label{app:algorithms}

\begin{algorithm}[H]
\caption{Block Emission Encoding}
\label{alg:encode}
\begin{algorithmic}[1]
\REQUIRE tokens $T = [t_1, \ldots, t_n]$, model $M$, context $C$, bit precision $B$, emission threshold $\epsilon$
\ENSURE encoded blocks $E$, token counts $K$

\STATE $\textit{low} \gets 0$, $\textit{high} \gets 1$
\STATE $E \gets []$, $K \gets []$, $\textit{count} \gets 0$
\STATE $M.\text{reset}(C)$ \COMMENT{Initialize model with context}

\FOR{each token $t$ in $T$}
    \STATE $\mathbf{p} \gets M.\text{get\_distribution}()$ \COMMENT{Next-token probabilities}
    \STATE $\textit{cdf} \gets \text{cumsum}(\mathbf{p})$ \COMMENT{Cumulative distribution}
    \STATE $\textit{range} \gets \textit{high} - \textit{low}$
    \STATE $\textit{new\_low} \gets \textit{low} + \textit{range} \cdot \textit{cdf}[t-1]$ \COMMENT{$\textit{cdf}[0] = 0$}
    \STATE $\textit{new\_high} \gets \textit{low} + \textit{range} \cdot \textit{cdf}[t]$

    \IF{$(\textit{new\_high} - \textit{new\_low}) < \epsilon$ \textbf{or} $\textit{count} \geq 2^5 - 1$}
        \STATE $E.\text{append}\left(\left\lfloor \frac{\textit{low} + \textit{high}}{2} \cdot (2^B - 1) \right\rfloor\right)$ \COMMENT{Emit block}
        \STATE $K.\text{append}(\textit{count})$
        \STATE $\textit{low} \gets \textit{cdf}[t-1]$, $\textit{high} \gets \textit{cdf}[t]$ \COMMENT{Fresh interval}
        \STATE $\textit{count} \gets 1$
    \ELSE
        \STATE $\textit{low} \gets \textit{new\_low}$, $\textit{high} \gets \textit{new\_high}$
        \STATE $\textit{count} \gets \textit{count} + 1$
    \ENDIF
    \STATE $M.\text{add\_token}(t)$ \COMMENT{Update model state}
\ENDFOR

\STATE $E.\text{append}\left(\left\lfloor \frac{\textit{low} + \textit{high}}{2} \cdot (2^B - 1) \right\rfloor\right)$ \COMMENT{Emit final block}
\STATE $K.\text{append}(\textit{count})$
\STATE \textbf{return} $(E, K)$
\end{algorithmic}
\end{algorithm}

\begin{algorithm}[H]
\caption{Block Emission Decoding}
\label{alg:decode}
\begin{algorithmic}[1]
\REQUIRE encoded blocks $E$, token counts $K$, model $M$, context $C$, bit precision $B$
\ENSURE decoded tokens $T$

\STATE $T \gets []$
\STATE $M.\text{reset}(C)$ \COMMENT{Initialize model with context}

\FOR{$i = 1$ to $|E|$}
    \STATE $v \gets E[i] / (2^B - 1)$ \COMMENT{Dequantize to $[0, 1]$}
    \STATE $\textit{low} \gets 0$, $\textit{high} \gets 1$

    \FOR{$j = 1$ to $K[i]$}
        \STATE $\mathbf{p} \gets M.\text{get\_distribution}()$
        \STATE $\textit{cdf} \gets \text{cumsum}(\mathbf{p})$
        \STATE $\textit{range} \gets \textit{high} - \textit{low}$
        \STATE $\textit{target} \gets (v - \textit{low}) / \textit{range}$ \COMMENT{Normalize position in interval}
        \STATE $t \gets \min\{k : \textit{cdf}[k] > \textit{target}\}$ \COMMENT{Find token via CDF lookup}
        \STATE $\textit{low} \gets \textit{low} + \textit{range} \cdot \textit{cdf}[t-1]$
        \STATE $\textit{high} \gets \textit{low} + \textit{range} \cdot \textit{cdf}[t]$
        \STATE $T.\text{append}(t)$
        \STATE $M.\text{add\_token}(t)$
    \ENDFOR
\ENDFOR
\STATE \textbf{return} $T$
\end{algorithmic}
\end{algorithm}

\subsection{Design Rationale}

\paragraph{Block Emission.} Standard arithmetic coding maintains a single interval $[\textit{low}, \textit{high}]$ that narrows with each encoded symbol. When encoder and decoder probability distributions differ slightly (due to floating-point non-determinism, hardware variations, or batching effects), these small mismatches accumulate over the sequence. Block emission addresses this by emitting the interval midpoint (quantized to $B$ bits) whenever the range falls below threshold $\epsilon$, then restarting with a fresh $[0,1]$ interval. This periodic reset limits how much drift can accumulate, making the scheme robust to minor probability discrepancies between encoder and decoder.

\paragraph{Implementation Details.} In our experiments we use $B = 58$ bits for each emitted midpoint and allocate a small number of additional bits ($b$) for a per-block token count that records how many tokens are packed into each block. The total overhead per block is therefore $B + b \approx 64$ bits, and a block contains approximately 64 tokens, so the amortized overhead is approximately $0.125$ bits per token.

\subsection{PMATIC Self-Compression Results}
\label{app:pmatic-results}

To validate the PMATIC fallback discussed in Section~\ref{app:nondeterminism}, we ran self-compression on a corpus of 90 QA transcripts (86,840 tokens, 408,036 bytes of original UTF-8 text) using Qwen3-4B as both the SLM and the arithmetic-coding distribution. With helper bits and per-block CRCs (512 tokens per block), the compressed output totaled 8{,}132 bytes: 5{,}523 bytes of payload, 1{,}001 bytes of helper bits, and 1{,}608 bytes of block-level overhead. This corresponds to 0.75 bits per token, or roughly $50\times$ compression relative to the tokenized representation and $\sim$2\% of the original UTF-8 size. The cross-entropy lower bound on this corpus is 0.45 bits per token, so the PMATIC overhead (helper bits plus block CRCs) accounts for roughly $0.30$ bpt, or about a $1.6\times$ multiplier over the information-theoretic floor. In practical terms, this means that even when full determinism cannot be guaranteed, the egress cost of transmitting a QA transcript grows by less than $2\times$ relative to an idealized lossless coder.

\section{Q\&A Compression Detailed Results (Correct-Answer Ablation)}
\label{app:qa-results-detailed}

This appendix presents an ablation where the LLM answering yes/no questions is given the correct answer from the dataset directly, rather than generating its own solution as a reference. Additionally, all $N=10$ questions are generated in a single batch with no iterative quality thresholding or early stopping. This provides an approximation of an upper bound on the protocol's potential: access to the correct answer ensures the LLM knows the final answer, but it does not provide the reasoning that produced it, so the LLM cannot compare the SLM's reasoning path against a reference derivation. In practice, the LLM's own solution (used in the main results, Table~\ref{tab:qa-asker}) may sometimes be more useful for answering questions about \emph{how} to arrive at the answer, even if the solution itself is occasionally wrong.

We compare two configurations:
\begin{itemize}
    \item \textbf{\BLC{} (Bit-Limited Chain-of-Thought)}: SLM=Haiku, LLM=Haiku, Q=Haiku. The small model asks questions answered by itself.
    \item \textbf{QA}: SLM=Haiku, LLM=Opus, Q=Haiku. The small model asks questions answered by the large model (with access to the correct answer).
\end{itemize}

\subsection{Recovery Rate by Difficulty (Claude 4.5)}

Table~\ref{tab:qa-accuracy} reports the recovery rate broken down by difficulty level. Problems are grouped as \textbf{Medium} (Haiku fails, Opus succeeds), \textbf{Hard} (Haiku and Sonnet fail, Opus succeeds), and \textbf{Very Hard} (all models fail). $n$ is the total number of non-easy problems per dataset.

\begin{table}[H]
\centering
\caption{Recovery rate (\%) on problems where Haiku initially fails (Claude 4.5). \BLC{} = Bit-Limited Chain-of-Thought (haiku$\to$haiku$\to$haiku); QA = QA Compression (haiku$\to$opus$\to$haiku). $\Delta$ = QA $-$ \BLC{}. Higher is better.}
\label{tab:qa-accuracy}
\scriptsize
\setlength{\tabcolsep}{3pt}
\begin{tabular}{l | r | rr | rr | rr | rr r}
\toprule
 & & \multicolumn{2}{c|}{Medium} & \multicolumn{2}{c|}{Hard} & \multicolumn{2}{c|}{Very Hard} & \multicolumn{3}{c}{All Non-Easy} \\
\cmidrule(lr){3-4} \cmidrule(lr){5-6} \cmidrule(lr){7-8} \cmidrule(lr){9-11}
Dataset & $n$ & \BLC{} & QA & \BLC{} & QA & \BLC{} & QA & \BLC{} & QA & $\Delta$ \\
\midrule
GSM8K & 145 & 19.0 & \textbf{23.8} & 20.0 & \textbf{60.0} & 37.1 & \textbf{48.6} & 23.4 & \textbf{31.0} & \small+7.6 \\
MATH (Algebra) & 259 & \textbf{28.3} & 20.0 & 48.0 & 48.0 & \textbf{43.7} & 43.1 & \textbf{40.5} & 38.2 & \small$-$2.3 \\
MATH (Geometry) & 190 & \textbf{50.0} & 43.3 & 28.6 & 28.6 & 37.4 & \textbf{39.6} & 38.4 & \textbf{38.9} & \small+0.5 \\
MATH (Num.\ Theory) & 61 & 55.6 & \textbf{88.9} & 0.0 & \textbf{66.7} & 34.7 & 34.7 & 36.1 & \textbf{44.3} & \small+8.2 \\
GPQA (MC) & 152 & \textbf{50.0} & 33.3 & 11.1 & \textbf{33.3} & 11.7 & \textbf{15.3} & 13.2 & \textbf{17.1} & \small+3.9 \\
MBPP & 106 & 48.1 & \textbf{59.3} & 5.9 & 5.9 & 3.2 & 3.2 & 15.1 & \textbf{17.9} & \small+2.8 \\
AIME & 51 & 0.0 & 0.0 & \textbf{10.7} & 7.1 & 41.2 & 41.2 & \textbf{19.6} & 17.6 & \small$-$2.0 \\
HLE$^\dagger$ & 306 & 14.0 & 14.0 & 3.8 & \textbf{7.1} & 3.0 & \textbf{4.0} & 5.2 & \textbf{7.8} & \small+2.6 \\
\bottomrule
\end{tabular}
\vspace{2pt}

{\footnotesize $^\dagger$\,For HLE, Very Hard problems were randomly subsampled to 100 (out of ${\sim}$1{,}900) due to evaluation cost.}
\end{table}

Q\&A guidance from Opus helps most on Medium problems, with GSM8K showing +7.6\% overall and MATH Number Theory showing the strongest Medium improvement (55.6\% to 88.9\%). On Very Hard problems where even Opus fails, Q\&A provides minimal benefit.

\subsection{100-Bit QA Compression (Claude 4.5)}
\label{app:qa-100q}

We also evaluated QA compression with 100 yes/no questions instead of 10. Table~\ref{tab:qa-100q-results} reports accuracy with standard deviation over 3 trials. Unlike the 10-question ablation above (which uses the gold answer as reference), this experiment uses the LLM's own generated solution.

\begin{table}[H]
\centering
\caption{QA Compression accuracy on non-easy problems (\%, mean\sd{std} over 3 trials) comparing 10-bit and 100-bit protocols (Claude 4.5). \BLC{} = haiku$\to$haiku$\to$haiku; QA = haiku$\to$opus$\to$haiku; QA+ = haiku$\to$opus$\to$opus.}
\label{tab:qa-100q-results}
\scriptsize
\setlength{\tabcolsep}{3pt}
\begin{tabular}{l r | rrr | rrr | rrr}
\toprule
 & & \multicolumn{3}{c|}{\BLC{}} & \multicolumn{3}{c|}{QA} & \multicolumn{3}{c}{QA+} \\
\cmidrule(lr){3-5} \cmidrule(lr){6-8} \cmidrule(lr){9-11}
Dataset & $n$ & 10-bit & 100-bit & $\Delta$ & 10-bit & 100-bit & $\Delta$ & 10-bit & 100-bit & $\Delta$ \\
\midrule
GSM8K & 145 & 22.5\sd{1.2} & 24.6\sd{1.4} & +2.1 & 26.7\sd{0.9} & 30.8\sd{0.3} & +4.1 & 29.7\sd{1.7} & 35.4\sd{1.7} & +5.7 \\
MATH (Algebra) & 259 & 38.1\sd{1.6} & 38.2\sd{1.4} & +0.1 & 39.1\sd{1.0} & 40.0\sd{1.9} & +0.9 & 40.0\sd{1.5} & 40.2\sd{1.6} & +0.2 \\
MATH (Geometry) & 190 & 33.7\sd{0.4} & 37.4\sd{0.9} & +3.7 & 39.6\sd{0.5} & 37.2\sd{1.1} & $-$2.4 & 38.2\sd{1.5} & 41.1\sd{3.0} & +2.9 \\
MATH (Num.\ Theory) & 61 & 38.3\sd{2.0} & 39.3\sd{1.3} & +1.0 & 43.7\sd{4.3} & 41.5\sd{0.8} & $-$2.2 & 40.4\sd{3.1} & 45.9\sd{1.3} & +5.5 \\
GPQA (MC) & 152 & 11.8\sd{2.7} & 11.4\sd{3.0} & $-$0.4 & 13.6\sd{0.3} & 13.8\sd{0.0} & +0.2 & 12.1\sd{2.0} & 13.2\sd{1.4} & +1.1 \\
AIME & 51 & 18.3\sd{1.8} & 17.0\sd{4.6} & $-$1.3 & 19.0\sd{2.4} & 19.6\sd{2.8} & +0.6 & 22.2\sd{1.8} & 16.3\sd{1.8} & $-$5.9 \\
HLE & 306 & 5.4\sd{0.6} & 6.1\sd{0.6} & +0.7 & 6.5\sd{0.0} & 5.9\sd{0.9} & $-$0.6 & 9.3\sd{0.8} & 8.8\sd{0.9} & $-$0.5 \\
MBPP & 106 & 18.2\sd{1.8} & 19.2\sd{0.9} & +1.0 & 18.2\sd{0.4} & 18.6\sd{0.9} & +0.4 & 26.1\sd{0.4} & 23.6\sd{0.0} & $-$2.5 \\
\bottomrule
\end{tabular}
\end{table}

Table~\ref{tab:qa-100q-med-hard} shows the same comparison restricted to Medium and Hard problems only (excluding Very Hard).

\begin{table}[H]
\centering
\caption{QA Compression accuracy on Medium + Hard problems only (\%, mean\sd{std} over 3 trials) comparing 10-bit and 100-bit protocols (Claude 4.5). Excludes Very Hard problems where even Opus fails.}
\label{tab:qa-100q-med-hard}
\scriptsize
\setlength{\tabcolsep}{3pt}
\begin{tabular}{l r | rrr | rrr | rrr}
\toprule
 & & \multicolumn{3}{c|}{\BLC{}} & \multicolumn{3}{c|}{QA} & \multicolumn{3}{c}{QA+} \\
\cmidrule(lr){3-5} \cmidrule(lr){6-8} \cmidrule(lr){9-11}
Dataset & $n$ & 10-bit & 100-bit & $\Delta$ & 10-bit & 100-bit & $\Delta$ & 10-bit & 100-bit & $\Delta$ \\
\midrule
GSM8K & 110 & 19.1\sd{2.0} & 21.8\sd{0.7} & +2.7 & 23.3\sd{1.1} & 27.6\sd{0.4} & +4.3 & 27.6\sd{1.5} & 33.9\sd{1.1} & +6.3 \\
MATH (Algebra) & 85 & 28.6\sd{1.5} & 28.6\sd{1.1} & +0.0 & 27.5\sd{2.2} & 29.8\sd{3.9} & +2.3 & 31.0\sd{1.5} & 32.5\sd{0.6} & +1.5 \\
MATH (Geometry) & 51 & 28.1\sd{1.8} & 37.9\sd{3.7} & +9.8 & 37.3\sd{3.2} & 36.6\sd{6.1} & $-$0.7 & 37.3\sd{1.6} & 46.4\sd{5.1} & +9.1 \\
MATH (Num.\ Theory) & 12 & 66.7\sd{11.8} & 61.1\sd{3.9} & $-$5.6 & 72.2\sd{15.7} & 55.6\sd{7.9} & $-$16.6 & 63.9\sd{15.7} & 75.0\sd{6.8} & +11.1 \\
GPQA (MC) & 15 & 40.0\sd{14.4} & 33.3\sd{5.4} & $-$6.7 & 40.0\sd{14.4} & 40.0\sd{5.4} & +0.0 & 31.1\sd{6.3} & 42.2\sd{6.3} & +11.1 \\
AIME & 34 & 10.8\sd{1.4} & 7.8\sd{5.0} & $-$3.0 & 6.9\sd{1.4} & 9.8\sd{3.7} & +2.9 & 9.8\sd{1.4} & 6.9\sd{2.8} & $-$2.9 \\
HLE & 206 & 6.5\sd{0.6} & 7.6\sd{1.4} & +1.1 & 7.9\sd{0.6} & 7.6\sd{1.4} & $-$0.3 & 13.1\sd{1.4} & 11.7\sd{0.4} & $-$1.4 \\
MBPP & 44 & 36.4\sd{1.9} & 42.4\sd{2.1} & +6.0 & 37.9\sd{2.8} & 35.6\sd{3.9} & $-$2.3 & 51.5\sd{2.8} & 43.9\sd{2.8} & $-$7.6 \\
\bottomrule
\end{tabular}
\end{table}

\subsection{Claude 3.5/4 Model Family Results}

Table~\ref{tab:qa-results-v35} shows Q\&A compression results using the older Claude 3.5/4 model family.

\begin{table}[H]
\centering
\caption{Recovery rate (\%) on problems where Haiku initially fails (Claude 3.5/4). \BLC{} = Bit-Limited Chain-of-Thought (haiku$\to$haiku$\to$haiku, self-refinement baseline); QA = QA Compression (haiku$\to$opus$\to$haiku, knowledge transfer). $\Delta$ = QA $-$ \BLC{}. Bold indicates the highest value in each row (excluding $\Delta$).}
\label{tab:qa-results-v35}
\scriptsize
\setlength{\tabcolsep}{3pt}
\begin{tabular}{l | r | rr r | rr r | rr r | rr r}
\toprule
 & & \multicolumn{3}{c|}{Medium} & \multicolumn{3}{c|}{Hard} & \multicolumn{3}{c|}{Very Hard} & \multicolumn{3}{c}{All Non-Easy} \\
\cmidrule(lr){3-5} \cmidrule(lr){6-8} \cmidrule(lr){9-11} \cmidrule(lr){12-14}
Dataset & $n$ & \BLC{} & QA & $\Delta$ & \BLC{} & QA & $\Delta$ & \BLC{} & QA & $\Delta$ & \BLC{} & QA & $\Delta$ \\
\midrule
GSM8K & 109 & 30.0 & \textbf{50.0} & \small+20.0 & \textbf{20.0} & 0.0 & \small$-$20.0 & \textbf{52.3} & 45.5 & \small$-$6.8 & 38.5 & \textbf{45.9} & \small+7.4 \\
MATH (Algebra) & 342 & 29.5 & \textbf{32.0} & \small+2.5 & \textbf{50.0} & 41.7 & \small$-$8.3 & \textbf{52.4} & 49.5 & \small$-$2.9 & \textbf{44.2} & 43.0 & \small$-$1.2 \\
MATH (Geometry) & 269 & 23.9 & 23.9 & \small 0.0 & 12.5 & 12.5 & \small 0.0 & \textbf{15.8} & 13.8 & \small$-$2.0 & \textbf{19.0} & 17.8 & \small$-$1.2 \\
MATH (Num.\ Theory) & 154 & 43.2 & \textbf{46.3} & \small+3.1 & \textbf{50.0} & 33.3 & \small$-$16.7 & 24.5 & 24.5 & \small 0.0 & 37.0 & \textbf{38.3} & \small+1.3 \\
GPQA (MC) & 156 & 11.1 & \textbf{22.2} & \small+11.1 & \textbf{33.3} & 11.1 & \small$-$22.2 & \textbf{12.3} & 11.6 & \small$-$0.7 & \textbf{13.5} & 12.2 & \small$-$1.3 \\
MBPP & 100 & 50.0 & \textbf{57.1} & \small+7.1 & 20.0 & 20.0 & \small 0.0 & \textbf{12.3} & 9.9 & \small$-$2.4 & \textbf{18.0} & 17.0 & \small$-$1.0 \\
AIME & 87 & 0.0 & 0.0 & \small 0.0 & 0.0 & 0.0 & \small 0.0 & 0.0 & 0.0 & \small 0.0 & 0.0 & 0.0 & \small 0.0 \\
HLE & 164 & 5.9 & 5.9 & \small 0.0 & 1.2 & \textbf{6.2} & \small+5.0 & 2.0 & \textbf{4.0} & \small+2.0 & 2.4 & \textbf{5.5} & \small+3.1 \\
\bottomrule
\end{tabular}
\end{table}

\subsection{Cross-Model Validation with GPT-OSS}
\label{app:gptoss-results}

To validate that the QA protocol generalizes beyond Claude models, we evaluate using GPT-OSS-120B~\citep{openai2025gptoss120bgptoss20bmodel}, an open-source reasoning model, as the SLM.

We test three configurations:
\begin{itemize}
    \item \textbf{\BLC{} (g/g/g)}: SLM=GPT-OSS, LLM=GPT-OSS, Q=GPT-OSS. Self-refinement baseline.
    \item \textbf{QA (g/o/g)}: SLM=GPT-OSS, LLM=Opus, Q=GPT-OSS. Cross-model knowledge transfer.
    \item \textbf{QA+ (g/o/o)}: SLM=GPT-OSS, LLM=Opus, Q=Opus. Full cross-model protocol.
\end{itemize}

Problems are filtered to those where GPT-OSS initially fails but Opus succeeds, analogous to the Medium+Hard filtering used for Haiku in the main results.

\begin{table}[H]
\centering
\caption{Recovery rate (\%) with GPT-OSS as SLM on non-easy problems (mean $\pm$ std over 3 trials). $n$ = number of problems where GPT-OSS initially fails.}
\label{tab:gptoss-qa}
\small
\begin{tabular}{l | r | c c c}
\toprule
Dataset & $n$ & \BLC{} (g/g/g) & QA (g/o/g) & QA+ (g/o/o) \\
\midrule
GSM8K & 59 & $37.3 \pm 3.7$ & $37.9 \pm 5.6$ & $\mathbf{41.8 \pm 0.8}$ \\
MATH (Algebra) & 240 & $19.9 \pm 1.7$ & $\mathbf{20.4 \pm 1.0}$ & $19.7 \pm 0.9$ \\
MATH (Geometry) & 197 & $17.1 \pm 0.2$ & $18.8 \pm 1.8$ & $\mathbf{24.4 \pm 0.4}$ \\
MATH (Num.\ Theory) & 64 & $28.6 \pm 1.9$ & $\mathbf{30.2 \pm 0.7}$ & $29.7 \pm 4.6$ \\
MBPP & 88 & $11.4 \pm 3.2$ & $12.1 \pm 1.4$ & $\mathbf{12.9 \pm 1.9}$ \\
AIME & 83 & $5.6 \pm 2.5$ & $6.8 \pm 2.0$ & $\mathbf{13.7 \pm 2.0}$ \\
HLE$^\dagger$ & 275 & $5.2 \pm 0.8$ & $7.4 \pm 1.1$ & $\mathbf{9.4 \pm 0.8}$ \\
\bottomrule
\end{tabular}
\vspace{2pt}

{\footnotesize $^\dagger$\,For HLE, very hard problems were randomly subsampled to 100 due to evaluation cost.}
\end{table}

\subsection{Detailed Heatmaps by Dataset (Claude 4.5)}

The following figures show Q\&A compression accuracy heatmaps for each dataset, with SLM fixed to Haiku. Each heatmap shows accuracy broken down by difficulty level (Medium, Hard+Very Hard, All Non-Easy), with rows indicating the LLM (question answerer) and columns indicating the question-generating model (Q).

\begin{figure}[H]
    \centering
    \includegraphics[width=\linewidth]{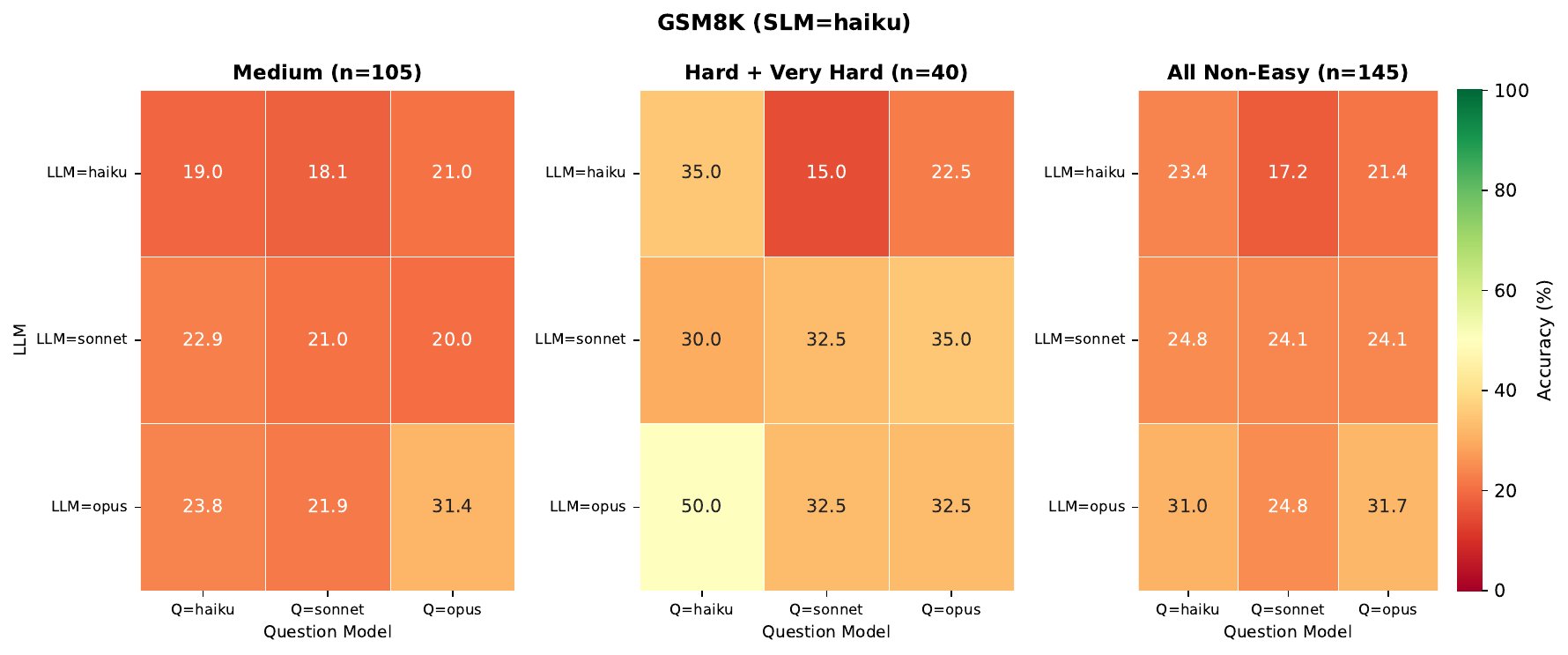}
    \caption{GSM8K Q\&A compression accuracy (SLM=Haiku, Claude 4.5). GSM8K shows consistent improvement from Q\&A guidance, particularly on Medium problems where Opus can provide useful feedback.}
    \label{fig:gsm8k-qa-v45}
\end{figure}

\begin{figure}[H]
    \centering
    \includegraphics[width=\linewidth]{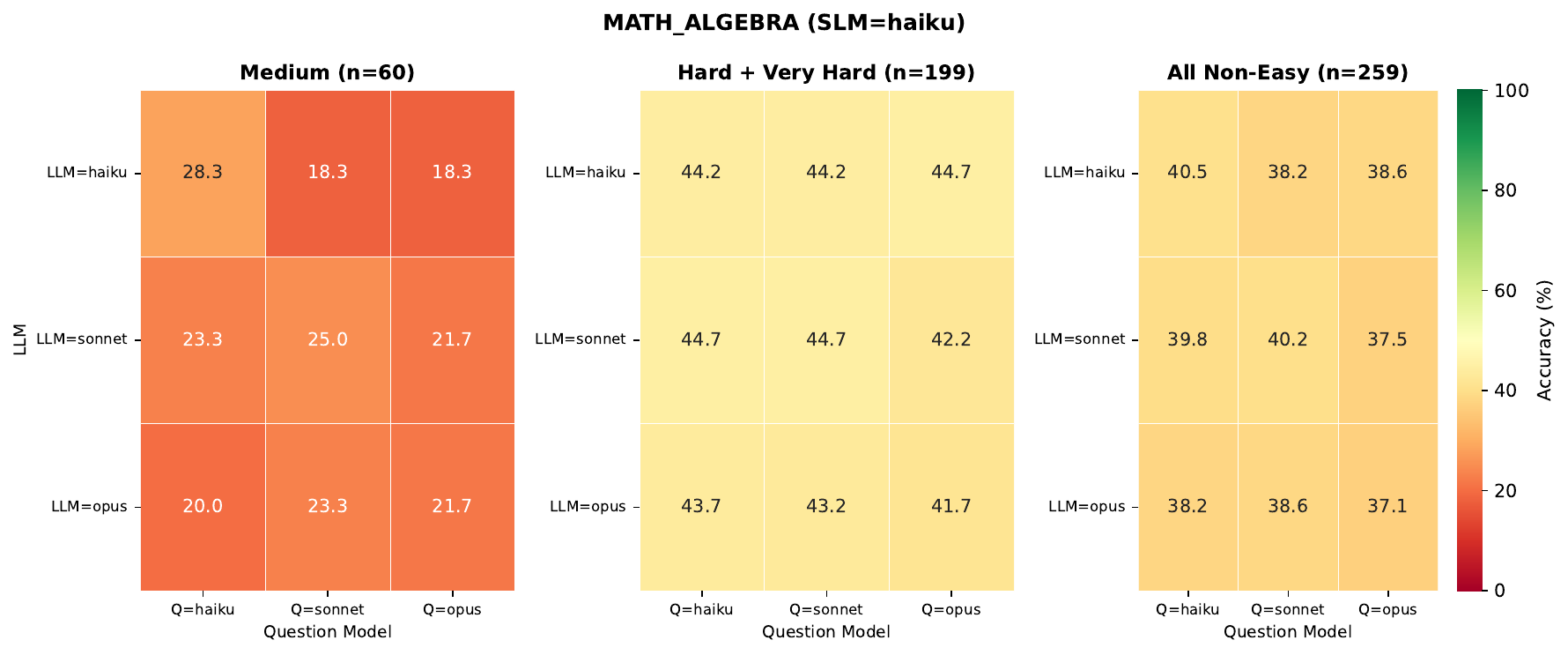}
    \caption{MATH Algebra Q\&A compression accuracy (SLM=Haiku, Claude 4.5).}
    \label{fig:math-algebra-qa-v45}
\end{figure}

\begin{figure}[H]
    \centering
    \includegraphics[width=\linewidth]{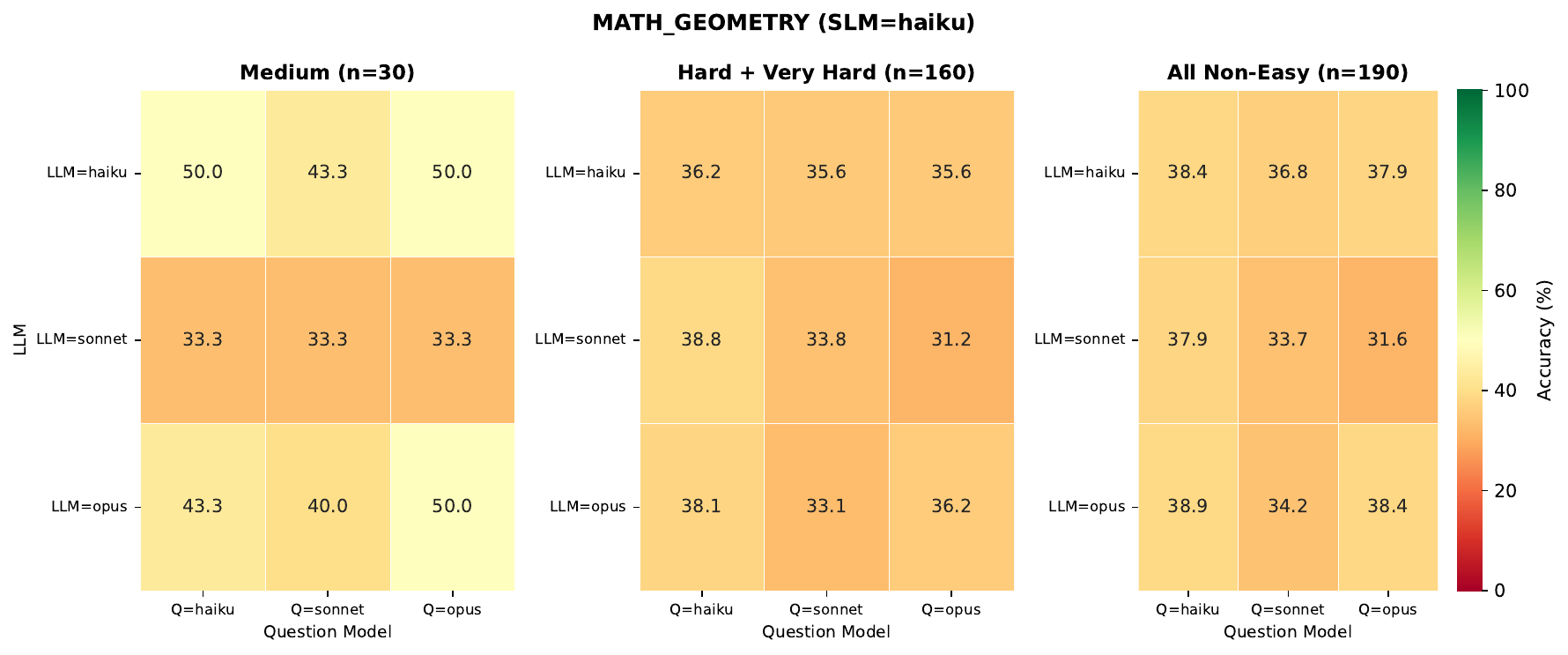}
    \caption{MATH Geometry Q\&A compression accuracy (SLM=Haiku, Claude 4.5).}
    \label{fig:math-geometry-qa-v45}
\end{figure}

\begin{figure}[H]
    \centering
    \includegraphics[width=\linewidth]{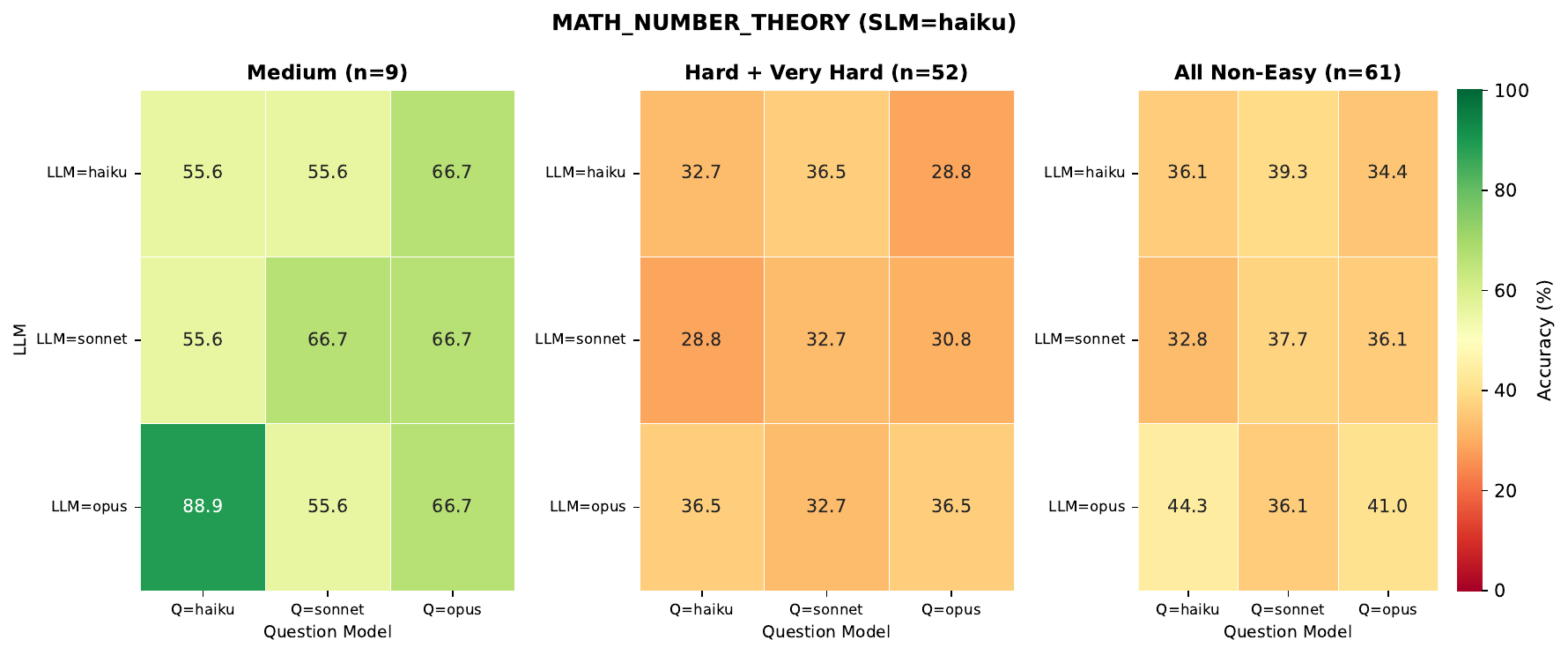}
    \caption{MATH Number Theory Q\&A compression accuracy (SLM=Haiku, Claude 4.5). Number Theory shows strong improvement on Medium problems (55.6\% to 88.9\% with Opus guidance).}
    \label{fig:math-nt-qa-v45}
\end{figure}

\begin{figure}[H]
    \centering
    \includegraphics[width=\linewidth]{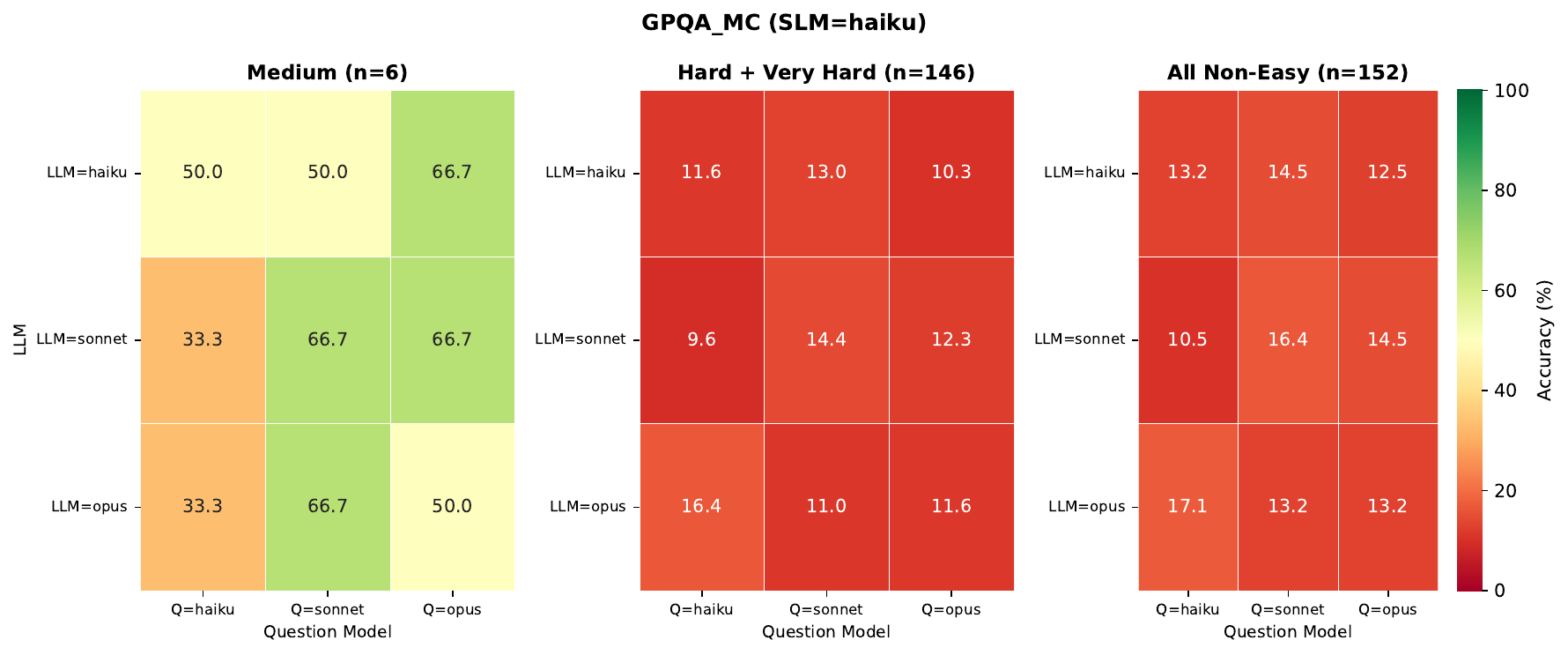}
    \caption{GPQA (MC) Q\&A compression accuracy (SLM=Haiku, Claude 4.5). The high proportion of Very Hard problems limits improvement potential.}
    \label{fig:gpqa-mc-qa-v45}
\end{figure}

\begin{figure}[H]
    \centering
    \includegraphics[width=\linewidth]{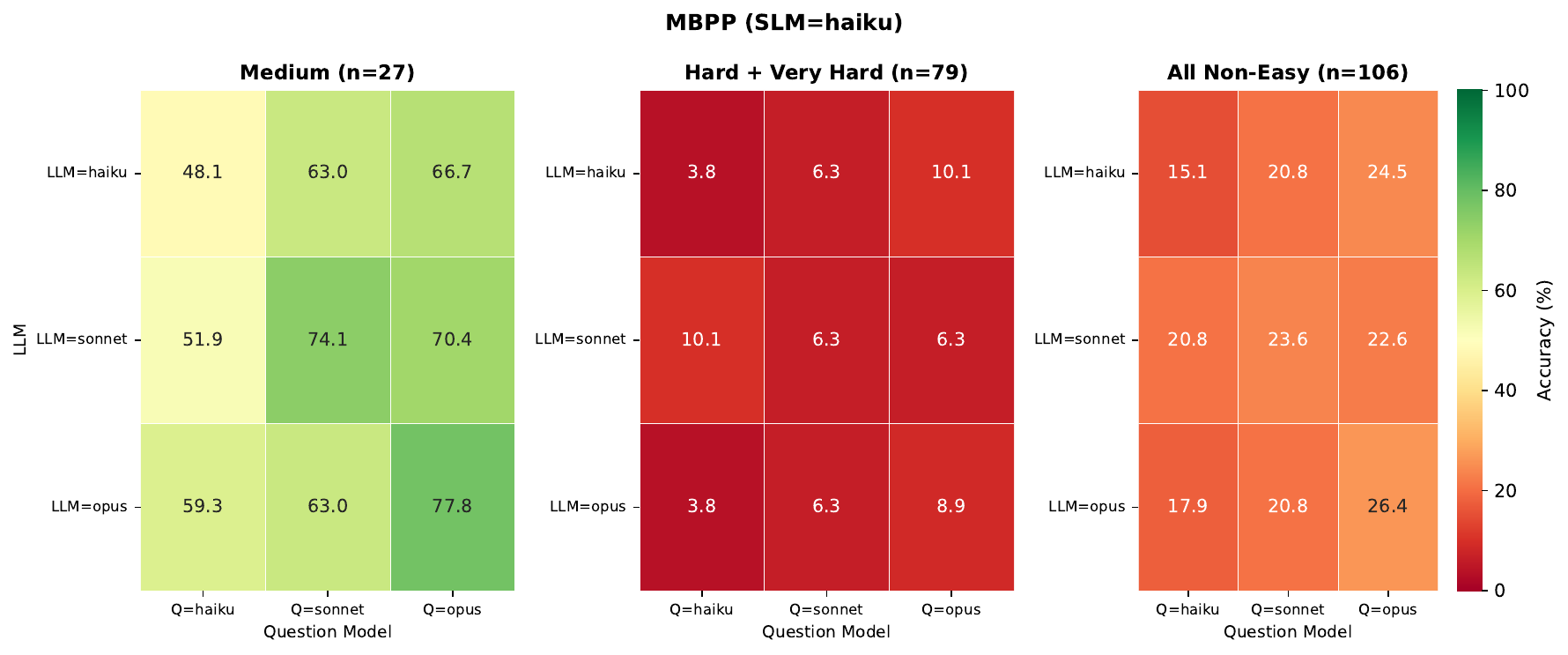}
    \caption{MBPP Q\&A compression accuracy (SLM=Haiku, Claude 4.5). Code generation proves challenging for Q\&A compression on Hard and Very Hard problems.}
    \label{fig:mbpp-qa-v45}
\end{figure}

\begin{figure}[H]
    \centering
    \includegraphics[width=\linewidth]{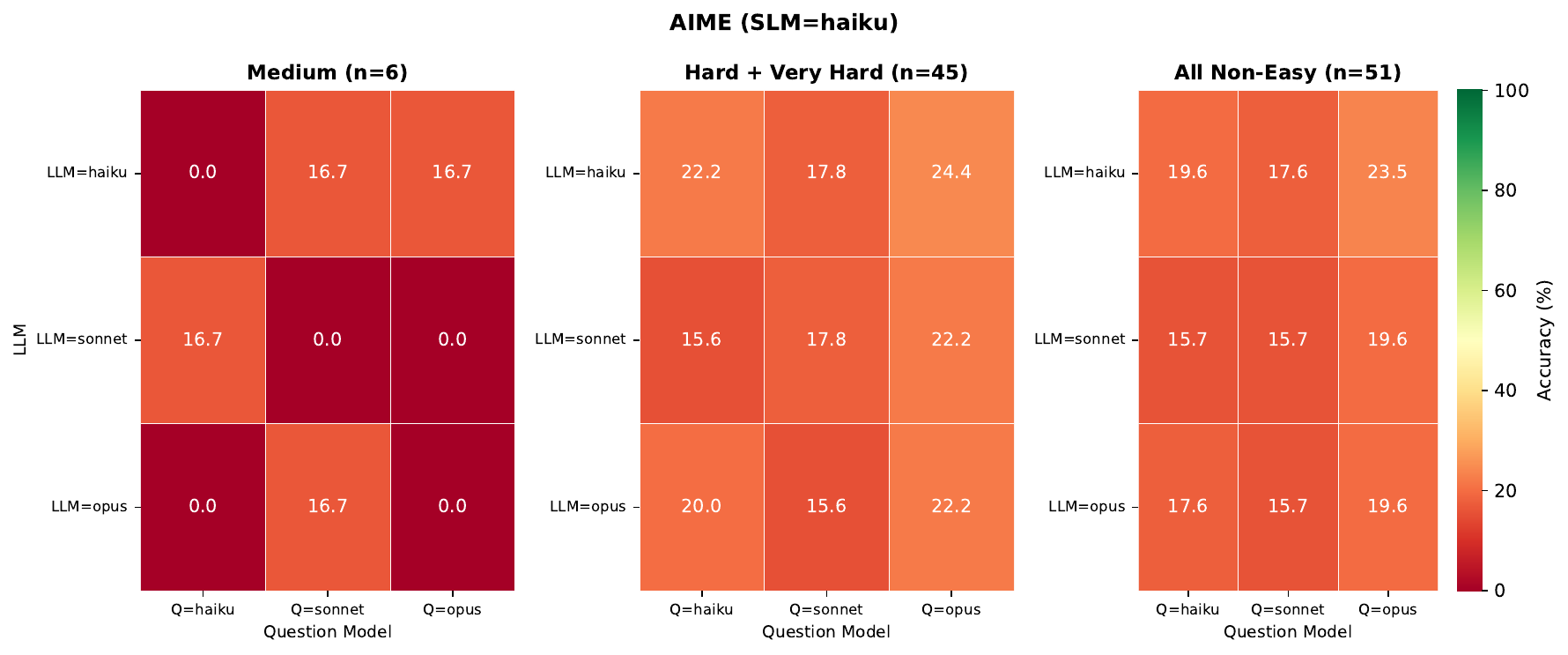}
    \caption{AIME Q\&A compression accuracy (SLM=Haiku, Claude 4.5). Competition math problems show limited benefit from Q\&A guidance.}
    \label{fig:aime-qa-v45}
\end{figure}

\begin{figure}[H]
    \centering
    \includegraphics[width=\linewidth]{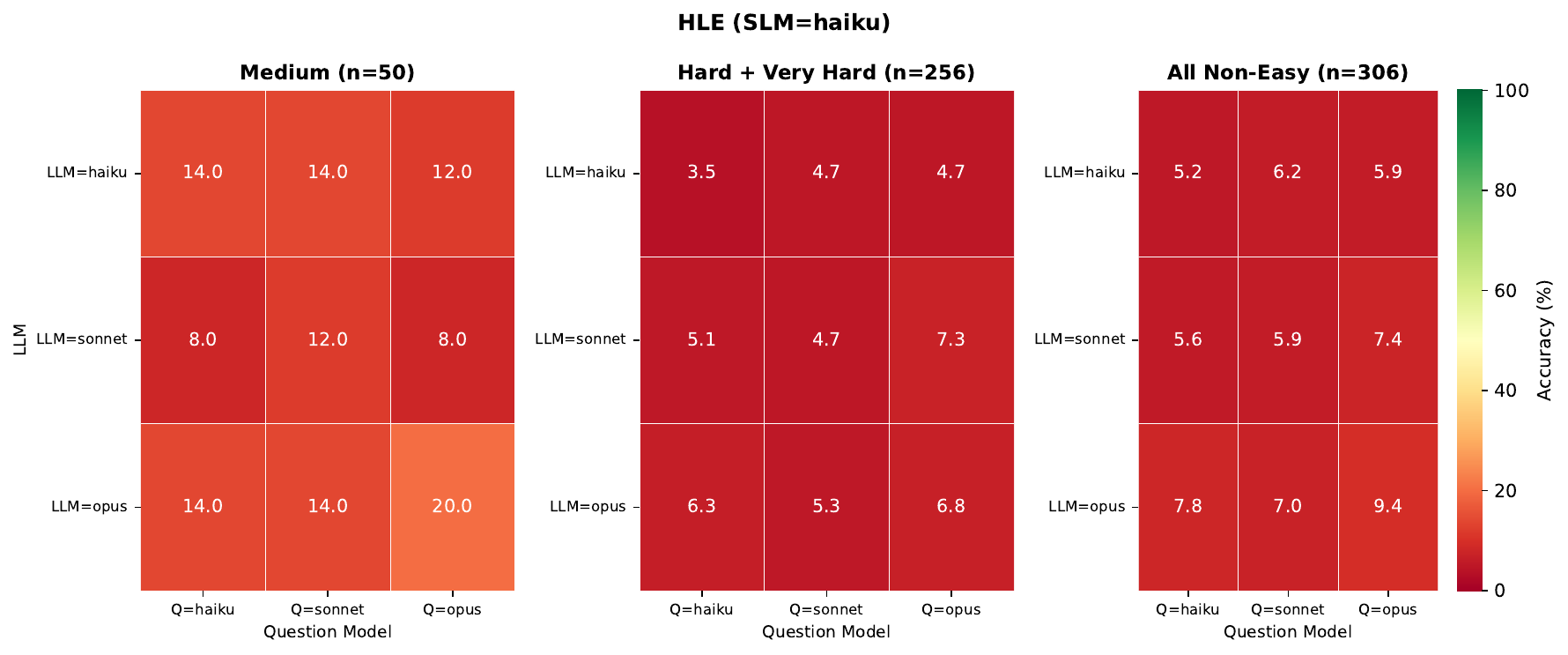}
    \caption{HLE Q\&A compression accuracy (SLM=Haiku, Claude 4.5). Despite the high Very Hard rate (89\%), Q\&A shows modest improvements on Hard problems.}
    \label{fig:hle-qa-v45}
\end{figure}

\subsection{Detailed Heatmaps by Dataset (Claude 3.5/4)}

\begin{figure}[H]
    \centering
    \includegraphics[width=\linewidth]{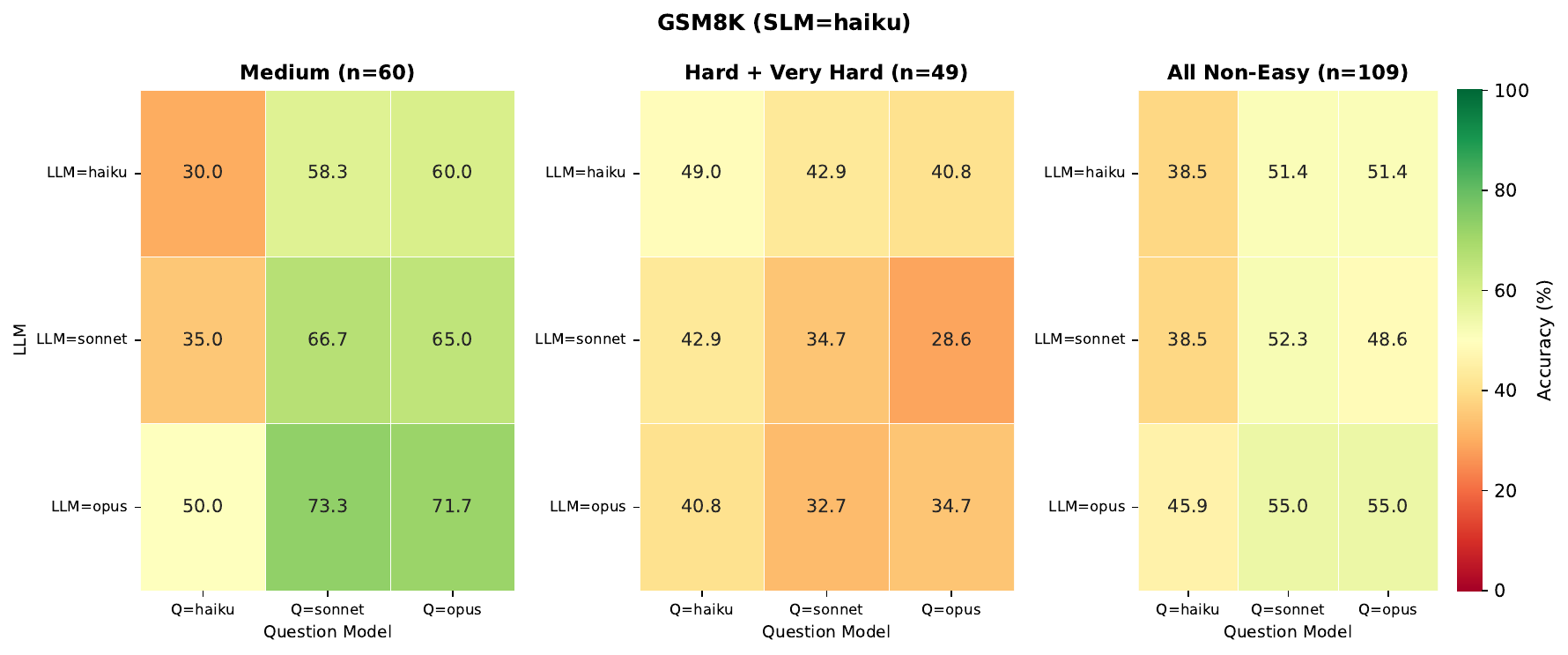}
    \caption{GSM8K Q\&A compression accuracy (SLM=Haiku, Claude 3.5/4). Strong improvement on Medium problems (30\% to 50\%).}
    \label{fig:gsm8k-qa-v35}
\end{figure}

\begin{figure}[H]
    \centering
    \includegraphics[width=\linewidth]{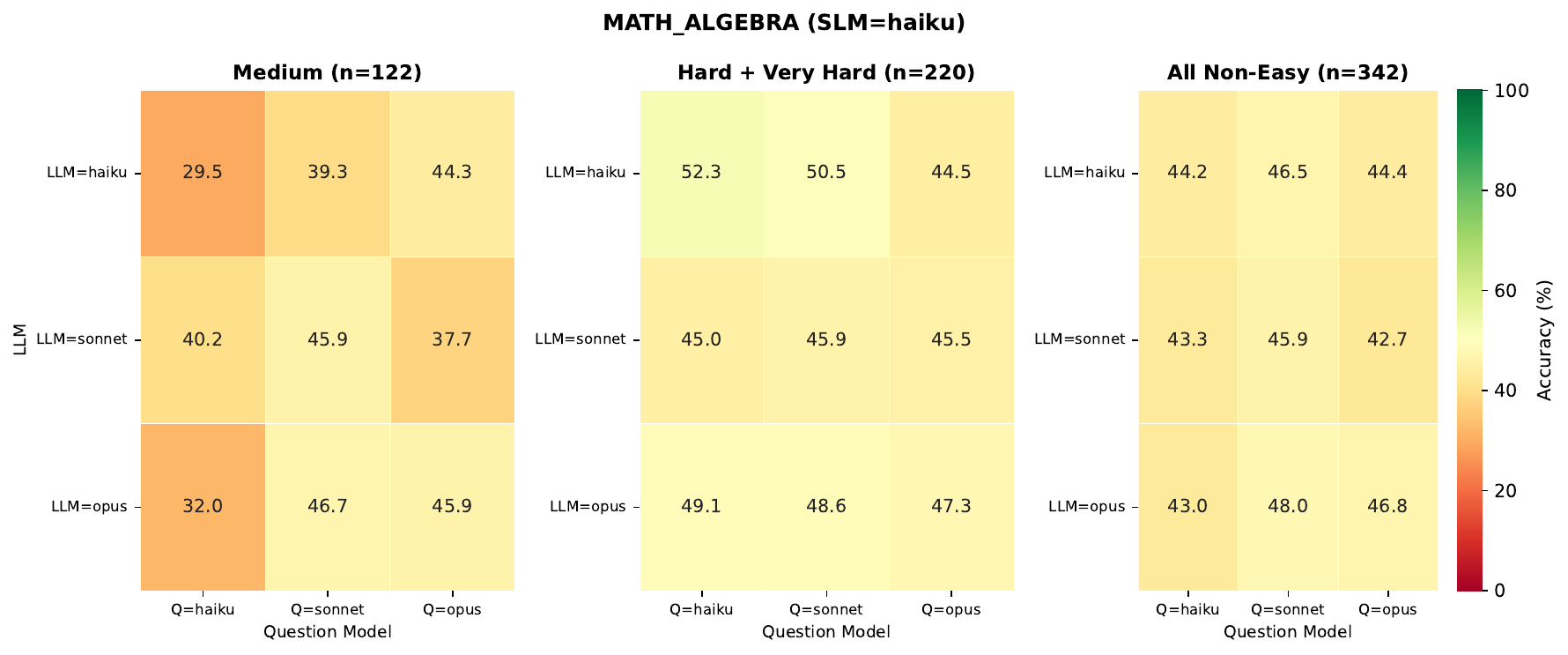}
    \caption{MATH Algebra Q\&A compression accuracy (SLM=Haiku, Claude 3.5/4).}
    \label{fig:math-algebra-qa-v35}
\end{figure}

\begin{figure}[H]
    \centering
    \includegraphics[width=\linewidth]{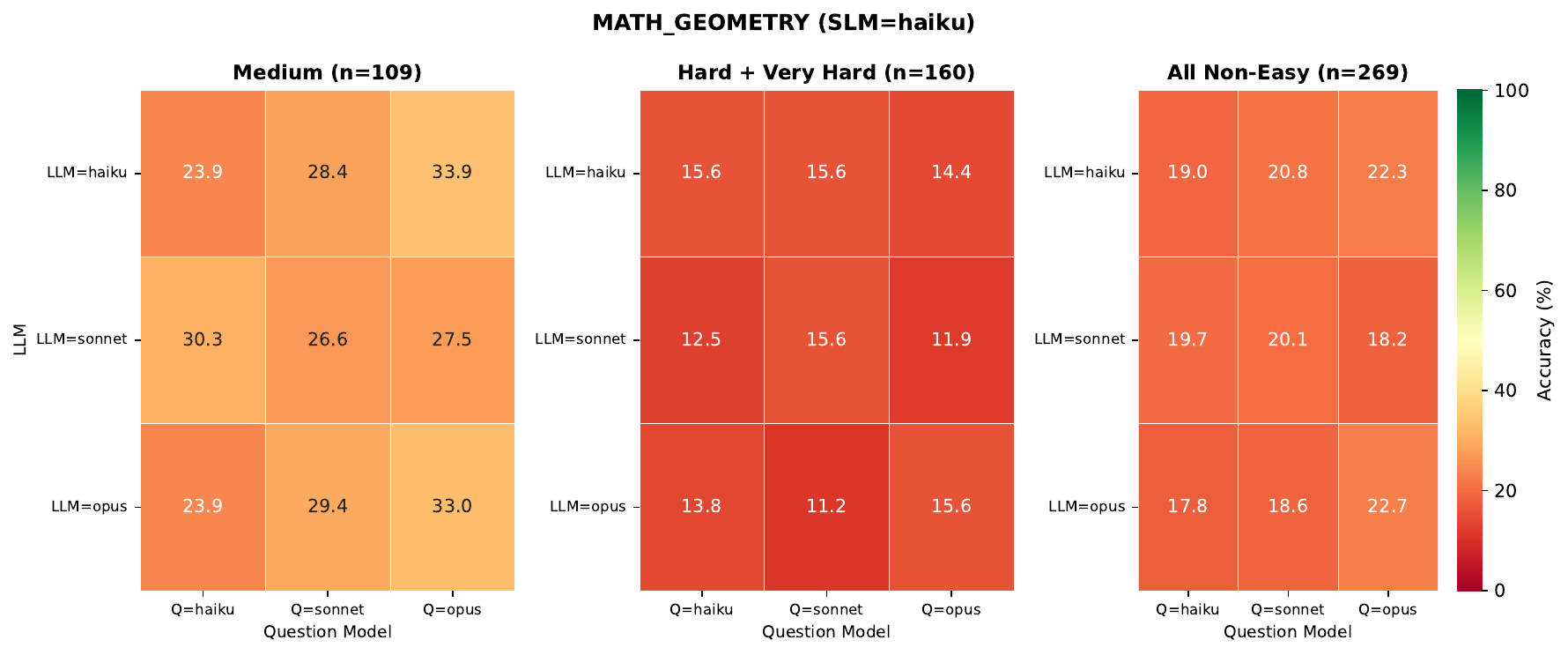}
    \caption{MATH Geometry Q\&A compression accuracy (SLM=Haiku, Claude 3.5/4).}
    \label{fig:math-geometry-qa-v35}
\end{figure}

\begin{figure}[H]
    \centering
    \includegraphics[width=\linewidth]{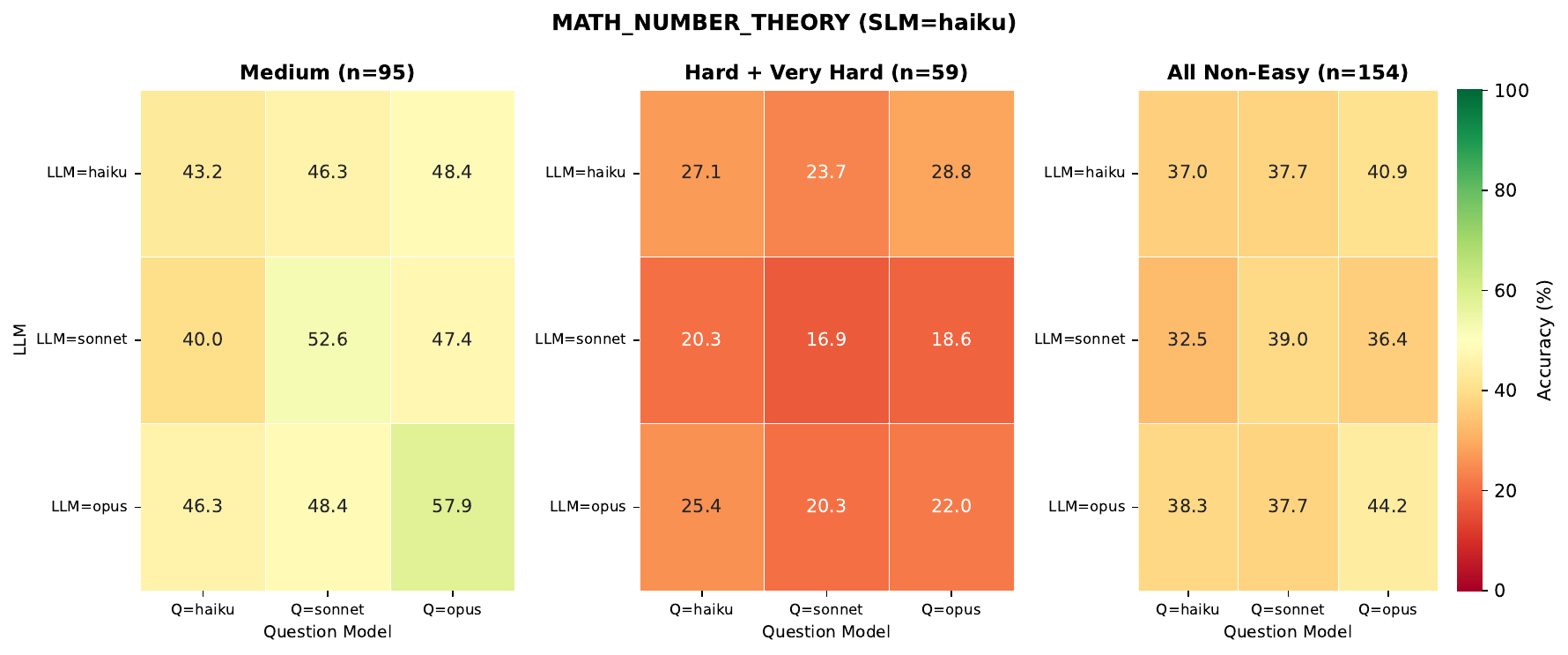}
    \caption{MATH Number Theory Q\&A compression accuracy (SLM=Haiku, Claude 3.5/4).}
    \label{fig:math-nt-qa-v35}
\end{figure}

\begin{figure}[H]
    \centering
    \includegraphics[width=\linewidth]{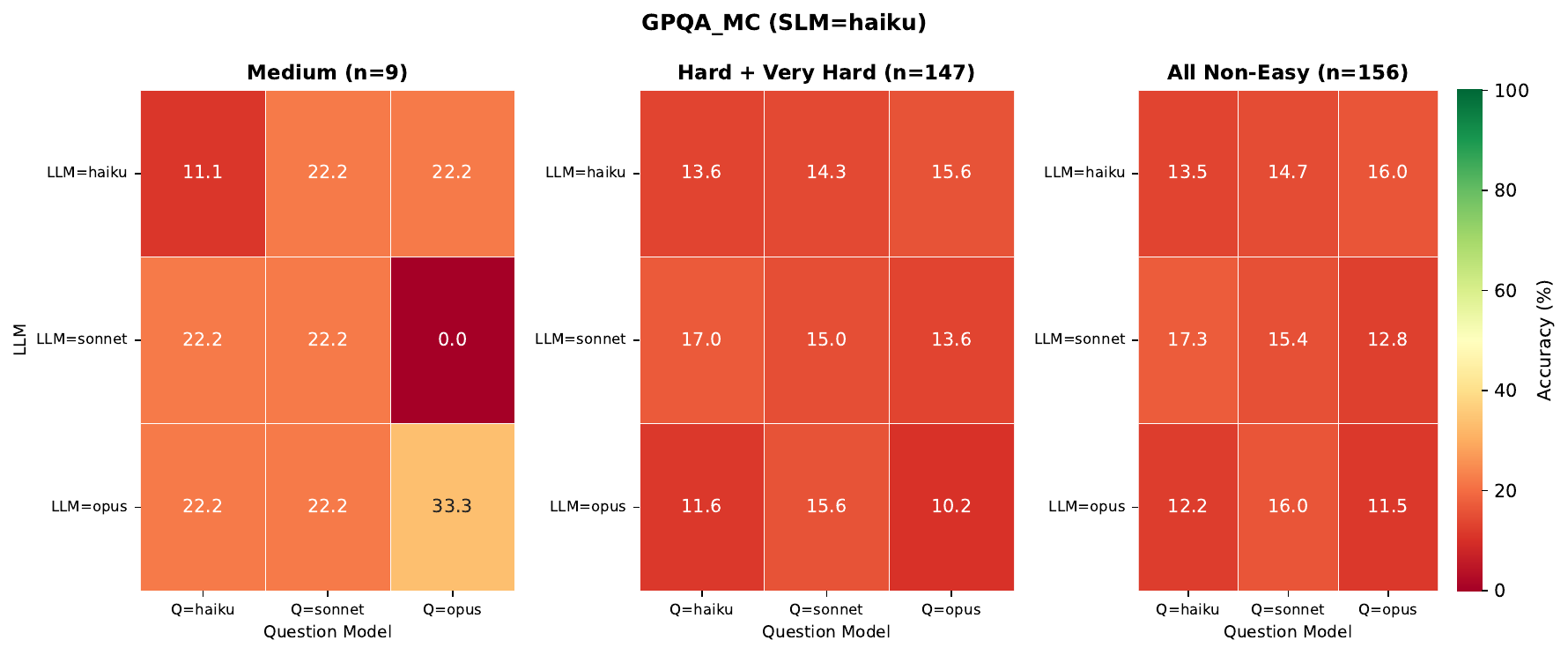}
    \caption{GPQA (MC) Q\&A compression accuracy (SLM=Haiku, Claude 3.5/4).}
    \label{fig:gpqa-mc-qa-v35}
\end{figure}

\begin{figure}[H]
    \centering
    \includegraphics[width=\linewidth]{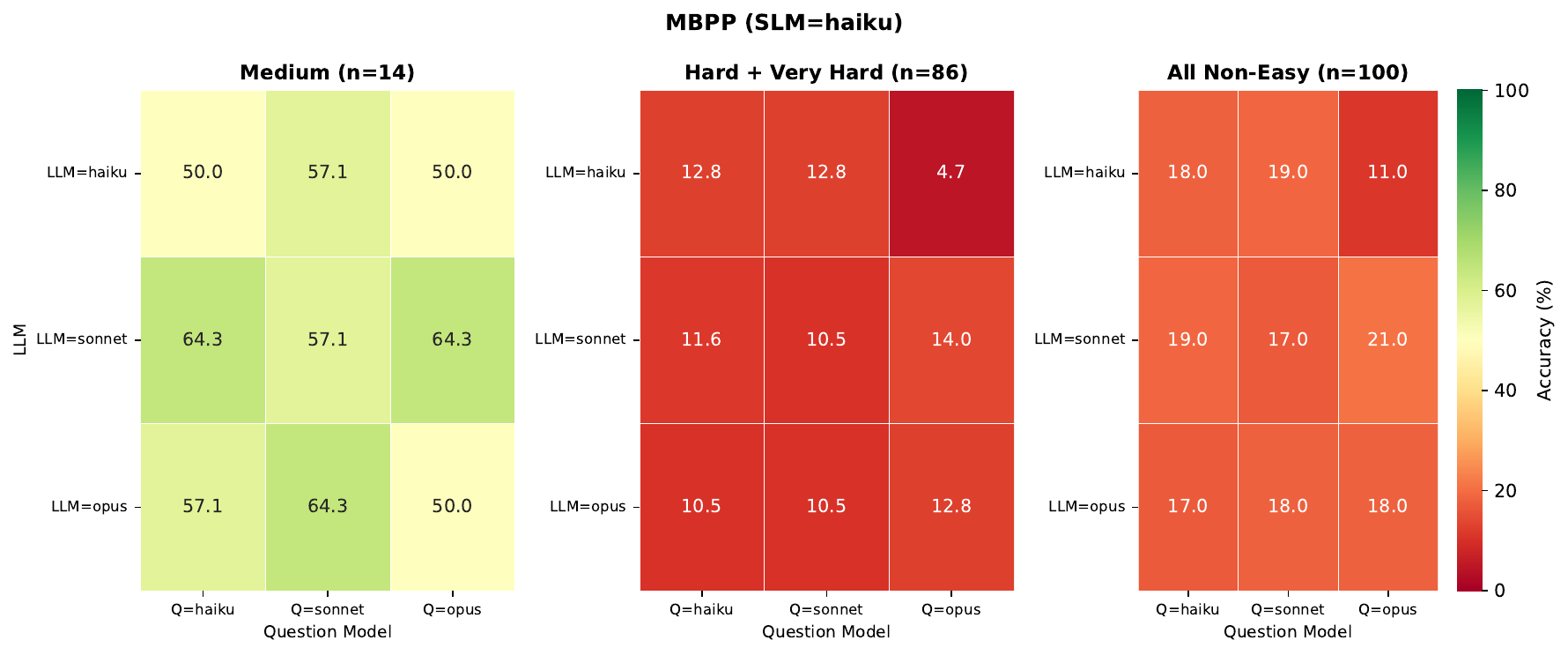}
    \caption{MBPP Q\&A compression accuracy (SLM=Haiku, Claude 3.5/4).}
    \label{fig:mbpp-qa-v35}
\end{figure}

\begin{figure}[H]
    \centering
    \includegraphics[width=\linewidth]{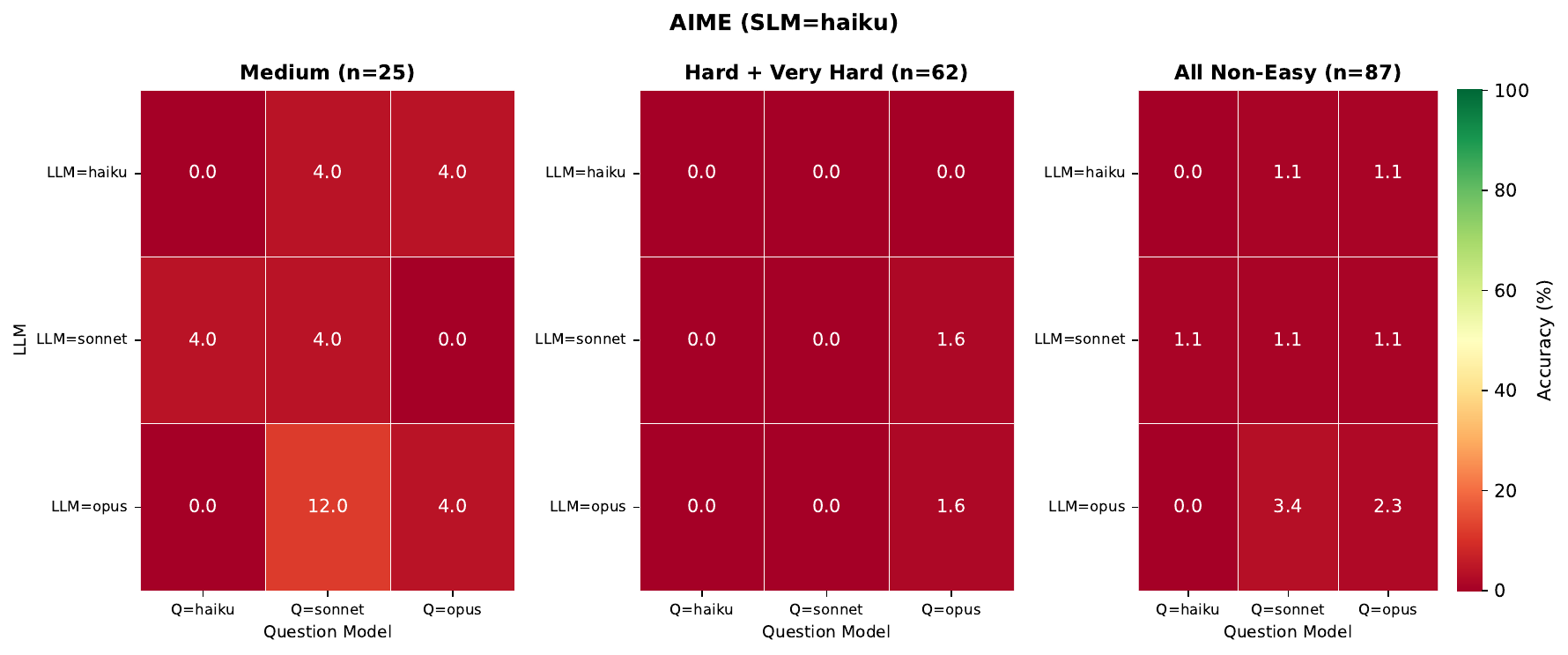}
    \caption{AIME Q\&A compression accuracy (SLM=Haiku, Claude 3.5/4). The older Haiku model achieves 0\% on all non-easy AIME problems, even with Opus guidance.}
    \label{fig:aime-qa-v35}
\end{figure}

\begin{figure}[H]
    \centering
    \includegraphics[width=\linewidth]{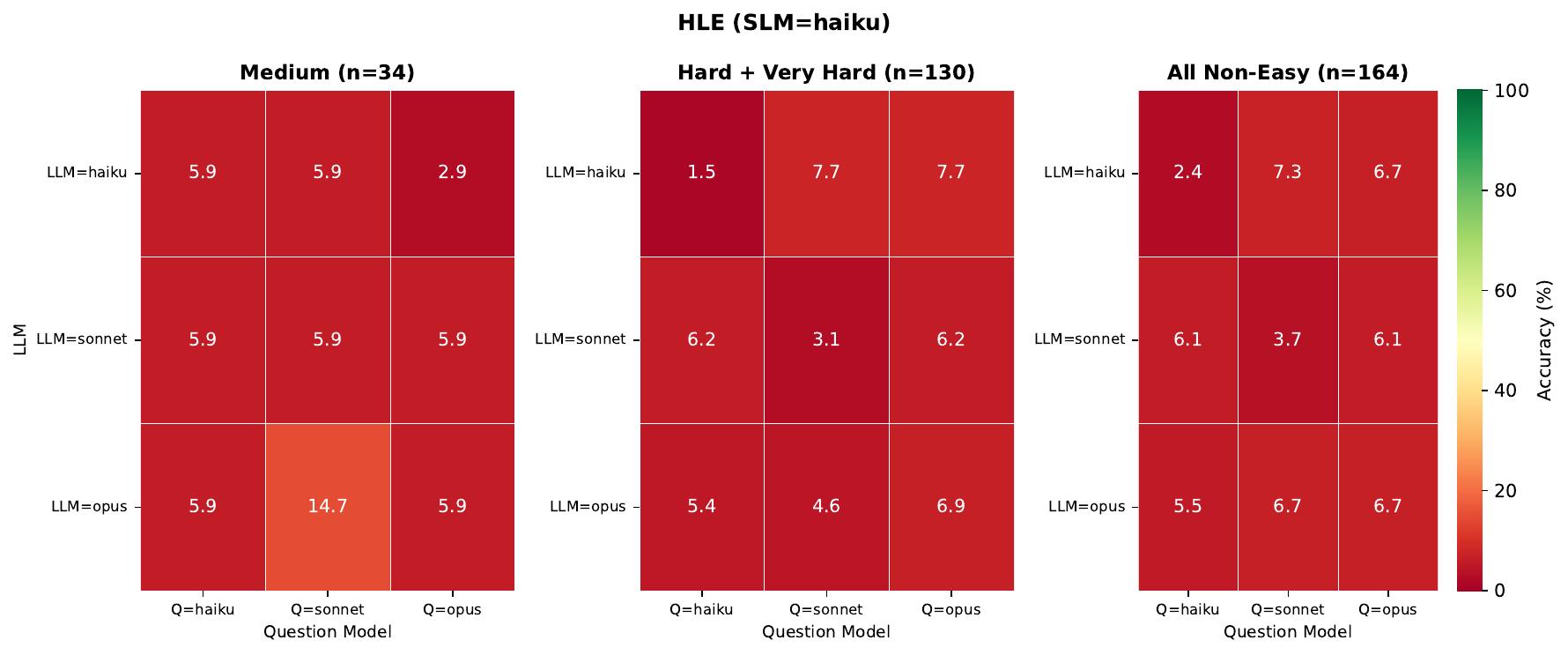}
    \caption{HLE Q\&A compression accuracy (SLM=Haiku, Claude 3.5/4). Q\&A shows improvement on Hard problems (1.2\% to 6.2\%).}
    \label{fig:hle-qa-v35}
\end{figure}

\subsection{Easy-Problem Regression Analysis}
\label{app:easy-regression}

To check whether the QA protocol causes regression on problems Haiku already solves, we run the protocol on \emph{easy} problems (those where all three models (Haiku, Sonnet, Opus) answer correctly at the baseline). Since all problems start correct, any change in accuracy is purely regression.

Table~\ref{tab:easy-reeval} reports the regression rate for three conditions: (1)~\textbf{Haiku re-eval}, where Haiku simply re-solves each problem with no protocol (establishing how often Haiku fails to reproduce its own baseline answer); (2)~\textbf{\BLC{}}, the self-refinement baseline (SLM=Haiku, LLM=Haiku); and (3)~\textbf{QA}, the full protocol (SLM=Haiku, LLM=Opus). Both \BLC{} and QA use the iterative protocol with objective judge and early stopping.

\begin{table}[H]
\centering
\caption{Easy-problem regression rates (\%) under different conditions. All problems start correct (100\% initial accuracy). ``Regression'' is the percentage of problems that become incorrect after re-evaluation. Lower is better. $^\dagger$For AIME and HLE, the high regression rates are primarily due to Haiku's inconsistency on these problems (see Haiku re-eval column), not the QA protocol itself.}
\label{tab:easy-reeval}
\small
\begin{tabular}{l r | r | rr}
\toprule
 & & Haiku & \multicolumn{2}{c}{Iterative QA Protocol} \\
 & & Re-eval & \multicolumn{2}{c}{Regression (\%)} \\
Dataset & $n$ & (\%) & \BLC{} & QA \\
\midrule
GSM8K & 150 & --- & 0.7 & 0.0 \\
MATH (Algebra) & 150 & --- & 8.7 & 7.3 \\
MATH (Geometry) & 150 & --- & 17.3 & 11.3 \\
MATH (Num.\ Theory) & 150 & --- & 8.0 & 4.7 \\
GPQA (MC) & 46 & --- & 10.9 & 10.9 \\
MBPP & 150 & --- & 1.3 & 0.7 \\
AIME$^\dagger$ & 39 & 59.0 & 74.4 & 69.2 \\
HLE$^\dagger$ & 30 & 83.3 & 80.0 & 83.3 \\
\bottomrule
\end{tabular}
\end{table}

On standard benchmarks (GSM8K, MATH, GPQA, MBPP), the QA protocol causes modest regression: 0--17\% of easy problems become incorrect, with most datasets below 10\%. On these datasets, QA regression is consistently equal to or lower than the \BLC{} self-refinement baseline, indicating that Opus guidance does not introduce additional regression.

On frontier benchmarks (AIME and HLE), regression appears severe (69--83\%), but a sanity check reveals that this is largely an artifact of Haiku's inconsistency rather than the QA protocol. When Haiku simply re-solves these ``easy'' problems with no protocol at all, 59\% of AIME and 83\% of HLE problems regress. The QA protocol's regression rates are comparable to this baseline, confirming that the protocol does not meaningfully increase regression. Rather, the underlying difficulty classification is unreliable for these benchmarks, where a single correct answer at baseline does not guarantee consistent reproducibility.

\subsection{Example Q\&A Transcripts}
\label{app:qa-transcripts}

The following are example transcripts from the Q\&A compression protocol (SLM=Haiku, LLM=Opus, Q=Haiku, Claude 4.5). For each dataset, we selected an arbitrary problem where Haiku initially answered incorrectly but produced a correct answer after 10 binary questions, with the additional constraint that the transcript contains both ``Yes'' and ``No'' responses. The ``No'' answers are typically the most informative, as they correct a flawed assumption or redirect the SLM's reasoning.

\newsavebox{\qabox}
\newenvironment{qatranscript}[1]{%
  \begin{lrbox}{\qabox}%
  \begin{minipage}{0.93\linewidth}%
  \smallskip
  {\bfseries\large #1}\par\medskip
}{%
  \smallskip
  \end{minipage}%
  \end{lrbox}%
  \par\medskip\noindent\fcolorbox{gray!40}{gray!8}{\usebox{\qabox}}\par\medskip
}
\newcommand{\qaseparator}{\par\smallskip\noindent\rule{\linewidth}{0.4pt}\par\smallskip}
\newcommand{\qaitem}[3]{\textbf{Q#1:} #2 \hfill \textbf{#3}\\[2pt]}

\begin{qatranscript}{GSM8K -- Medium}
\textbf{Problem:} John buys milk for 2 dollars, eggs for 3 dollars, light bulbs for 3 dollars, cups for 3 dollars, and roach traps for 4 dollars. If there is a 10\% tax on all nonfood items how much did everything cost?

\medskip
\textbf{Gold Answer:} 16
\qaseparator
\small
\qaitem{1}{Is milk correctly identified as a food item that should not be taxed?}{Yes}
\qaitem{2}{Are eggs correctly identified as a food item that should not be taxed?}{Yes}
\qaitem{3}{Is the categorization correct that light bulbs are a nonfood item subject to tax?}{Yes}
\qaitem{4}{Is the categorization correct that cups are a nonfood item subject to tax?}{Yes}
\qaitem{5}{Is the categorization correct that roach traps are a nonfood item subject to tax?}{Yes}
\qaitem{6}{Is the sum of all nonfood items before tax correctly calculated as \$10?}{Yes}
\qaitem{7}{Is the 10\% tax calculation on the \$10 nonfood subtotal correctly computed as \$1?}{Yes}
\qaitem{8}{Should the tax be applied to the nonfood items only, rather than to all items?}{Yes}
\qaitem{9}{Is the final total of \$16 correct when adding the \$5 food subtotal and the \$11 taxed nonfood subtotal?}{No}
\qaitem{10}{Does the solution correctly account for all five items that John purchased?}{Yes}

\medskip
\textit{Haiku's initial calculation was wrong. Q9 revealed an arithmetic error in the final total, which Haiku corrected in its revised answer.}
\end{qatranscript}

\begin{qatranscript}{MATH Geometry -- Medium}
\textbf{Problem:} Corner $A$ of a rectangular piece of paper of width 8 inches is folded over so that it coincides with point $C$ on the opposite side. If $BC = 5$ inches, find the length in inches of fold $l$.

\medskip
\textbf{Gold Answer:} $5\sqrt{5}$
\qaseparator
\small
\qaitem{1}{Is the fold line $l$ the same as the line segment $BC$?}{No}
\qaitem{2}{When corner $A$ is folded to point $C$, is the fold line the perpendicular bisector of segment $AC$?}{Yes}
\qaitem{3}{Given that $BA = BC = 5$ and $B$ is on the top edge while $A$ is the top-right corner, is $|8 - b| = 5$ giving $b = 3$ correct?}{Yes}
\qaitem{4}{Is the height of the rectangle $h = 10$ inches?}{Yes}
\qaitem{5}{Is point $C$ located at $(0, 6)$ on the left edge, making $AC = 4\sqrt{5}$ inches?}{Yes}
\qaitem{6}{Given coordinates $A = (8, 10)$, $B = (3, 10)$, $C = (0, 6)$, is $BC = 5$ inches?}{Yes}
\qaitem{7}{If the fold line $BC$ has length 5 inches, is this the answer?}{No}
\qaitem{8}{Should we verify that $\vec{BC} \perp \vec{AC}$?}{No}
\qaitem{9}{Is there additional info constraining the height $h$ to exactly 10 inches?}{No}
\qaitem{10}{Should we verify that folding $A$ along line $BC$ truly maps $A$ onto $C$?}{Yes}

\medskip
\textit{Haiku initially confused the fold line with segment $BC$. Q1 and Q7 corrected this misconception, leading Haiku to compute the correct fold length.}
\end{qatranscript}

\begin{qatranscript}{MBPP -- Medium}
\textbf{Problem:} Write a function to find the maximum difference between available pairs in the given tuple list.
\qaseparator
\small
\qaitem{1}{Should the maximum difference be calculated between the largest and smallest values found anywhere across all tuples?}{No}
\qaitem{2}{Does ``available pairs'' mean we should only consider pairs of values within the same tuple?}{Yes}
\qaitem{3}{Should the function return 0 when the input list is empty?}{Yes}
\qaitem{4}{If a tuple contains only a single value, should that tuple be ignored?}{Yes}
\qaitem{5}{Are we guaranteed that all elements in the tuples are numeric values?}{Yes}
\qaitem{6}{Should the function handle negative numbers correctly?}{Yes}
\qaitem{7}{Does ``maximum difference between pairs'' require finding the difference for every possible pair combination?}{No}
\qaitem{8}{If the tuple list contains empty tuples, should the function still work?}{Yes}
\qaitem{9}{Should the function consider the absolute difference rather than signed differences?}{Yes}
\qaitem{10}{Is it acceptable to flatten all values from all tuples into a single list?}{Yes}

\medskip
\textit{Q1 corrected Haiku's initial interpretation of ``available pairs''; differences should be computed within each tuple, not across all tuples.}
\end{qatranscript}

\begin{qatranscript}{AIME -- Hard}
\textbf{Problem:} Find the number of cubic polynomials $p(x) = x^3 + ax^2 + bx + c$, where $a, b, c \in \{-20,\ldots,20\}$, such that there is a unique integer $m \neq 2$ with $p(m) = p(2)$.

\medskip
\textbf{Gold Answer:} 738
\qaseparator
\small
\qaitem{1}{Is the factorization $(m-2)[m^2 + (2+a)m + (4+2a+b)] = 0$ correct?}{Yes}
\qaitem{2}{Should we assume $p(m) = p(2)$ is the complete condition?}{Yes}
\qaitem{3}{Is it correct that $q(m) = m^2 + (2+a)m + (4+2a+b) = 0$ must have exactly one integer solution?}{Yes}
\qaitem{4}{For discriminant $= 0$, is $a$ being even necessary and sufficient for the double root to be an integer?}{Yes}
\qaitem{5}{Should roughly 12 even values of $a$ lead to valid $b \in [-20, 20]$?}{Yes}
\qaitem{6}{Should Case 2 include quadratics with two distinct real roots but only one integer root?}{Yes}
\qaitem{7}{Does Case 2 also need to account for quadratics with no real roots?}{Yes}
\qaitem{8}{For each valid $(a,b)$ pair, is $c$ free to vary over all 41 values in $\{-20,\ldots,20\}$?}{No}
\qaitem{9}{Is the answer $(\text{\# valid } (a,b) \text{ pairs}) \times 41$?}{No}
\qaitem{10}{Should the solution verify a concrete example (e.g., $a=0, b=-3$)?}{Yes}

\medskip
\textit{Q8 and Q9 corrected a critical error: Haiku assumed $c$ was free to vary independently, but the condition $p(m) = p(2)$ constrains $c$ through $p(2) = 8 + 4a + 2b + c$.}
\end{qatranscript}

\begin{qatranscript}{HLE -- Medium}
\textbf{Problem:} Let $X$ be the moduli space of nondegenerate lattices in $\mathbb{R}^2$ with unit area. Compute $H_1(X, \mathbb{Z})$.

\medskip
\textbf{Gold Answer:} $\mathbb{Z}$
\qaseparator
\small
\qaitem{1}{Is $X$ correctly identified as $\mathrm{GL}_2(\mathbb{R})^+ / \mathrm{GL}_2(\mathbb{Z})$?}{No}
\qaitem{2}{Does the space of unit-area lattices have two connected components (det $= \pm 1$)?}{No}
\qaitem{3}{Does the Iwasawa decomposition $\mathrm{SO}(2) \cdot A \cdot N$ apply, with $A \cdot N$ contractible?}{Yes}
\qaitem{4}{Is $X$ homotopy equivalent to $\mathbb{H} / \mathrm{PSL}_2(\mathbb{Z})$?}{Yes}
\qaitem{5}{Is $\mathbb{H} / \mathrm{PSL}_2(\mathbb{Z})$ a non-compact space that deformation retracts onto a circle?}{Yes}
\qaitem{6}{Are the singular orbifold points at $i$ (order 2) and $\rho = e^{2\pi i / 3}$ (order 3) correctly identified?}{Yes}
\qaitem{7}{Does non-compactness contribute a generator to $H_1$ from wrapping around the cusp?}{No}
\qaitem{8}{Is $H_1(X, \mathbb{Z}) = \mathbb{Z}$ consistent with $X \simeq S^1$?}{Yes}
\qaitem{9}{Should we verify the quotient doesn't introduce torsion or additional generators?}{Yes}
\qaitem{10}{Is it necessary to distinguish oriented vs.\ unoriented lattices?}{Yes}

\medskip
\textit{Q1 and Q2 rejected Haiku's initial identification of the moduli space, and Q7 corrected a wrong claim about the cusp. After these corrections, Haiku correctly identified $X \simeq \mathbb{H} / \mathrm{PSL}_2(\mathbb{Z}) \simeq S^1$ and computed $H_1 = \mathbb{Z}$.}
\end{qatranscript}

\end{document}